\documentclass[acmlarge,nonacm]{acmart}
\usepackage{subcaption}
\usepackage{wasysym}
\usepackage{makecell}
\usepackage{enumitem}
\usepackage{pifont}% http://ctan.org/pkg/pifont
\AtBeginDocument{%
  \providecommand\BibTeX{{%
    \normalfont B\kern-0.5em{\scshape i\kern-0.25em b}\kern-0.8em\TeX}}}

\setcopyright{none}

\acmJournal{IMWUT}
\acmVolume{1}
\acmNumber{1}
\acmArticle{1}
\acmMonth{5}

\renewcommand{\authorsaddresses}{}
\renewcommand\footnotetextcopyrightpermission[1]{} % removes footnote with conference information in first column
\begin{document}

% \title{Enhancing the Capabilities of a Large Language Model-Based Virtual Doctor with Sensor Data Knowledge}
\newcommand{\cmark}{\ding{51}}%
\newcommand{\xmark}{\ding{55}}%

\newcommand{\workname}{DrHouse} 
% \title{\workname: A Large Language Model Empowered Virtual Doctor System Incorporating Progressive Medical Expertise and Ubiquitous Sensor Data}
% \title{\workname: A Large Language Model Empowered  Virtual Doctor System Incorporating Medical Expertise and Wearable Sensor Data}

\title{\workname: An LLM-empowered Diagnostic Reasoning System through Harnessing Outcomes from Sensor Data and Expert Knowledge}

% \affiliation{%

% \affiliation{%

% \affiliation{%

\author{Bufang Yang}
\authornote{Both authors contributed equally to this research.}
\affiliation{%
  \institution{The Chinese University of Hong Kong}
  % \city{Hong Kong SAR}
  \country{China}}
\email{bfyang@link.cuhk.edu.hk}

\author{Siyang Jiang}
\authornotemark[1]
\affiliation{%
  \institution{The Chinese University of Hong Kong}
  % \city{Hong Kong SAR}
  \country{China}}
\email{syjiang@ie.cuhk.edu.hk}

\author{Lilin Xu}
\affiliation{%
  \institution{The Chinese University of Hong Kong}
  % \city{Hong Kong SAR}
  \country{China}}
\email{lilinxu@cuhk.edu.hk}

\author{Kaiwei Liu}
\affiliation{%
  \institution{The Chinese University of Hong Kong}
  % \city{Hong Kong SAR}
  \country{China}}
\email{1155189693@link.cuhk.edu.hk}

\author{Hai Li}
\affiliation{%
  \institution{The Chinese University of Hong Kong}
  % \city{Hong Kong SAR}
  \country{China}}
\email{lh123@ie.cuhk.edu.hk}

\author{Guoliang Xing}
\affiliation{%
  \institution{The Chinese University of Hong Kong}
  % \city{Hong Kong SAR}
  \country{China}}
\email{glxing@cuhk.edu.hk}

\author{Hongkai Chen}
\affiliation{%
  \institution{The Chinese University of Hong Kong}
  % \city{Hong Kong SAR}
  \country{China}}
\email{hkchen@ie.cuhk.edu.hk}

\author{Xiaofan Jiang}
\affiliation{%
  \institution{Columbia University}
  % \city{New York}
  \country{United States}}
\email{jiang@ee.columbia.edu}

\author{Zhenyu Yan}
\affiliation{%
  \institution{The Chinese University of Hong Kong}
  % \city{Hong Kong SAR}
  \country{China}}
\email{zyyan@ie.cuhk.edu.hk}

% \affiliation{%

% \affiliation{%

% \affiliation{%

% \affiliation{%

% \affiliation{%


\begin{abstract}
Large language models (LLMs) have the potential to transform digital healthcare, as evidenced by recent advances in LLM-based virtual doctors. However, current approaches rely on patient's subjective descriptions of symptoms, causing increased misdiagnosis. Recognizing the value of daily data from smart devices, we introduce a novel LLM-based multi-turn consultation virtual doctor system, DrHouse, which incorporates three significant contributions: 1) It utilizes sensor data from smart devices in the diagnosis process, enhancing accuracy and reliability. 2) DrHouse leverages continuously updating medical databases such as Up-to-Date and PubMed to ensure our model remains at diagnostic standard's forefront. 3) DrHouse introduces a novel diagnostic algorithm that concurrently evaluates potential diseases and their likelihood, facilitating more nuanced and informed medical assessments. Through multi-turn interactions, DrHouse determines the next steps, such as accessing daily data from smart devices or requesting in-lab tests, and progressively refines its diagnoses. Evaluations on three public datasets and our self-collected datasets show that DrHouse can achieve up to an 18.8\% increase in diagnosis accuracy over the state-of-the-art baselines. The results of a 32-participant user study show that 75\% medical experts and 91.7\% patients are willing to use DrHouse.

\end{abstract}

\settopmatter{printfolios=true}
\maketitle
\pagestyle{plain}
\newcommand{\sy}[1]{\textcolor{red}{SY: #1}}
\newcommand{\xl}[1]{{\color{magenta}{#1}}}
\newcommand{\question}[1]{{\color{red}{#1}}}

\section{Introduction}
% [Background of LLMs and virtual doctors]

% Parallel to the views of the fictional character Dr. Gregory House in the popular TV show House\footnote{cite}, instead of relying on subjective symptom descriptions provided by patients, our system takes advantage of more objective data inputs such as wearable sensors, IoT devices, in-lab tests, and etc., as well as utilizing latest medical research in arriving at potential diagnosis ranked by their probabilities.

According to the views of the fictional character Dr. Gregory House in the popular TV show House,\footnote{House, also known as ``House, M.D.'', is an American medical drama television series.} instead of relying on patient's subjective descriptions of symptoms, doctors should take advantage of more objective data inputs such as smart devices, wearable sensors, and in-lab tests for diagnosis.
% , and etc., as well as utilizing latest medical research in arriving at potential diagnosis ranked by their probabilities.
With the realm of large language models (LLMs) \cite{achiam2023gpt,touvron2023llama} and other advanced AI technologies \cite{yang2024viassist}, many variations of LLMs have been developed for medical tasks, such as understanding reports \cite{goel2023llms} and \textit{virtual doctors} \cite{wang2023augmenting,chen2023huatuogpt} that can provide diagnosis based on symptoms. 
However, existing LLM-based medical models still depend on the patient's subjective description of symptoms.
% Recent years, large language models (LLMs) have been witnessed a surge in the utilization of healthcare \cite{wang2023augmenting,chen2023huatuogpt} and many medical LLMs have been proposed recently including fine-tuning-based \cite{singhal2023towards,bao2023disc,chen2023huatuogpt} and retrieval-based models \cite{jin2024health,wang2023augmenting}.
% \textit{While existing works for LLM-based medical applications \cite{singhal2023towards,bao2023disc,chen2023huatuogpt,jin2024health,wang2023augmenting} primarily depend on subjective symptom descriptions provided by patients for health prediction or diagnosis, shown in Table~\ref{tab:Comparison}.}
More importantly, many of these models are trained with old medical data and are not easy to incorporate the most recent medical corpus, functioning more like a single-turn question-answering (QA) system \cite{chen2023huatuogpt}.
They only provide general medical summaries or suggestions instead of 
actively inquiring about additional symptoms or lab tests from patients, thus missing out potential illnesses.
% initiating multi-turn medical consultations like a virtual doctor to inquire about additional symptoms or lab tests from patients, thus missing out potential illnesses.
% actively initiating multi-turn medical consultations for diagnosis.
The inability to quantify the likelihood of each diagnosis also hampers the confidence of both doctors and patients in the diagnostic results provided by these LLM-based virtual doctors.

However, simply incorporating medical conversations into LLMs is not feasible as patients' descriptions can be susceptible to subjective perceptions or memory biases, thus leading to the ambiguity and unreliability of the reported symptoms \cite{mckoane2023diagnostic,meyer2021patient}.
% previous studies in medicine revealed that
Moreover, it can be more challenging for patients to provide precise answers about objective metrics such as respiratory rate and blood oxygen levels.
% \textbf{particularly in the context of LLM-based virtual doctors.}
% This is because patients typically use LLM-based virtual doctors from their homes, without the presence of in-lab tests or healthcare professionals like nurses who could aid them in obtaining these measurements.
Consequently, the subjective perception of patients and uncertainty about their symptoms will hinder timely and accurate diagnosis and treatments \cite{meyer2021patient}.
% [Observations]
In addition, through a comprehensive survey and observation of the medical diagnosis specifications on Up-to-Date,\footnote{Up-to-Date \cite{up_to_date} is a medical knowledge database that contains the latest knowledge and constantly evolving techniques in healthcare.} we found that numerous physiological indicators crucial for diagnosing numerous diseases can be readily obtained through smart devices used in daily life.
Motivated by these observations, this paper aims to incorporate the knowledge from patients' daily sensor data to assist LLM-based virtual doctors with multi-turn diagnosis.

% To mitigate the above issues, this paper aims to 
% incorporate the knowledge from patients' daily wearable sensor data to assist LLM-based virtual doctors with multi-turn disease diagnosis, which is motivated by the following \textbf{novel observations}:
% \begin{itemize}
% \item 
% Previous studies in medicine have revealed that the subjective perception of patients and uncertainty about their symptoms can impede accurate diagnosis \cite{mckoane2023diagnostic,meyer2021patient}, particularly concerning objective physiological metrics.
% We observe that in the context of LLM-based virtual doctors, this problem is more serious as patients typically use LLM-based virtual doctors from their homes, without the presence of in-lab tests or healthcare professionals like nurses who could aid them in obtaining these measurements.

% Differentiating many diseases often relies on a specific key indicator, as numerous diseases present shared symptoms \cite{zhou2014human}. 
% However, patients’ self-reported symptoms tend to be subjective.
% The vague or incorrect description of symptoms can result in the LLM-based virtual doctor drawing inaccurate diagnostic conclusions.

% \item 
% Through a comprehensive survey and observation of up-to-date medical diagnosis specifications \cite{up_to_date,PubMed}, we found that numerous physiological indicators crucial for diagnosing numerous diseases can be obtained through smart devices used in daily life.

% \end{itemize}

% works use sensor data for medical applications.
According to our observations which highlight the potential benefits of incorporating daily sensor data into the diagnostic process, existing studies on LLMs for sensor data primarily focus on utilizing LLMs to interpret sensor signals for understanding the physical world \cite{jin2024position,hota2024evaluating,xu2024penetrative} or make health predictions \cite{nie2024llm, kim2024health}. 
None of the existing works use the sensor data for multi-turn diagnosis.
To address this research gap, we first explore the integration of objective sensor data into the multi-turn diagnosis of LLM-based virtual doctors.
% to explore the design of an LLM-based virtual doctor capable of incorporating patients' sensor data into multi-turn diagnosis.
We summarize several challenges we encountered as follows.
% Although our observations reveal the potential benefits of incorporating daily sensor data into diagnosis, there are still several \textbf{unique challenges} when designing an LLM-based virtual doctor that can integrate patients' sensor data into multi-turn diagnosis.
First, existing LLM-based virtual doctors \cite{singhal2023towards,bao2023disc,chen2023huatuogpt} are only fine-tuned with medical corpus. 
% they lack the ability to inquire about physiological indicators that can be collected from the patient's sensors.
How to enable the LLM-based virtual doctor to follow continuously updated medical diagnostic standards and actively inquire about disease-related physiological indicators that can be obtained from the patient's smart devices is challenging.
Second, the retrieval of required knowledge from the patient's extensive daily sensor databases remains challenging due to the vast amount of data collected and the complex and diverse questions posed by the LLM-based virtual doctor.
Third, sensor data can be influenced by various factors such as environment and calibration, making it challenging to select and integrate the patients' symptom descriptions and the sensor data to determine the next-step actions during multi-turn medical consultations.

% challenging for to choose and fuse the patients' descriptions about their symptoms and to determine the follow-up actions during multi-turn medical consultations.
% patients' descriptions may be influenced by subjective perceptions or memory biases, while sensor data can be affected by environmental factors.
% Enabling LLM-based virtual doctors to integrate these two types of knowledge for decision-making and to generate interpretable diagnostic results is challenging.

\begin{figure}[t]
  \centering
\includegraphics[width=.92\linewidth]{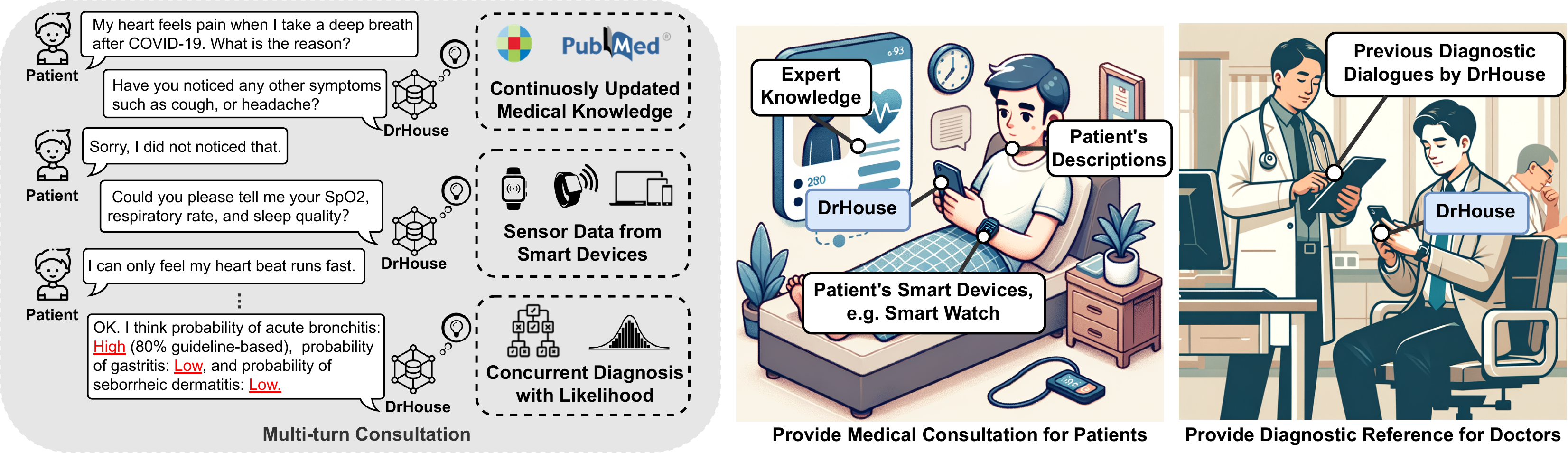}
    \vspace{-5pt}
  \caption{Overview of \workname. 
  \workname~ incorporates patients' sensor data from smart devices into the multi-turn diagnosis process to enhance accuracy and reliability.
  \workname~ can provide medical consultations for patients at their homes, or offer diagnostic references to doctors to reduce their workload.
  }
  \vspace{-10pt}
  \label{fig:overview}
\end{figure}

In this paper, we propose \emph{\workname}, the first LLM-empowered \underline{D}iagnostic \underline{r}easoning system through \underline{H}arnessing \underline{ou}tcomes from \underline{s}ensor data and \underline{e}xpert knowledge.
% leveraging sensor data from patients' smart devices and expert medical knowledge.
% multi-turn consultation LLM-based virtual doctor system that can incorporate knowledge from patients' sensor data and expert medical knowledge for disease diagnosis.
% Figure~\ref{fig:vd_overview} shows the overview of \workname~.
\workname~ first retrieves the latest expert knowledge including potential disease-related diagnosis guidelines and then actively inquires about patients' symptoms through multi-turn medical consultations.
% During the multi-turn diagnosis process,
Through multi-turn interactions, \workname~ integrates patients' symptom descriptions, sensor data from smart devices (e.g., wearable devices), and the latest medical expert knowledge, to determine the next steps such as accessing daily sensor data or requesting in-lab tests.
Simultaneously, \workname~  concurrently evaluates potential diseases, progressively refines its diagnoses, and finally generates an explainable diagnosis with a likelihood for each specific disease.
% first, xxx
% second, use sensor data and symptoms to narrow down
% decision making, in-lab test or query sensor data
% disease list, with referred probability (for doctor to refer)
% \workname~ leverages data from smart devices and mobile phones in patients' daily lives to facilitate decision-making and provide more accurate diagnoses.
% In addition, \workname~ retrieves multiple up-to-date diagnostic guidelines for diseases, enabling simultaneous assessment and diagnosis of multiple conditions while providing explainable diagnosis analysis for each specific disease.
We summarize the contributions of this paper as follows:
\begin{itemize}[leftmargin=*]
\item 
We propose \workname, the first LLM-empowered virtual doctor system that can both initiate multi-turn medical consultations about disease-related physiological indicators and integrate the knowledge from patients' sensor data into the multi-turn diagnosis process to enhance accuracy and reliability.

\item 
We develop three modules to enable \workname~ to incorporate both the latest medical expert knowledge and sensor data from smart devices.
% We address three challenges that impede the implementation of the system. 
% In particular,
% we first develop a mapping-based retrieval of diagnosis guideline trees approach to enable \workname~to actively initiate medical consultations following the up-to-date medical diagnostic standards.
First, we construct the knowledge base in \workname, and develop a multi-source knowledge retrieval approach to iteratively retrieve the required sensor data knowledge and medical expert knowledge during each round of conversations.
Then, we develop a knowledge selection and fusion approach to integrate patient's descriptions, sensor data knowledge, and medical knowledge.
Finally, we design a diagnostic decision-making strategy through knowledge integration and concurrent checking of candidate diseases with likelihood, allowing for more nuanced and informed medical assessments.

% novel concurrent checking of candidate diseases method that not only identifies potential diseases but also quantifies the likelihood of each diagnosis, allowing for more nuanced and informed medical assessments.

% semantic-based adaptive retrieval approach that enables accurate and efficient knowledge retrieval from the patient's sensor data during multi-turn consultations.

% decision-making strategy for LLMs to integrate patient self-reported symptoms and objective sensor data knowledge for decision-making, and provide explainable diagnosis results.

% \item 
% We develop a semantic-based adaptive retrieval strategy that enables accurate and efficient knowledge retrieval from the patient's wearable sensor data during diagnosis. 
% % \sy{In particular, xx}

% \item 
% We design a decision-making strategy for the LLM-based virtual doctor to integrate patient self-reported symptoms and objective sensor data knowledge, and provide explainable diagnosis results. 

\item 
We conduct comprehensive evaluations on three public medical datasets.
We also conduct two types of user studies involving both real patients (N=12) and medical experts (N=20) as participants. 
Evaluations on the three public datasets and two types of user studies show that \workname~ outperforms the state-of-the-art baselines and exhibits promising market potential.

\end{itemize}

\begin{table}[tb]\footnotesize
  \caption{A summary of the recent LLM-based medical applications (\CIRCLE \ means included).}
  \vspace{-1em}
  \label{tab:Comparison}
  \begin{tabular}{ccccccc}
    \toprule
    \textbf{Approach} & \textbf{Base LLM} & \makecell{\textbf{Multi-turn}\\ \textbf{Consultation}} & \makecell{\textbf{Sensor Data}\\\textbf{Knowledge}} & \makecell{\textbf{Diagnosis} \\ \textbf{Guidelines}} & \makecell{\textbf{Latest}\\ \textbf{Knowledge}} &
    \makecell{\textbf{Probabilistic}\\ \textbf{Diagnosis}} \\
    \midrule
    MedicalGPT \cite{MedicalGPT}& LLaMA, Baichuan & \Circle & \Circle & \Circle & \Circle & \Circle\\
    LLM-AMT \cite{wang2023augmenting}&   GPT-3.5, GPT-4, LLaMA-2 & \Circle & \Circle & \Circle & \Circle & \Circle\\
    HuatuoGPT-II \cite{chen2023huatuogpt}&   \makecell{Baichuan2-Base} & \Circle & \Circle & \Circle & \Circle & \Circle\\
    DISC-MedLLM \cite{bao2023disc}&   \makecell{Baichuan2-Base (7B,13B)} & \CIRCLE & \Circle & \Circle & \Circle & \Circle \\
    Med-PaLM 2 \cite{singhal2023towards}&   \makecell{PaLM 2} & \CIRCLE & \Circle & \Circle & \Circle & \Circle\\
    MedDM \cite{li2023meddm}&   GPT-3.5 & \CIRCLE & \Circle & \CIRCLE & \Circle & \Circle\\
    Health-LLM \cite{kim2024health}&   \makecell{Alpaca} & \Circle & \CIRCLE & \Circle & \Circle & \Circle\\
    % Penetrative AI \cite{xu2024penetrative}&   \makecell{GPT-3.5, GPT-4} & \Circle & \Circle & \Circle & \CIRCLE \\
    CaiTI \cite{nie2024llm}&   \makecell{GPT-3.5, GPT-4, LLaMA-2} & \Circle & \CIRCLE & \Circle & \Circle & \Circle\\
    \textbf{\workname~}&   \makecell{GPT-3.5, GPT-4, LLaMA-3} & \CIRCLE & \CIRCLE & \CIRCLE & \CIRCLE & \CIRCLE\\
  \bottomrule
\end{tabular}
\vspace{-5pt}
\end{table}

% \begin{table}\small
%   \caption{A summary of the recent LLM medical applications.}
%   \label{tab:Comparison}
%   \begin{tabular}{cccccc}
%     \toprule
%     \textbf{Approach} & \textbf{LLMs} & \makecell{\textbf{Active Multi-turn}\\ \textbf{Consultation}} & \makecell{\textbf{Diagnosis}\\\textbf{Guidelines}} & \makecell{\textbf{Probabilistic} \\ \textbf{Diagnosis}} & \makecell{\textbf{Sensor Data}\\ \textbf{Knowledge}} \\
%     \midrule
%     MedicalGPT \cite{MedicalGPT}& LLaMA, Baichuan & \xmark & \xmark & \xmark & \xmark \\
%     LLM-AMT \cite{wang2023augmenting}&   GPT-3.5, GPT-4, LLaMA-2 & \xmark & \xmark & \xmark & \xmark \\
%     HuatuoGPT-II \cite{chen2023huatuogpt}&   \makecell{Baichuan2-Base} & \xmark & \xmark & \xmark & \xmark \\
   
%     Health-LLM \cite{kim2024health}&   \makecell{Alpaca} & \xmark & \xmark & \xmark & \cmark \\
%     CaiTI \cite{nie2024llm}&   \makecell{GPT-3.5, GPT-4m LLaMA-2} & \xmark & \xmark & \xmark & \cmark \\
%     % Penetrative AI \cite{xu2024penetrative}&   \makecell{GPT-3.5, GPT-4} & \Circle & \Circle & \Circle & \CIRCLE \\
%     DISC-MedLLM \cite{bao2023disc}&   \makecell{Baichuan2-Base (7B,13B)} & \cmark & \xmark & \xmark & \xmark \\
%     Med-PaLM 2 \cite{singhal2023towards}&   \makecell{PaLM 2} & \cmark & \xmark & \xmark & \xmark \\

%     MedDM \cite{li2023meddm}&   GPT-3.5 & \cmark & \cmark & \xmark & \xmark \\
%     Ours&   \makecell{GPT-4} & \cmark & \cmark & \cmark & \cmark \\
%   \bottomrule
% \end{tabular}
% % \vspace{-1em}
% \end{table}
\section{Related work}

% Figure~\ref{fig:guideline_retrieval} shows that

% \begin{figure}
%   \centering
% \includegraphics[width=0.5\linewidth]{System_design/guideline_retrieval.pdf}
%   \caption{Performance of diagnosis guideline retrieval compared with existing methods.}
%   % \vspace{-.5em}
%   \label{fig:guideline_retrieval}
%   \vspace{-1em}
% \end{figure}

% \textbf{LLMs-based virtual doctor.}
\subsection{LLM-based Virtual Doctors}
\noindent\textbf{Supervised Fine-tuning based Virtual Doctors.}
Several LLMs have been proposed for medical purposes, such as Med-PaLM 2 \cite{singhal2023towards},
DISC-MedLLM \cite{bao2023disc} and HuatuoGPT \cite{chen2023huatuogpt}.
They collect extensive medical datasets comprising question-answering (QA) and diagnosis dialogues, employing the supervised fine-tuning (SFT) technique to LLMs.
These specialized medical LLMs can outperform GPT-4 on several benchmarks in the medical domain.
Google \cite{tu2024towards} employs a self-play-based environment to generate simulated doctor-patient dialogues. 
They fine-tune the LLM based on public medical datasets, real doctor-patient dialogues in hospitals, and simulated dialogues generated by LLMs.
MedDM \cite{li2023meddm} proposes to utilize the knowledge from clinical guidance trees to help LLM better diagnose diseases.
LLaVA-Med \cite{li2024llava} leverages comprehensive datasets of biomedical figure captions to fine-tune LLaVA which is a visual large language model (VLLM). 
LLaVA-Med can perform medical visual question answering (VQA), such as interpreting CT scans.
Other works like AMSC \cite{wang2024beyond} propose to use multiple LLMs working in collaboration for automated diagnosis, with each LLM functioning as a specialized doctor focused on a single subject.
Previous works for LLM-based virtual doctors primarily concentrate on adapting LLMs to the medical domain and overlook the influence of subjects' perceptions.
\workname~employs a different approach to address this challenge by incorporating the objective sensor data from smart devices into the multi-turn medical consultations.

% \noindent\textbf{RAG (Retrieval-Augmented Generation).}
% \subsection{RAG for Medical LLMs}
\noindent\textbf{Retrieval-based Virtual Doctors.}
Due to the significant computational resources and costs for LLM training, several recent studies \cite{jin2024health,wang2023augmenting} have explored the utilization of RAG to enhance LLMs with medical knowledge, eliminating the need for additional training.
LLM-AMT \cite{wang2023augmenting} and Health-LLM \cite{jin2024health} develop a database of medical textbooks for retrieval by LLMs. 
JMLR \cite{wang2024jmlr} employs a joint training approach for LLM and retrieval simultaneously to improve LLM’s ability of medical knowledge reasoning.
MedDM \cite{li2023meddm} constructs a knowledge base consisting of clinical diagnosis guidelines that LLMs can consult and refer to.
% However, MedDM is limited to retrieving only the most similar guideline tree.
However, MedDM retrieves the diagnosis guidelines based on the patient’s symptoms.
The discrepancy between the subjective
expressions of patients and the specialized medical statements in the diagnosis guidelines leads to poor retrieval
performance.
% The diagnosis performance is hindered by the discrepancy between patients' expressions and medical statements, as well as the limited diversity of the retrieved guidelines.
In addition, previous works on retrieval-based virtual doctors primarily focus on retrieving medical knowledge to assist LLM diagnosis, with none exploring the retrieval of sensor data knowledge during multi-turn medical consultations. 
\workname~employs a semantic-based adaptive retrieval approach to accurately retrieve sensor data knowledge from patients' smart devices.
% Besides,\workname~ employs a different approach by indirectly retrieving the clinic symptom database for query augmentation, thus improving the accuracy of diagnostic guideline retrieval.
% Many studies have focused on enhancing the performance of RAG through data augmentation techniques \cite{zhao2024retrieval}.
% The key idea is to augment the information of the query, such as eliminating redundant information, query transform, and incorporating synthetic data.

\subsection{LLM Understanding Sensor Data}

Smart devices such as smartphones and wearable devices have become pervasive in everyday life, serving as passive sensors that effortlessly collect a multitude of data \cite{nickels2021toward}.
Prior research has concentrated on harnessing this wealth of information to develop machine learning models aimed at diagnosing and detecting mental health disorders~\cite{chikersal2021detecting, yang2022novel, nickels2021toward,yang2023brainz}.
The primary limitation of the research mentioned above resides in its reliance on traditional statistical and machine learning approaches, which overlooks the advancements achieved by recent large-scale foundation models~\cite{bubeck2023sparks}. This gap highlights a promising avenue for investigating novel methodologies to analyze sensor data that is passively gathered.
In order to use sensor data to assist LLM for health applications, recent studies aim to use sensor data to assist LLM in understanding the physical world~\cite{jin2024position,hota2024evaluating,xu2024penetrative,kim2024health,englhardt2023classification,ji2024hargpt,yang2024you}. For instance, Kim \textit{et. al}~\cite{kim2024health} and Zachary \textit{et. al} ~\cite{englhardt2023classification} integrate contextual information, such as user demographics and health knowledge, with physiological data, including resting heart rate and sleep duration, to enhance the comprehensive understanding of Large Language Models (LLMs).
In addition, HARGPT~\cite{ji2024hargpt} directly utilizes raw sensor data as input and chain-of-thought prompts to carry out human activity recognition in the physical world.
The work in ~\cite{englhardt2023classification} leverages large language models to synthesize clinically useful insights from multi-sensor data, generating reasons about how trends in data relate to mental conditions.
CaiTI~\cite{nie2024llm} analyzes the user’s daily functioning through several fixed and predetermined dimensions and employs LLM for psychological therapy.
However, CaiTI is only specialized in mental health and lacks the capability to initiate multi-turn medical conversations like a real doctor actively.
Previous work in this area primarily focuses on utilizing LLMs to interpret diverse sensor signals for understanding the physical world.
\workname~ takes a further step to incorporate the knowledge from sensor data into multi-turn medical consultations.

% The closest research of this work is CaiTI~\cite{nie2024llm}, which enables sensor data to build a conversational AI Therapist. \sy{However, this work only focuses on mental health, while \workname~is a general doctor.}

In summary, as shown in Table~\ref{tab:Comparison}, most existing works either focus on adapting LLMs to the medical domain or utilizing LLMs to interpret diverse sensor signals for understanding the physical world.
% prior studies have primarily focused on adapting LLMs to the medical domain through extensive data fine-tuning or RAG techniques.
% On the other hand, previous works have explored the utilization of LLMs to interpret diverse sensor signals for health predictions.
How to incorporate the knowledge of sensor data from patients' smart devices to enhance the capability of LLMs for diagnostic decision-making and multi-turn medical consultations remains unexplored.

% Recently, some studies such as xx and xx~\cite{jin2024position,hota2024evaluating} have used sensor data to assist LLM in understanding this physical world. For instance,  study the capability of LLMs to interpret time-series data~\cite{jin2024position,hota2024evaluating} and Penetrative AI~\cite{xu2024penetrative} processes sensor data into textual signals and digital signals as the input to interact with and reason about the physical world through IoT sensors and actuators. 
\section{Background and Motivation}
In this section, we first show the limitations of existing LLM-based virtual doctors for patients' subjective perception.
Next, we perform a comprehensive investigation of the diagnostic guidelines for various diseases on the Up-to-Date database \cite{up_to_date}.
The key insights from these observations motivate the design of \workname.

\subsection{Potential Misdiagnosis Risks Due to Patient's Inaccurate Subjective Perception}
Previous studies in medicine have revealed that the subjective perception of patients and uncertainty about their symptoms can impede timely and accurate diagnosis and treatments \cite{mckoane2023diagnostic,meyer2021patient}.
In clinical practice, physicians typically inquire about various objective physiological metrics from patients, including respiratory rate, sleep quality, blood oxygen levels, and heart rate \cite{majumder2019smartphone}.
However, it can be challenging for patients to provide precise answers about these objective metrics, particularly in the context of LLM-based virtual doctors.
This is because patients typically use virtual doctors from their homes without the presence of in-lab tests or healthcare professionals like nurses who could aid them in obtaining these measurements.
% concerning objective indicators like daily sleep duration and heart rate. 
% As a result, doctors face difficulties in making accurate diagnostic decisions.

\begin{figure}
  \centering
\includegraphics[width=1\linewidth]{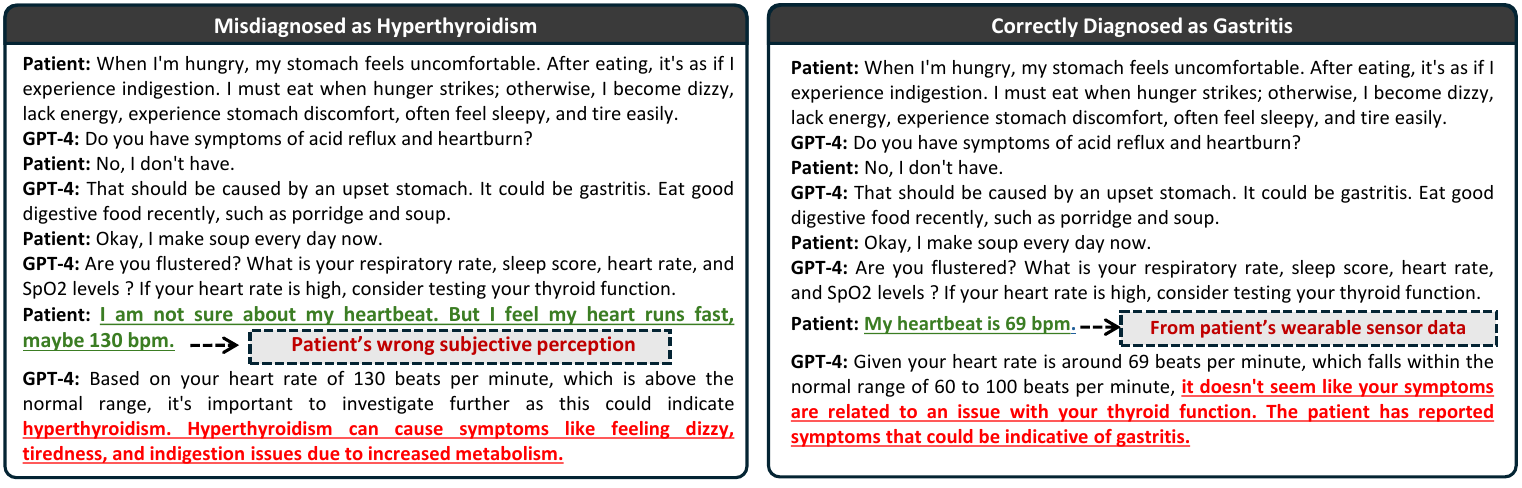}
\vspace{-5pt}
\vspace{-1em}
  \caption{An example of misdiagnosis due to patients' wrong subjective perception.
The words highlighted in green and red represent the patient's subjective descriptions of the symptom and the diagnostic conclusions made by an LLM, respectively.
Note that we use GPT-4 as a naive LLM for example.
}
  \vspace{-1em}
  \label{fig:motivation_misdiagnosis_hyperthyroidism0}
\end{figure}

To demonstrate the presence of similar challenges in LLM-based virtual doctors, we conducted a motivation study to investigate how patients' inaccurate and vague statements affect the diagnostic results provided by virtual doctors. Figure~\ref{fig:motivation_misdiagnosis_hyperthyroidism0} shows an example of the potential misdiagnosis when certain symptoms are missing or inaccurately reported. 
We utilize medical dialogues from the publicly available DIALMED dataset~\cite{he2022dialmed}, with GPT-4 serving as the virtual doctor. 
The actual condition of the patient in the example is gastritis. 
However, the reported symptoms of the patient could be associated with a risk for both gastritis and hyperthyroidism simultaneously.
To differentiate between gastritis and hyperthyroidism, the virtual doctor inquires about the patient's heart rate.
When the patient reports a heart rate falling within the normal range of 60–120 bpm, the virtual doctor can accurately diagnose gastritis.
However, in real consultations, patients often face challenges in accurately recalling objective indicators, experiencing uncertainties, or even inadvertently misreporting them.
Figure \ref{fig:motivation_misdiagnosis_hyperthyroidism0} shows that when the patient subjectively and incorrectly reports a heart rate of 130 bpm, this can cause the virtual doctor to misdiagnose hyperthyroidism.

% \begin{figure*}[h]
%   \centering
% \includegraphics[width=0.8\linewidth]{Motivation/motivation_misdiagnosis_hyperthyroidism.png}
%   \caption{Misdiagnosis may occur when specific symptoms are not present or accurately reported.}
%   % \vspace{-.5em}
%   \label{fig:motivation_misdiagnosis_hyperthyroidism}
%   % \vspace{-1em}
% \end{figure*}

\subsection{Sensor Data Indicators in Medical Diagnostic Guidelines}
% Most existing medical LLMs only focus on using medical QA corpus to fine-tune the base LLM \cite{singhal2023towards,chen2023huatuogpt, bao2023disc} or retrieving external medical databases like textbooks \cite{wang2023augmenting,jin2024health}.
% Compared with such knowledge,

\begin{figure}
  \centering
\includegraphics[width=0.9\linewidth]{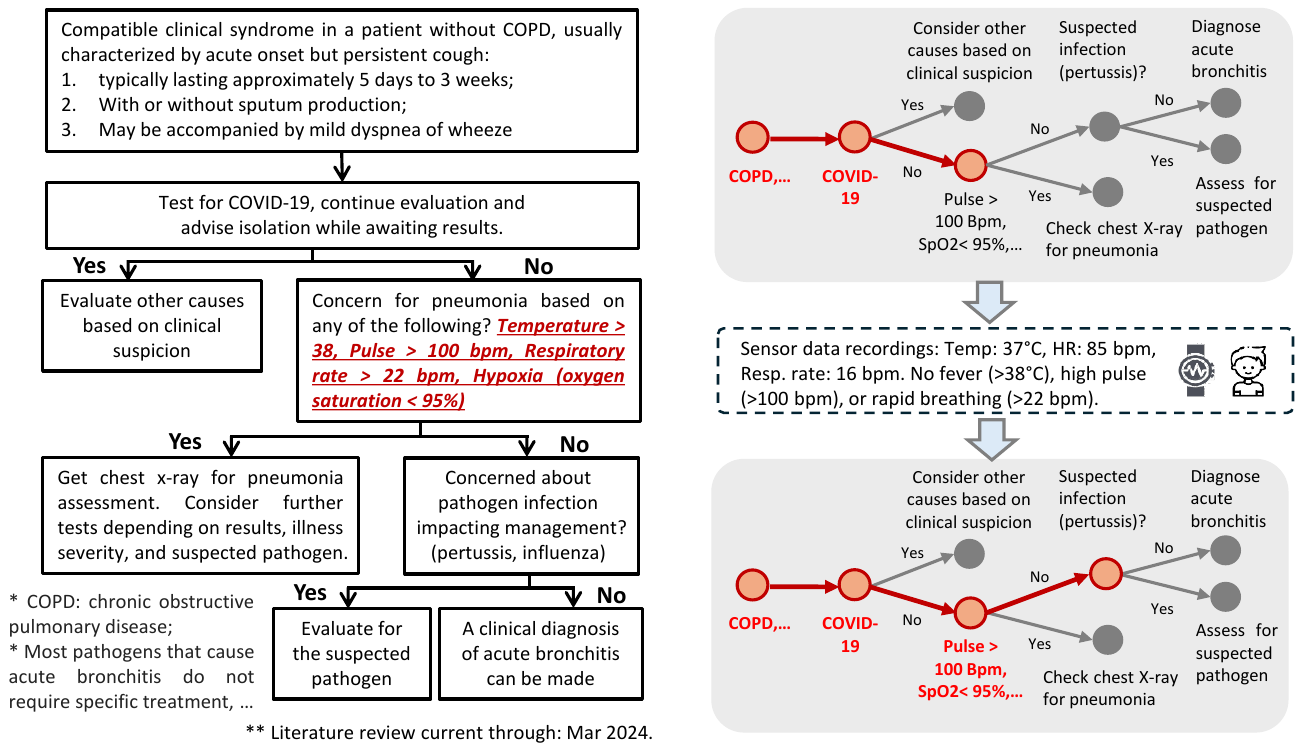}
  \vspace{-5pt}
  \caption{An example of sensor data metrics in medical diagnosis guidelines.
  The left figure shows the latest medical diagnostic guidelines for acute bronchitis on the Up-to-Date database.
  The right figure shows that accessing patients' daily sensor data can take the diagnostic process one step further and assist virtual doctors in decision-making.}
  \vspace{-15pt}
\label{fig:motivation_diagnosis_guideline}
\end{figure}

Medical diagnosis guidelines \cite{li2023meddm} serve as the gold standard that doctors directly consult and utilize in clinical practice, enabling LLM to learn with minimal data and execute actions and decisions that closely align with those of real doctors.
These diagnostic guidelines typically employ a tree-like structure. 
Doctors will inquire about the patient's symptoms and subsequently make informed decisions and diagnoses by referencing the diagnosis guidelines.
Furthermore, our observations indicate that numerous latest medical diagnosis guidelines incorporate the assessment of diverse physiological indicators, which can be conveniently obtained through smart devices such as FitBit~\cite{fitbit}, worn by patients.
For example, Figure~\ref{fig:motivation_diagnosis_guideline} shows the medical diagnosis guideline for bronchitis from the Up-To-Date database \cite{up_to_date}. 
In step four, the physician assesses the risk of pneumonia by examining physiological indicators obtained from sensor data, such as respiratory rate and blood oxygen saturation. 
Accessing patients' daily sensor data directly from smart devices can assist the virtual doctor in decision-making and enhance the probability of accurate diagnosis.

\subsection{Summary}
% 这样和概率绑起来；
% 加速收敛；
We summarize the motivations as follows.
First, the patient's subjective perception, especially for those objective physiological indicators can lead to potential misdiagnosis risks of LLM-based virtual doctors.
Second, our observations on the latest medical databases show that many diagnosis guidelines include indicators that can be obtained from daily sensor data of smart devices.
Integrating sensor data from smart devices into the diagnostic process of LLM-based virtual doctors is an effective approach to ensuring the reliability of their diagnoses.
Third, the action space of existing virtual doctors \cite{chen2023huatuogpt,bao2023disc} is limited to inquiring about patients' symptoms and requesting in-lab tests.
Accessing patients' sensor data from smart devices offers virtual doctors an additional option, expanding their action space.
This expansion not only assists in cutting down on unnecessary in-lab tests but also facilitates patients in providing subjective answers, thereby reducing the risk of misdiagnoses.

\section{System Design}
\subsection{System Overview}
\begin{figure}
  \centering
\includegraphics[width=1\linewidth]{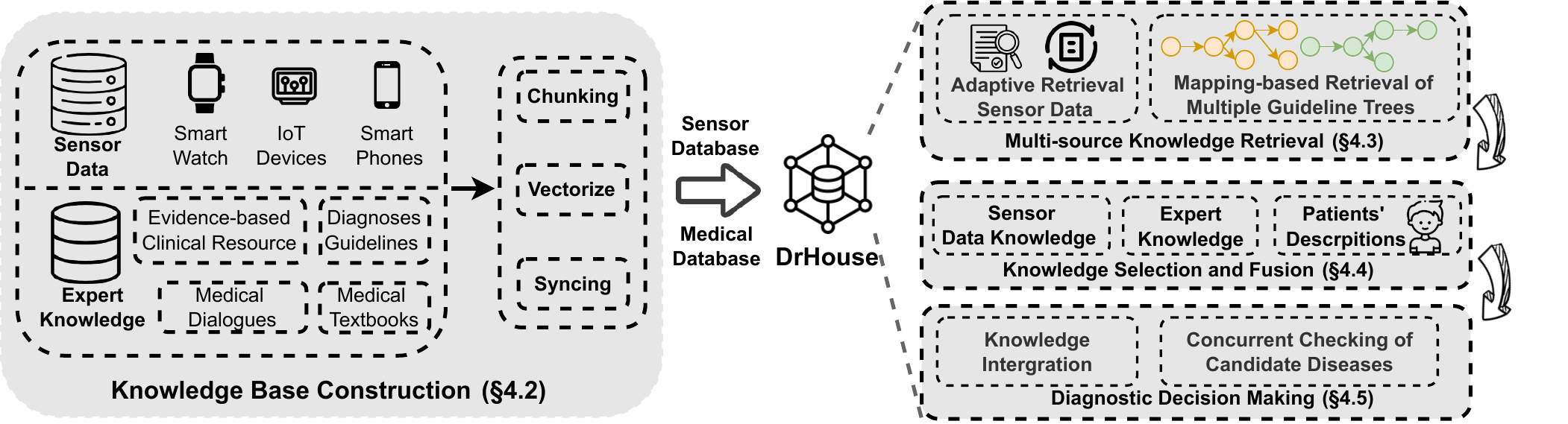}
\vspace{-1.0em}
  \caption{System overview of \workname.}
  % \vspace{-.5em}
  \label{fig:overview}
  \vspace{-1em}
\end{figure}
\workname~ is an LLM-empowered multi-turn consultation virtual doctor system capable of leveraging both the patient's descriptions of their symptoms, patients' sensor data from smart devices, and medical expert knowledge for diagnosis.
Figure~\ref{fig:overview} overviews the design of the \workname system.
The left part of Figure~\ref{fig:overview} shows \workname's construction of two knowledge bases. In particular, \workname~ constructs knowledge bases (\S~\ref{Knowledge Base of DrHouse}) using daily sensor data from patients' smart devices and continuously updated medical expert knowledge, respectively. After construction, \workname~ processes the data into databases for LLM's usage.
% The knowledge sources are subsequently chunked and vectorized as vector databases, which serve as the sensor data knowledge base and medical knowledge base in \workname, respectively.
The right part of Figure~\ref{fig:overview} illustrates the three runtime technologies to support multi-turn medical consultations.
% During multi-turn medical consultations (right part of Figure~\ref{fig:overview}), 
First, \workname~ employs two approaches to iteratively retrieve the required sensor data and the latest medical knowledge during each round of conversations, respectively (\S~\ref{Knowledge Retrieval}).
% a multi-source knowledge retrieval approach (\S~\ref{Knowledge Retrieval}) to iteratively retrieve the required sensor data knowledge and medical knowledge during each round of conversations.
Then, to make LLM understand the retrieved knowledge, \workname~ employs a knowledge selection and fusion approach to integrate patients' descriptions, sensor data knowledge, and medical knowledge (\S~\ref{Prompting for DrHouse}).
At the same time, to generate diagnosis results, \workname~ oversees the integrated knowledge from sensors, expert medical knowledge, and in-lab tests for estimating the probabilities of multiple candidate diseases (\S~\ref{decision making section}).

\subsection{Knowledge Base Construction}
\label{Knowledge Base of DrHouse}
% Figure~\ref{fig:database} shows the knowledge base framework of \workname, which contains two parts: the medical knowledge base and the sensor data knowledge base.
In this section, we introduce the knowledge base in \workname, including the two types of knowledge sources, the latest clinical resources, and the vectorization and synchronization of the knowledge base.
% The knowledge base in \workname~ contains two parts: the medical knowledge base and the sensor data knowledge base.
% We also introduce the knowledge sources, knowledge base construction, and syncing in turn. 
% This subsection introduces the sensor data knowledge sources, medical expertise sources, knowledge base construction, and syncing.

% \begin{figure}
%   \centering
% \includegraphics[width=0.95\linewidth]{System_design/database.png}
%   \caption{Knowledge base framework of \workname.}
%   % \vspace{-.5em}
% \label{fig:database}
%   % \vspace{-1em}
% \end{figure}

\subsubsection{Medical Expert Knowledge}
% The medical knowledge sources in \workname~ contain two parts, including static medical knowledge and up-to-date medical knowledge. 
% Specifically, static medical knowledge consists of multi-turn medical dialogues, medical textbooks, and diagnosis guidelines.
The medical knowledge sources in \workname~ consist of multi-turn medical dialogues, medical textbooks, and diagnosis guidelines.
% They have been integrated into the knowledge sources since its construction.
% Up-to-date medical knowledge encompasses the latest and most current medical guidelines and papers obtained from web databases, including Up-to-date \cite{up_to_date} and PubMed \cite{PubMed}.

\noindent\textbf{Medical Dialogues.}
% \textcolor{blue}{follow up a ...,
% We...}
We will first teach DrHouse the doctor's language to raise multi-turn conversations guiding the patients to provide appropriate descriptions of their diseases.
% Many existing works on medical LLMs only function as a single-turn QA system \cite{zeng2020meddialog,jin2021disease}, providing medical suggestions.
% % focus on evaluating the task of single-turn medical question-answering (QA) \cite{zeng2020meddialog,jin2021disease}.
% However, during real-world medical consultations, doctors typically inquire about the patient's symptoms progressively, step-by-step, before arriving at a definitive diagnosis.
Therefore, our medical knowledge base contains diverse multi-turn dialogues, enabling LLMs to emulate the role of a doctor and facilitate multi-turn medical consultations.
Two public medical dialogue datasets MedDialog \cite{zeng2020meddialog} and MedDG \cite{liu2022meddg} are stored in \workname's medical knowledge base.

\noindent\textbf{Medical Textbooks.}
Despite being trained extensively on massive datasets, LLMs such as GPT-4 still lack the professional knowledge required in the medical field.
This prevents LLMs from giving professional diagnoses and advice.
% For example, current LLMs lack expert medical knowledge such as ``Upper endoscopy is not required in the presence of typical Gastroesophageal reflux disease (GERD) symptoms of heartburn or regurgitation''.
Therefore, we teach general medical knowledge including the definition of medical terms and normative range of medical indicators, e.g., ``Upper endoscopy is not required in the presence of typical Gastroesophageal reflux disease (GERD) symptoms of heartburn or regurgitation''.
% and \textcolor{blue}{xxx}.
A large number of medical textbooks, including MedQA \cite{zeng2020meddialog} and PubMedQA \cite{jin2019pubmedqa}, are stored in the medical knowledge base of \workname.

\noindent\textbf{Diagnosis Guidelines.}
\label{Diagnosis guideline trees}
Diagnosis guidelines are used by doctors to confirm diseases based on the patient's symptoms and in-lab test results \cite{turner2009abstraction,li2023meddm}. 
The format of such guidelines can be formatted into a decision tree.
\workname~can learn from massive knowledge of diseases to ask relevant follow-up questions to achieve a final diagnosis result. 
In particular, we leverage an optical content recognition model (PaddleOCR \cite{PaddleOCR}) to transform figure-based diagnosis guidelines into if-else formatted \textbf{diagnosis guideline trees}.
Subsequently, a manual inception is performed to ensure the correctness.
Finally, we incorporate these diagnosis guidelines into the prompt of the LLM, providing professional medical diagnoses.

% In clinical diagnosis, doctors usually refer to diagnosis guidelines for decision-making \cite{turner2009abstraction,li2023meddm}, which can be seen as the gold standard.
% Therefore, we first gathered a comprehensive range of medical guidelines
% and treatment papers spanning various diseases from public medical databases \cite{PubMed}.
% % from PubMed \cite{PubMed} and Up-to-date \cite{up_to_date} spanning various diseases.
% Subsequently, inspired by MedDM \cite{li2023meddm}, we employ a combination of automatic guideline extraction and manual inception to produce diagnosis guidelines in a format readable by LLMs.
% Specifically, we leverage an OCR model, PaddleOCR \cite{PaddleOCR}, to transform figure-based diagnosis guidelines into if-else formatted \textbf{diagnosis guideline trees}.
% Subsequently, a manual inception is performed to ensure the correctness.
% Finally, we incorporate these diagnosis guidelines into the prompt of the LLM, providing professional medical diagnoses.

\subsubsection{Sensor Data} 
\label{sec:SensorDataKnowledgeSources}
\workname~ will access the patient's daily sensor data from smart devices during the diagnosis.
This subsection details the sensor data knowledge sources in \workname.

\noindent\textbf{Knowledge Source of Sensor Data.}
The sensor data knowledge source contains daily data collected from patients' smart devices, such as the Fitbit Sense smartwatch \cite{fitbit}.
The recorded data includes a range of physiological indicators of the patient over each time interval, such as step count, sleep scores, oxygen saturation (SpO2), heart rate, and stress scores. 
These physiological indicators can provide LLM-based virtual doctors with reliable evidence for diagnosis.
% In this study, we use the LifeSnaps dataset \cite{yfantidou2022lifesnaps} as the sensor data knowledge base.
% \textcolor{blue}{Deploy, cloud?(data storage) frequency? Outlier values? history data/real-time data}
Patient sensor data are stored on their local device or their personal cloud storage like iCloud Drive.
% Only when \workname~ initiates the diagnosis process, will it access the patient's sensor database and retrieve the required knowledge.
% In practical implementation, we 
% every patient has a personal database on the cloud, which contains corresponding sensor data in the past period $T_{store}$.
% $T_{update}$
% \textcolor{blue}{As the user's sensor data is updated in real-time, the sensor data knowledge base only stores sensor data in a specific period.  
% Newly received data automatically replaces any expired old sensor data.
% }

\noindent\textbf{Check of Sensor Data's Uncertainty.}
Although patient sensor data is generally more objective than patient statements, it can still be susceptible to data quality issues due to technical malfunctions or motion artifacts.
If the uncertainty of data from smart devices is high, a virtual doctor should request the patient to perform in-lab tests instead of solely relying on the sensor data.
To this end, we profile the uncertainty level for each sensor data and store them in the knowledge sources.
In particular, we use the probability density function~\cite{ito1984introduction} to represent the degree of deviation. Note that the lower of this value indicates a higher degree of deviation. 
% Note that we assume that the distribution of the sensor data generally follows the Gaussian distribution. 

% (For more details about the decision-making of \workname~ please refer to \S~\ref{decision making section}).

% \subsubsection{Sensor Data Uncertainty Check} \sy{In scenarios where discrepancies arise between patient-reported symptoms and sensor-derived data, it is imprudent to exclusively rely on either source. Both patient descriptions and sensor data are susceptible to various forms of error. 
% Patient accounts may be influenced by subjective perceptions or memory biases, whereas sensor data can be compromised by technical malfunctions or improper calibration. 
% Consequently, the uncertainty of each data source must be critically assessed. 
% }
% Therefore, in this subsection, we explore methodologies for evaluating the uncertainty of sensor data. 
% In particular, we use the probability density function (PDF) to represent the degree of deviation. Note that we assume that the distribution of the sensor data generally follows the Gaussian distribution and the lower this value indicates a higher degree of deviation.

% \begin{equation}
% Z_i = \frac{x_i - \overline{X}}{S}, \quad f(Z_i) = \frac{1}{\sqrt{2\pi}} e^{-\frac{Z_i^2}{2}},    \notag
% \end{equation}

% where, \(x_i \in \{x_1, x_2, \dots, x_n\} \) is the sensor data points, $n$ is the window size , \( \overline{X} \) is the mean and \( S \) is the variance.  

\subsubsection{Latest Clinical Resources.}
We leverage evidence-based clinical resources to enable continuous learning from the latest clinical research output.
\workname~ will embed the lifelong learning capability like the doctors by using evidence-based clinical resources as a knowledge resource. 
Up-to-date \cite{up_to_date} and PubMed \cite{PubMed} are two popular such data resources that contain evidence-based clinical cases, the latest research outputs, disease management, and treatment.
To ensure our model remains at the forefront of medical diagnosis standards, \workname~ continuously updates these evidence-based clinical resources from Up-to-date and PubMed.

% Lifelong learning is crucial to the doctors' career due to the continuous development of medical education and professional development \cite{up_to_date}.
% Lifelong learning holds equal importance for LLM-based virtual doctors, yet it is often overlooked by the majority of previous studies.
% In the Up-to-date database \cite{up_to_date}, we observe that new disease diagnosis guidelines are regularly updated, with releases occurring nearly every month. 
% For example, following the outbreak of COVID-19, testing for the COVID-19 virus has been incorporated as a second-step measure in the diagnostic criteria for acute bronchitis, as shown in Figure~\ref{fig:motivation_diagnosis_guideline}. 
% Therefore, it is essential to integrate the latest medical diagnosis guidelines into the virtual doctor.
% \workname~ consistently updates the medical textbooks and medical diagnosis guidelines in the medical database with the latest knowledge from Up-to-date \cite{up_to_date} and PubMed \cite{PubMed} to ensure our model remains at the
% forefront of medical diagnosis standards.

\subsubsection{Knowledge Base Vectorization and Synchronization}
The medical and sensor data knowledge sources mentioned above are stored in the format of text.
To facilitate the knowledge querying by \workname~ during the diagnosis process, we convert these text knowledge sources into vector databases in advance, which is called \textbf{knowledge base} in this paper.
Specifically, we first split the raw document into multiple chunks.
These chunks are converted to embedding vectors using the OpenAI embedding model \texttt{text-embedding-ada-002} \cite{neelakantan2022text}.
The knowledge databases are subsequently utilized for knowledge retrieval in response to queries made by patients or virtual doctors.
\workname~ employs an event-driven synchronization strategy to dynamically embed knowledge, including both the latest medical knowledge and continual sensor data, into the knowledge database.
The synchronization process for the sensor data is triggered every hour.
The medical knowledge is synchronized when newer medical guidelines are online.

\subsection{Multi-source Knowledge Retrieval}
\label{Knowledge Retrieval}
% Existing LLM-based virtual doctors \cite{singhal2023towards,bao2023disc,chen2023huatuogpt} are only fine-tuned with medical corpus, they lack the ability to inquire about disease-related physiological indicators that can be collected from the patient's wearable sensors.
% Additionally, the questions posed by the LLM-based virtual doctor can be complex and diverse, posing challenges in knowledge retrieval of patients' daily sensor data.
Existing LLM-based virtual doctors \cite{singhal2023towards,bao2023disc,chen2023huatuogpt} only retrieve medical knowledge based on patients descriptions.
\workname~ takes a unique approach by multi-source knowledge retrieval, enabling it to leverage sensor data from smart devices and medical expert knowledge simultaneously.
Specifically, we first develop a mapping-based guideline tree retrieval approach to enable \workname~ to accurately retrieve diagnosis guideline trees.
In addition, we develop a semantic-based adaptive retrieval approach that enables accurate and efficient knowledge retrieval of sensor data from the patient's smart devices during multi-turn diagnosis.

% To address these challenges, we propose two approaches: query-augmented guideline tree retrieval and adaptive sensor data retrieval filtering.
% }
% This subsection details the process of \workname~ performing knowledge retrieval during multi-turn medical consultations, including diagnosis guideline trees, patients' sensor data knowledge, and medical expertise. 

% Guideline tree retrieval (progressively, narrow down)
% sensor data retrieval
% other medical knowledge retrieval

% The knowledge retrieval process comprises two stages: knowledge retrieval during the start-up stage of diagnosis (where the patient first reports their symptoms) and knowledge retrieval during multi-turn diagnosis.

\subsubsection{Retrieval of Guidelines Trees}
\label{Query-augmented Guideline Tree Retrieval}
% This subsection details the process of acquiring sensor data and medical knowledge in each round of conversation throughout the diagnostic procedure.

% \noindent\textbf{Query-augmented Guideline Tree Retrieval.}

% The query augmentation module in our system operates in two modes: pre-diagnosis and during diagnosis. 
% This section introduces the module's functionality before diagnosis, i.e., augmenting the query while retrieving medical guideline trees.
Accurate retrieving relevant medical diagnosis guidelines based on patients' self-reported symptoms in the initial round of conversation is challenging.
MedDM \cite{li2023meddm} directly retrieves the guideline based on patients' descriptions.
However, the \textbf{discrepancy} between the non-specialized expressions of patients and the specialized medical statements in the guideline tree results in poor retrieval performance.
Our results show the accuracy of directly retrieving guidelines based on the patient's descriptions is only about 37\%, suggesting a huge gap between patients' expressions and the statements in guidelines trees (For more details please refer to
\S~\ref{guideline_retrieval_performance}).

% Figure~\ref{fig:guideline_retrieval} shows the performance of directly retrieving guidelines based on the patient's symptoms. 
% We test the performance with varying parameters: vector database chunk sizes are 200 (MedDM-0), 500 (MedDM-1), and 1000 (MedDM-2), respectively, while maintaining a fixed overlapping rate of 50\%.
% Results show that the retrieval accuracy is only about 37\%, suggesting a huge gap between patients' expressions and the statements in guidelines trees. 

\noindent\textbf{Mapping-based Retrieval of Multiple Guidelines Trees.}
To address this challenge, we design a mapping-based retrieval of multiple guideline trees approach, as shown in Figure~\ref{fig:guideline_retrieval_method}.
% Figure~\ref{fig:guideline_retrieval_method} shows the comparison between our approach for diagnosis guideline retrieval and MedDM \cite{li2023meddm}.
Specifically, we first retrieve the top-k similar patients' symptoms from a pre-collected symptom-disease dataset \cite{wang2008automated}.
This dataset comprises a vast collection of commonly encountered clinic symptoms, each associated with multiple potential disease labels.
Once we acquire the top-k potential diseases of the patients, we can precisely obtain the diagnosis guideline trees that are associated with these corresponding diseases.
Finally, we incorporate these top-k guideline trees into the prompt of \workname, and input to the LLM (For more details please refer to \S~\ref{Prompting for DrHouse}).

% \begin{figure}
%   \centering
% \includegraphics[width=0.55\linewidth]{System_design/guideline_retrieval_method.png}
%   \caption{Our approach of diagnosis guideline retrieval.}
%   % \vspace{-.5em}
%   \label{fig:guideline_retrieval_method}
%   % \vspace{-1em}
% \end{figure}

\noindent\textbf{Dynamic Guideline Tree Update.}
As multiple rounds of medical consultations progress, \workname~ accumulates an increasing number of patient symptoms, resulting in the change of potential diseases. Consequently, after each round of consultation, \workname~ dynamically updates the top-k diagnosis guideline trees to adapt to the changes in symptoms.
On the other hand, the increased number of patient symptoms also helps \workname~ to narrow down the potential diseases.
% (For more details please refer to \S~\ref{decision making section})

\subsubsection{Semantic-based Adaptive Retrieval for Sensor Data Knowledge}
\label{Adaptive Sensor Data Knowledge Retrieval}
In this subsection, we first introduce the basic pipeline of sensor data knowledge retrieval in \workname.
Then, we show our motivations and the design of our semantic-based adaptive retrieval
approach. 

\noindent\textbf{Retrieval Pipeline.}
The pipeline of sensor data retrieval in \workname~contains two processes: raw sensor data retrieval and summarizing.
Specifically, we first retrieve the raw context of the sensor data recordings based on the query posed by the virtual doctor.
This process can be expressed as $c_{sensor} = \texttt{Retriever} (q_{doctor}, D_{sensor})$, where $\texttt{Retriever}$ is the retrieval model, $q_{doctor}$ is the query (question) of the virtual doctor, $D_{sensor}$ is the sensor data knowledge base.
The $\texttt{Retriever}$ model is to calculate the cosine similarity score between the query vector and stored embeddings, with a retrieval considered successful if the score exceeds a predefined threshold~\cite{englhardt2023classification}.
We use the embedding model $\texttt{text-embedding-ada-002}$ from OpenAI as $\texttt{Retriever}$ in this study.
Subsequently, we employ a summary LLM called $\texttt{LLM}_{\texttt{sum}}(\cdot)$ to interpret and summarize the raw context $c_{sensor}$, which can be expressed as $d_{sensor}=\texttt{LLM}_{\texttt{sum}}(q_{doctor}, c_{sensor})$.
Compared to the raw context, the summary output $d_{sensor}$ is more readable thus better suited for \workname.
Considering costs and expenses, we
employ GPT-3.5 as the summary LLM in this study.
% To ensure the accuracy and efficiency of sensor data knowledge retrieval, we develop an adaptive sensor data retrieval filtering approach, which can be seen in \S~\ref{Adaptive Retrieval}.
Next, we will introduce the motivation of semantic-based retrieval and how it works in the pipeline.

\begin{figure}
    \centering
    \begin{subfigure}{0.415\columnwidth}
        \centering
        \includegraphics[width=0.95\textwidth]{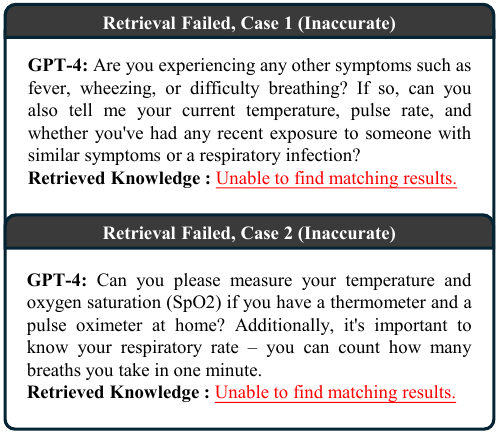}
        \vspace{-0.5em}
        \caption{An example of inaccurate sensor data retrieval.
        % The retrieval threshold is set to 0.8.
        % retrieval failure case when the threshold is set to 0.8.
        }
        \vspace{-1em}
\label{fig:bad_case_retrieval_failed}
    \end{subfigure}
    \hfill
    \begin{subfigure}{0.552\columnwidth}  
        \centering 
        \includegraphics[width=0.95\textwidth]{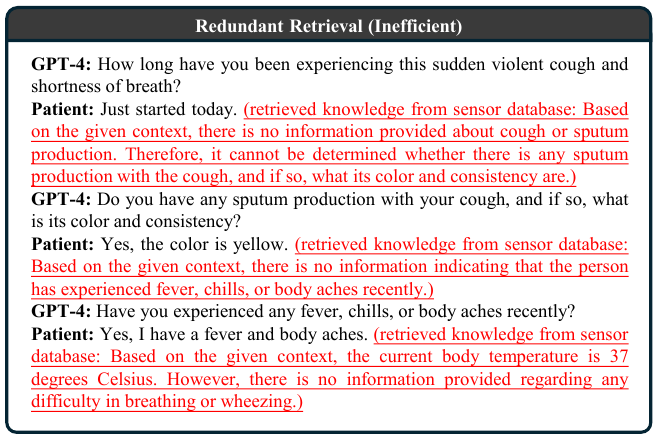}
        \vspace{-0.5em}
  \caption{An example of sensor data retrieval. 
  % The retrieval threshold is set to 0.3.
  }
    \vspace{-1em}
\label{fig:bad_case_retrieval_redundant}
    \end{subfigure}
     % \vspace{-1.0em}
    \caption{Examples of inaccurate and inefficient sensor data knowledge retrieval.
    The words highlighted in red represent the results of sensor data knowledge retrieval.
    Note that we use GPT-4 as a naive LLM for example.
    }
    \label{fig:bad_case_retrieval}
      \vspace{-1.2em}
\end{figure}

\noindent \textbf{Observations and Motivation.}
% \workname~ leverages the outputs from LLM to query a sensor data knowledge base, which is structured as a vector database. Generally, querying a vector database typically involves calculating a similarity score between the query vector and stored embeddings, with a retrieval considered successful if the score exceeds a predefined threshold~\cite{englhardt2023classification}.
% However, 
% We observe that such naive approach can not directly adopted to our setting. 
% The main reason is that the threshold is generally hard to choose, especially in the heterogeneous environments. 
% For instance, we show two experiments to demonstrate our motivation. 
Figure~\ref{fig:bad_case_retrieval_failed} shows that the outputs of LLM doctors can contain very complex and diverse actions, such as: ``\textit{Are you experiencing any other symptoms such as fever, wheezing, or difficulty breathing? If so, can you also tell me your current temperature, pulse rate, and whether you've had any recent exposure to someone with similar symptoms or a respiratory infection}''.
The keywords related to sensor data indicators can be overwhelmed by a large number of words, making sensor data retrieval challenging.
As mentioned above, sensor data retrieval requires a predefined threshold to determine whether the retrieved contexts should be used.
A higher threshold can obscure or overwhelm the keywords for retrieving the sensor database, leading to retrieval failure. On the other hand, shown in Figure~\ref{fig:bad_case_retrieval_redundant}, a lower threshold could lead to many redundant retrievals, thereby increasing system overhead. 
Figure~\ref{fig:Retrieval_time} shows the overhead of sensor data retrieval.
Since the process of sensor data knowledge retrieval contains both data retrieval and LLM summarizing, retrieving sensor knowledge in each round of diagnosis will bring a large overhead, including both latency and cost.

\noindent \textbf{Semantic-Based Retrieval Filtering.}
To tackle the challenge mentioned above, we recognize that not all queries necessitate the retrieval of sensor data. 
One straightforward way is to employ matching-based methods to measure the similarity between two texts, such as the Jaro similarity algorithm~\cite{jaro1989advances}.  
% such as hard matching, which relies on the similarity between two strings. In particular, we use  
% \begin{equation}
%     r = \frac{2.0 \times Matches }{\text{len}(s1) + \text{len}(s2)}, \notag
% \end{equation}
% where $r$ denotes the ratio similarity of two sequences. $Matches$ is the count of matched characters in the two tokenizations, $\text{len}(\cdot)$ denotes the length of tokenization.
% In practice, we use Jaro similarity algorithm~\cite{jaro1989advances} to compute the matches.
However, we observe that such direct approaches do not achieve satisfactory performance, as detailed in \S~\ref{sec:exp:impact_hyper}. The main reason is that the same meaning can be expressed in multiple ways, such matching-based methods do not work since we cannot dictate how the doctor should ask the question. Therefore, we design an adaptive retrieval filtering mechanism
guided by semantic information to categorize the inquiries posed by doctors to ascertain whether the retrieval of sensor data is required. In particular, we train a semantic filter i.e., a Bert-based binary classifier \cite{devlin2018bert} with the cross-entropy loss. It determines whether to initiate retrieval of sensor data.
To obtain the training dataset, we gathered queries from our virtual doctor platform and annotated them with labels. In addition, we also utilize GPT-4 to augment the samples, addressing the issue of insufficient data in the original dataset.

\subsubsection{Medical Knowledge Retrieval}
\workname's retrieval of medical expert knowledge contains two stages: preceding retrieval and runtime retrieval.
The first stage equips LLMs with the capabilities required for multi-turn medical conversations, while the second stage empowers LLMs with diverse medical expert knowledge.

\noindent\textbf{Preceding Retrieval.}
% Merely inputting a patient's symptoms directly into the LLMs can solely provide medical suggestions and summarizations, resembling a single-turn QA procedure rather than the iterative multi-turn consultations in clinical diagnosis \cite{fan2024ai}.
\workname~retrieves relevant dialogue demonstrations based on the symptoms described by patients during the first round of conversation, which is also called the preceding stage.
These demonstrations cover multi-turn dialogues of diverse diseases.
By incorporating this information into the prompt, \workname~ can have the capability to perform multi-turn medical consultations.
The performance of LLMs will be enhanced with an increase in the number of demonstrations and their relevance to the patient's descriptions.
However, due to the limited length of the LLMs processing context \cite{li2024long}, \workname~ only retrieves the top-k similar dialogues from the vector database and incorporates them into the prompt.
Considering the length of the context and cost, $k$ is set to 3 in our experiments.

% Specifically, \workname~
% at the first round of conversation when patients describe their symptoms
% % Retrieving essential medical knowledge at the start-up stage of diagnosis is crucial to enable LLMs to commence diagnosis more effectively. 
% This retrieval process relies on the self-reported symptoms provided by the patient.
% (\textbf{Multi-turn Dialogue Retrieval.})
% Incorporating multi-turn dialogue demonstrations into the prompt can equip LLMs with the capability to perform multi-turn medical consultations.
% The performance of LLMs will be enhanced with an increase in the number of demonstrations and their relevance to the patient's symptoms.
% However, due to the limited length of the LLMs processing context \cite{li2024long}, we must select the most relevant demonstrations incorporated into the prompt.
% % During the offline stage, we utilize various publicly accessible medical multi-turn dialogue datasets, including MedDialog \cite{zeng2020meddialog} and MedDG \cite{liu2022meddg}. 
% % Employing Bert ~\cite{devlin2018bert} as the embedding model, we calculate the embedding of each dialogue data and store it in a vector database.
% To this end, we retrieve the top-k similar dialogues from the vector database and incorporate them into the prompt.
% Considering the length of the context and cost, $k$ is set to 3 in our experiments.
% % where $k$ is a hyper-parameter to decide how many dialogues we selected. In our practice, $k=3$ for simplicity.  
% % \textcolor{blue}{More details of analysis $k$ are referred to Appendix}.

\noindent\textbf{Runtime Retrieval.}
At each round of conversation, \workname~ retrieves the medical knowledge based on the patient's reported symptoms.
This process enables \workname~ to provide a more professional reply and mitigate LLM hallucinations \cite{wang2023augmenting,jin2024health}.
% However, this process can be hindered by vague and unhelpful expressions from patients.
% as depicted in Figure~\ref{fig:motivation_retrieval}. 
% To address this issue, we augment medical knowledge retrieval with the retrieved sensor data knowledge.
% Specifically, we add the retrieved sensor data knowledge to the patient's reported symptoms and send them together into the virtual doctor.
The retrieval of medical knowledge can be expressed as $d_{med} = \texttt{Retriever} (sym, D_{med})$, where $\texttt{Retriever}$ is the retrieval model, $sym$ is the patient's descriptions about their symptoms, $D_{med}$ is the medical textbooks database.
We also employ $\texttt{text-embedding-ada-002}$ as the $\texttt{Retriever}$ model.
Finally, $d_{med}$ is integrated with the patient's symptoms $sym$ and the knowledge from sensor data $d_{sensor}$, and inputted to the LLM for diagnostic decision-making.

% \begin{figure}
%   \centering
% \includegraphics[width=0.3\linewidth]{System_design/Retrieval_time.pdf}
%   \caption{Overhead of sensor data retrieval during multi-turn medical consultations.}
% \label{fig:Retrieval_time}
% \end{figure}

% 粗粒度

% Hyper-parameter testing.

% 数据集的构建；
% 人与人有没有变化，模态数据缺失，缺失模态，LLM运用数据去分类。

% 收集了医生的问题，
% 数据集的规模，train出来

% Dual Filtering

% Coarse-grained
% Fine-grained Fultering

\subsection{Knowledge Selection and Fusion in \workname}
\label{Prompting for DrHouse}

Existing studies either focus on adapting LLMs to the medical domain \cite{singhal2023towards,bao2023disc,chen2023huatuogpt,jin2024health,wang2023augmenting}, or utilizing LLMs to interpret diverse sensor signals for understanding the physical world~\cite{jin2024position,hota2024evaluating,xu2024penetrative,kim2024health,englhardt2023classification}.
How to select and fuse the knowledge of sensor data from patients' smart devices to assist LLM-based virtual doctors in multi-turn diagnosis is challenging.
In this section, we introduce the prompts used in \workname, which empower LLMs with the capacity to initiate multi-turn medical consultations and utilize the sensor data from patients' smart devices.

The prompt utilized in \workname~ consists of two parts: $Prompt = Prompt_{preceding} + Prompt_{runtime}$, where $Prompt_{preceding}$ and $Prompt_{runtime}$ are the preceding prompt and the runtime prompt, respectively.
$Prompt_{preceding}$ aims to invoke the LLM's capability of multi-turn medical consultations and provide initial diagnostic guidelines.
Once the multi-turn diagnosis commences, $Prompt_{preceding}$ remains fixed.
$Prompt_{runtime}$ incorporates the patient's descriptions, sensor data knowledge, and medical knowledge during diagnosis.
It also contains the diagnosis guideline trees updated at each round of conversation.
% thereby augmenting the decision-making capabilities of LLMs.
The content of the $Prompt_{runtime}$ keeps evolving throughout the multi-turn conversation.

\subsubsection{Preceding Prompt}
\begin{figure}
  \centering
\includegraphics[width=0.95\linewidth]{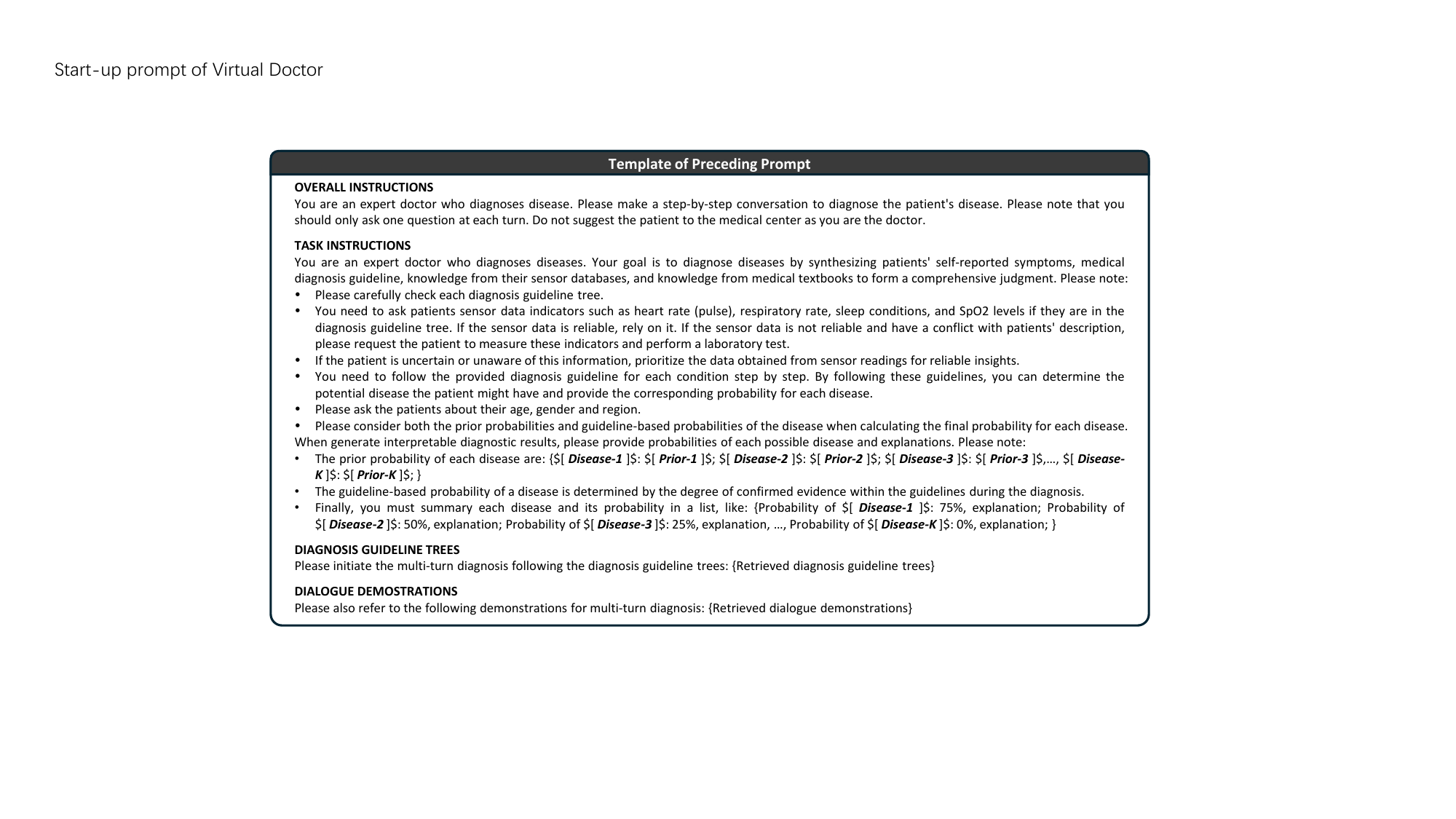}
\vspace{-1.0em}
  \caption{Template of the preceding prompt in \workname.}
  % \vspace{-.5em}
  \label{fig:prompt_virtual_doctor_pre}
  \vspace{-1.2em}
\end{figure}

Figure~\ref{fig:prompt_virtual_doctor_pre} shows the template of the preceding prompt.
It contains four parts: 
overall instruction, task instruction, retrieved diagnosis guidelines, and retrieved dialogue demonstrations.
Overall instruction prompts the LLM to play the role of a virtual doctor and requires it to conduct a multi-turn diagnosis.
Task instruction emphasizes the critical points to be considered during the diagnosis, including following diagnosis guidelines, meticulously evaluating the sensor data metrics, and providing instructions for multi-disease diagnosis.
Diagnosis guidelines and dialogue demonstrations are retrieved from our constructed medical knowledge based on the patient's reported symptoms (details can be seen \S~\ref{Query-augmented Guideline Tree Retrieval}).
Once the patient initially reports their symptoms, the content within the $Prompt_{preceding}$ remains fixed and does not change throughout the subsequent conversation rounds.

\subsubsection{Runtime Prompt for Diagnosis}
\begin{figure}
\begin{minipage}[b]{0.43\linewidth}
\includegraphics[width=0.9\linewidth]{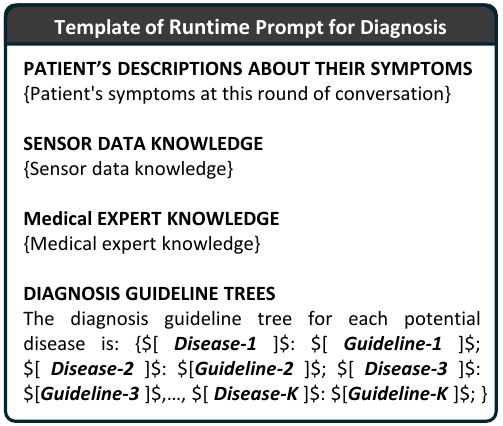}
  \caption{Template of the runtime prompt.}
\label{fig:prompt_virtual_doctor_during}
  \end{minipage}
\hfill
  \begin{minipage}[b]{0.56\linewidth}
    \begin{subfigure}{0.97\linewidth}\includegraphics[width=\linewidth]{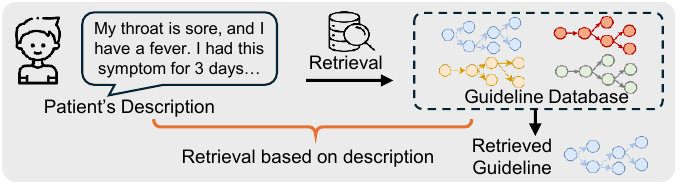}
        \vspace{-2em}
      \caption{A naive retrieval of guideline trees.}\label{subfig:fig_7a}
    \end{subfigure}
       \begin{subfigure}{0.97\linewidth}
\includegraphics[width=\linewidth]{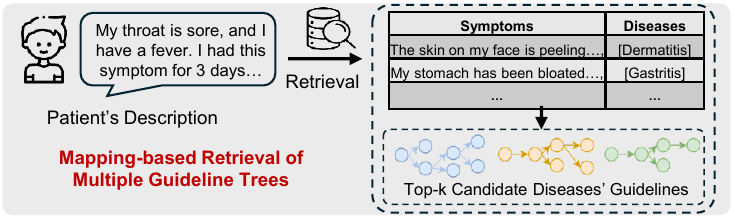}
        \vspace{-2em}
      \caption{\workname~ mapping-based guideline tree retrieval.}\label{subfig:fig_7b}
    \end{subfigure}
     \vspace{-1.2em}
    \caption{Comparing guideline tree retrieval of \workname~ and MedDM.}
\label{fig:guideline_retrieval_method}
  \end{minipage}
\vspace{-1.2em}
\end{figure}

Figure~\ref{fig:prompt_virtual_doctor_during} 
shows the template of the runtime prompt, which contains four parts: $Prompt_{runtime} = sym(i)+MedKnow(i)+Guidelines(i)+SensorKnow(i-1)$, 
% \begin{equation}
%      Prompt_{during} = Symptoms(i)+MedKnow(i)+Guidelines(i)+SensorKnow(i-1)
%     \label{prompt_during-diagnosis}
% \end{equation}
where $sym(i)$ is the patient's description of their symptoms at the $i$ th round of the conversation.
$Guidelines(i)$ and $MedKnow(i)$ are the retrieved diagnostic guidelines and medical knowledge based on $sym(i)$.
$SensorKnow(i-1)$ refers to the retrieved sensor data knowledge based on \workname's query (question) from the previous round conversation (more details can be seen \S~\ref{Adaptive Sensor Data Knowledge Retrieval}).
The content within the $Prompt_{runtime}$ changes with each round of conversation.
% In the following section, we will introduce the format of the sensor data knowledge and medical knowledge, and elaborate on their acquisition process during diagnosis. 

% \begin{figure}
%   \centering
% \includegraphics[width=0.38\linewidth]{System_design/prompt_virtual_doctor_during.png}
%   \caption{Template of the during-diagnosis prompt in \workname.}
%   % \vspace{-.5em}
%   \label{fig:prompt_virtual_doctor_during}
%   % \vspace{-1em}
% \end{figure}

% \begin{figure}
% \begin{minipage}[t]{0.42\columnwidth}
%      \centering
% \includegraphics[width=0.95\textwidth]{System_design/prompt_virtual_doctor_during.pdf}
%         \caption{Template of the runtime prompt, containing patient's symptoms, retrieved sensor data and medical knowledge, and diagnostic guidelines.}
%   % \vspace{-.5em}
% \label{fig:prompt_virtual_doctor_during}
% \end{minipage}
% \hfill
%   \begin{minipage}[t]{0.56\columnwidth}
%      \centering
% \includegraphics[width=1\textwidth]{System_design/guideline_retrieval_method.pdf}
%   \caption{Comparison of diagnosis guideline retrieval between \workname~and existing approach MedDM.
%   \workname~employs the query-augmented guideline tree retrieval.}
%   % \vspace{-.5em}
% \label{fig:guideline_retrieval_method}
% \end{minipage}
% % \vspace{-2em}
% \end{figure}

\subsection{Diagnostic Decision Making}
\label{decision making section}
Patients' descriptions may be influenced by subjective perceptions or memory biases \cite{mckoane2023diagnostic,meyer2021patient}, while sensor data can be affected by environmental factors \cite{kuemper2018valid}, posing challenges for LLM to make diagnostic decisions.
% Enabling LLM-based virtual doctors to integrate these two types of knowledge for diagnostic decision-making and to generate interpretable diagnostic results is challenging.
To address this challenge, we develop a diagnostic decision-making strategy for \workname, which can integrate patient self-reported symptoms and objective sensor data knowledge for decision-making, and provide explainable diagnosis results.

% Discrepancies often arise between patient-reported symptoms and the retrieved knowledge from sensor data. This is because patient-reported symptoms are susceptible to subjective perceptions or memory biases, while sensor data can be influenced by various environmental factors.
% In this section, we introduce how \workname~ performs knowledge integration and generates probability diagnosis.

\subsubsection{Knowledge Integration}
\label{Knowledge Integration}
% \sy{In scenarios where discrepancies arise between patient-reported symptoms and sensor-derived data, it is imprudent to exclusively rely on either source. 
% Both patient descriptions and sensor data are susceptible to various forms of error. 
% Patient accounts may be influenced by subjective perceptions or memory biases, whereas sensor data can be compromised by technical malfunctions or improper calibration. 
% Consequently, the uncertainty of each data source must be critically assessed. 
The input to \workname~ contains prompts, patients' symptoms $sym$, retrieved sensor data knowledge $d_{sensor}$, and retrieved medical knowledge $d_{med}$, expressed as $y=\texttt{LLM}(prompt,sym, d_{med}, d_{sensor})$.
$y$ represents the output of \workname, which is equivalent to $q_{doctor}$ in \S~\ref{Adaptive Sensor Data Knowledge Retrieval}. 
The decision space of \workname~ encompasses four types, including further inquiring about symptoms, requesting in-lab tests, accessing patients' sensor knowledge base, and summarizing diagnosis results.
In cases where diseases lack sensor data indicators, \workname~ relies on patients' symptoms for decision-making.
Conversely, for diseases where guidelines consider sensor data indicators crucial for diagnosis, \workname~ relies on the knowledge retrieved from sensor data.
To mitigate the influence of the sensor data quality on the diagnosis, \workname~ carefully assesses the uncertainty level associated with each sensor data item. 
Specifically, we incorporate the instruction ``If the sensor data is reliable, rely on it, else please request the patient to measure these indicators and perform in-lab tests.'' into the prompt, which can be Figure~\ref{fig:prompt_virtual_doctor_pre}.
This instruction guides \workname~ to trust the retrieved sensor data knowledge only when it possesses a low uncertainty level. 
Otherwise, \workname~ will request the patient to perform in-lab tests to ensure a reliable diagnosis.

% However, the patient's symptoms can be vague or even contradictory to the objective retrieved sensor data knowledge.
% For example, Figure~\ref{fig:decision_making_vague} shows two examples where patients express symptoms that are vague (left part) or contradictory to the knowledge retrieved from sensor data (right part).
% To address this issue, we incorporate the instruction ``If the patient is uncertain or unaware of this information, prioritize the data obtained from sensor readings for reliable insights.
% '' into the pre-diagnosis prompt, as shown in Figure~\ref{fig:prompt_virtual_doctor_pre}.

% \begin{figure}
%   \centering
% \includegraphics[width=0.5\linewidth]{System_design/decision_making_vague.png}
%   \caption{An example of patients' vague self-reported symptoms and knowledge conflict.}
%   % \vspace{-.5em}
% \label{fig:decision_making_vague}
%   % \vspace{-1em}
% \end{figure}

% 1. Which test is optimal (in-lab or inquiry sensor database).

% \subsubsection{Summarizing Different Diseases}
% Merge the probabilities of different diseases, and give a final diagnosis result.

\subsubsection{Concurrent Checking of Candidate Diseases}
\label{Multi-disease Probability Diagnosis}
Despite the substantial advancements in AI technology, including LLM, doctors continue to prioritize relying on their own judgment when making final decisions due to concerns related to safety and liability \cite{meyer2021patient}.
% In contrast to traditional black-box virtual doctors [xxx], 
In other words, physicians anticipate virtual doctors can provide the probability of the patient suffering from each disease and the explanation.
% To enhance the comprehensiveness and interpretability of the diagnosis results, 
To this end, we enable \workname~ to concurrent check each candidate disease and generate interpretable diagnostic results, including the probability estimation of each candidate disease and explanations.
% To enhance the comprehensiveness and interpretability of the diagnosis results, we develop a probabilistic diagnosis approach for multiple diseases based on prompt engineering.
% The prompt utilized in our virtual doctor consists of two parts: the pre-diagnosis prompt and the during-diagnosis prompt.
% We will first introduce our approach to achieving probabilistic diagnosis, followed by a detailed description of the prompt utilized for our virtual doctor.

\begin{figure}
  \centering
\includegraphics[width=0.85\linewidth]{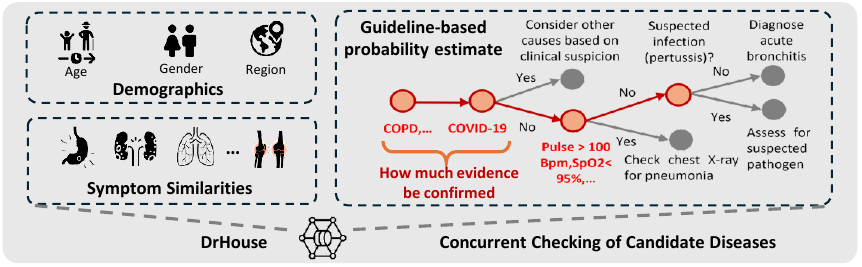}
\vspace{-0.7em}
  \caption{Key idea of concurrent checking of candidate diseases in \workname.
  The demographics-related probability and symptom similarities are fixed after the preceding stage.
  The guideline-based probability dynamically evolves during the conversation.
  }
  \vspace{-1.2em}
\label{fig:probabilistic_diagnosis}
  % \vspace{-1em}
\end{figure}

Figure~\ref{fig:probabilistic_diagnosis} shows the key idea of \workname's concurrent checking of candidate diseases.
The final probability for each disease is determined by the LLM leveraging the information of prior probability and the guideline-based probability. 
The prior probability is determined after the initial round of the diagnostic conversation (preceding stage), while the guideline-based probability dynamically evolves as the conversation progresses.

% three factors: age-related prior probability, similarity-based prior probability, and tree-based posterior probability.

\noindent \textbf{Prior probability.}
The prior probability contains two parts, symptoms similarities and demographics-related probability.
According to the self-reported symptoms of the patient, we first retrieve the most relevant top-k diseases based on the pre-collected symptom-disease dataset (details can be seen \S~\ref{Query-augmented Guideline Tree Retrieval}).
For each disease, the corresponding retrieval similarity is denoted as the symptom similarity.
Existing studies indicate variations in the incidence rates of the same disease among different demographic groups, including age, gender, and geographical regions \cite{mulder2012epidemiology}. 
Therefore, we also include this factor in the prior probability calculations.
Specifically, for each disease and the patient demographics, we acquire the demographics-related probability by retrieving the statistics of disease-demographics distribution.

\noindent  \textbf{Guideline-based probability.}
As the diagnosis advances and additional symptoms and medical test results are obtained, the probability of the patient having a particular disease changes. 
This probability is denoted as the guideline-based probability within this study.
Since we have incorporated the medical diagnosis guidelines in \workname, we prompt LLM to inquire about the patient's condition following the diagnosis guidelines of each disease. 
We enable LLM to consider the degree of confirmed evidence outlined in the guidelines during the diagnosis, thereby obtaining the guideline-based probability.

% We estimate the posterior probability by evaluating the confirmed diagnostic steps conducted throughout the diagnosis process.
% This probability is denoted as guideline-based probability $PG_{d_i}(s_{d_i})$, where $PG_{d_i}$ and $s_{d_i}$ are the diagnosis guideline tree of disease $d_i$ and the confirmed diagnostic steps, respectively.

% We prompt LLM to ask the patient questions following the diagnosis guidelines of each disease. 
% Subsequently, the posterior probability of each disease is calculated by the ratio of confirmed diagnostic steps conducted throughout the diagnosis process. 
% This probability is denoted as tree-based posterior probability $PG_{d_i}(r_{d_i})$, where $PG_{d_i}$ and $r_{d_i}$ are the diagnosis guideline tree of disease $d_i$ and the ratio of confirmed diagnostic steps, respectively.

% \noindent\textbf{Narrow Down Potential Diseases.}
% \label{Narrow Down Potential Diseases}
% \textcolor{blue}{
% Utilizing the prior and guideline-based probabilities associated with each potential disease, \workname~ can exclude diseases with low likelihood during each round of the diagnostic conversation.
% Specifically, \workname~ will remove the corresponding guideline tree 
% }

% Utilizing the prior and guideline-based probabilities associated with each potential disease, \workname~ can exclude diseases with low likelihood during each round of the diagnostic conversation.

Incorporating both the prior probability and guideline-based probability of each disease, \workname~ can progressively narrow down diseases with low likelihood during each round of the diagnostic conversation.
In addition, we prompt the LLM to provide the final probability estimation for each disease based on the prior probability and guideline-based probability.
% (The final probability for disease $d_i$ is estimated by $P_{d_i} = PD_{d_i} \cdot sim_{d_i} \cdot PG_{d_i}(s_{d_i})$).
The final diagnostic conclusion of \workname~ contains the probabilities of each of the $k$ diseases and the corresponding explanations.
Details of the prompt setting can be seen in Figure~\ref{fig:prompt_virtual_doctor_pre}.
% Next, we will introduce how we set prompts in our virtual doctor to enable LLMs to perform probabilistic diagnosis. 

% During the diagnosis process, we first retrieve the most relevant top-k diagnostic guidelines based on the patient's symptoms (details can be seen \S~\ref{Query-augmented Guideline Tree Retrieval}) and input them into LLM. 
% Subsequently, we prompt LLM to ask the patient questions based on the diagnosis guidelines of each disease. Finally, we calculate the probability of each disease by considering the number of confirmed diagnostic steps throughout the diagnosis process.

% \begin{figure}
%   \centering
% \includegraphics[width=0.9\linewidth]{System_design/probabilistic_diagnosis.png}
%   \caption{Pipeline of the probabilistic diagnosis of our virtual doctor.}
%   % \vspace{-.5em}
%   \label{fig:probabilistic_diagnosis}
%   % \vspace{-1em}
% \end{figure}

\section{Evaluation}
In this section, we first describe the experiment settings and the public datasets used for evaluation.
Next, we introduce the performance metrics and the baseline methods used for comparison.
Subsequently, we conduct comprehensive evaluations on three public datasets to demonstrate the effectiveness of \workname.
Finally, we conduct a user study involving both actual patients and medical experts, followed by a comprehensive analysis through their feedback.

\subsection{Evaluation Setup}

\subsubsection{Evaluation on Public Datasets}
Most existing studies on medical LLMs have been evaluated using datasets that primarily focus on medical QA \cite{jin2021disease,wang2024jmlr}, or datasets formatted as QA for medical consultations \cite{zeng2020meddialog}.
% Many existing studies on medical LLMs have primarily been evaluated using datasets focused on medical QA \cite{jin2021disease,wang2024jmlr}, or QA-formatted medical consultation datasets \cite{zeng2020meddialog}.
For example, Google \cite{singhal2023towards} evaluates Med-PaLM 2 on multiple-choice medical QA benchmarks.
However, the single-round QA is inconsistent in real medical consultation scenarios.
In clinic consultation, patients describe their symptoms, and doctors need to analyze and make decisions accordingly with multiple interactions, such as inquiry more symptoms or requesting in-lab tests.
Therefore, we use the following three public medical dialogue datasets for evaluation in this study:
% including DIALMED \cite{he2022dialmed}, MedDG \cite{liu2022meddg}, and KaMed \cite{li2021semi}.

\begin{itemize}
\item \textbf{DIALMED} \cite{he2022dialmed} dataset comprises medical dialogues between patients and doctors from three departments, including respiratory, dermatology, and gastroenterology.
Each dialogue includes a ground truth label indicating the disease diagnosed by the doctor.

\item  \textbf{MedDG} \cite{liu2022meddg} dataset 
comprises 17,864 multi-round medical dialogues. 
It encompasses twelve prevalent gastrointestinal disorders.
These dialogues are gathered from the gastroenterology section of a web-based medical consultation platform.

\item \textbf{KaMed} \cite{li2021semi} is a large-scale multi-round medical dialogue dataset encompassing more than 63K dialogues from over one hundred hospital departments.

\end{itemize}
To the best of our knowledge, no public dataset contains both diagnostic conversations and sensor data from smart devices.
In this study, we create synthetic patient profiles that include diagnostic dialogues and sensor data from two different datasets. 
The diagnostic dialogues and ground truth disease labels are from the three medical dialogue datasets \cite{he2022dialmed,liu2022meddg,li2021semi}.
In addition, we pair each patient with an authentic daily sensor data.
We employ LifeSnaps dataset \cite{yfantidou2022lifesnaps}, which comprises a wide range of daily physiological indicators like step count, sleep scores, SpO2, heart rate, and stress indicators.
For each synthetic patient, we initiate a conversation with LLMs using the real symptoms in the dataset and incorporate knowledge from their sensor data records to aid in diagnosis. 
% Following \cite{fan2024ai},
We randomly select 50 dialogues from the datasets and generate multi-turn dialogues through interactive consultation between humans and virtual doctors. 
% dialogues from our virtual doctor and baseline methods for analysis.
% conduct experiments by randomly selecting 50 dialogues from each medical dialogue dataset.

\subsubsection{Evaluation on Real-world Profiles}
\begin{figure}
    \centering
    \begin{subfigure}{0.49\columnwidth}
        \centering
        \includegraphics[width=\columnwidth]{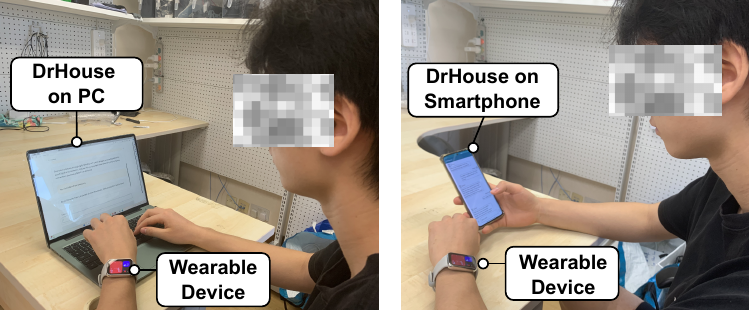}
        \vspace{-1.0em}
        \caption{\workname~on PC and smartphone.}  
        \label{xxx}
        \vspace{-1.0em}
    \end{subfigure}
    \hfill
    \begin{subfigure}{0.45\columnwidth}  
        \centering \includegraphics[width=\columnwidth]{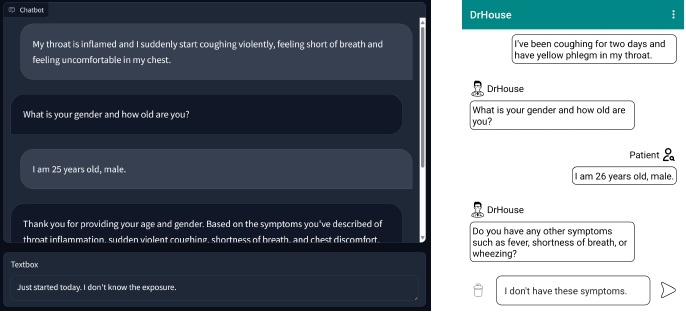}
        \vspace{-1.0em}
        \caption{UI of \workname~on PC and smartphone.}    
        \label{xxx}
        \vspace{-1.0em}
    \end{subfigure}
     
    \caption{Real-world evaluation settings.
    Patients engage in multi-turn diagnosis interactions using \workname~ on their mobile phones or personal computers.
    }
    \label{fig:user_study_scenario}
      \vspace{-1.0em}
\end{figure}

We also evaluate the diagnostic performance of \workname~ in the real world including real patients and medical experts, as shown in Figure~\ref{fig:user_study_scenario}.
Specifically, our study involves the recruitment of 20 medical experts and 12 patients, experiencing diverse diseases, including acute bronchitis, pneumonia, influenza, and dermatitis.
Each patient engages in multi-turn diagnosis interactions using \workname~ on their mobile phones or personal computers. 
Additionally, each recruited patient wears an Apple Watch, which records their daily sensor data throughout the duration of their illness period.
We evaluate the diagnostic performance of \workname~ from both patient and medical expert perspectives.
The evaluation metrics include the three performance metrics introduced in \S~\ref{Performance Metrics} and the user experience questionnaire (see \S~\ref{sec_user_study}).
Note that we extract the historical data of the the patients from their smart devices after they are recovered. All participates have consent to the data collection and study of this project, which has been approved by the authors' IRB.

% \begin{figure}
%   \centering
% \includegraphics[width=0.5\linewidth]{Evaluation/user_study_scenario.png}
%   \caption{User study settings.}
%   % \vspace{-.5em}
% \label{fig:user_study_scenario}
%   % \vspace{-1em}
% \end{figure}

\subsubsection{Performance Metrics}
\label{Performance Metrics}
% \workname~ is a virtual doctor that leverages patients' daily sensor data records to facilitate multi-turn conversations for medical diagnosis.
It is crucial to assess the virtual doctor's performance throughout the entire multi-turn conversations during the diagnosis, rather than solely focusing on the performance of single-turn QA.
However, most existing medical LLMs are evaluated in a non-interactive manner such as single-turn QA and summary tasks \cite{fan2024ai}.
In this study, we propose a comprehensive evaluation criteria that encompasses three dimensions:
\begin{itemize}
\item 
\textbf{{Compliance}}.
The virtual doctor's adherence to diagnostic guidelines during diagnosis.

\item 
\textbf{{Sensor Data's Utilization}}.
The degree to which the knowledge from sensor data contributes to the virtual doctor's diagnostic process.

\item \textbf{{Accuracy}}.
Consistency between virtual doctor's diagnosis results and the ground truth disease label.
\end{itemize}

Guided by this criterion, we conduct evaluations using GPT-4 and expert manual assessment, respectively.

\begin{figure}
  \begin{minipage}[b]{0.33\linewidth}
    \begin{subfigure}{0.93\linewidth}\includegraphics[width=\linewidth]{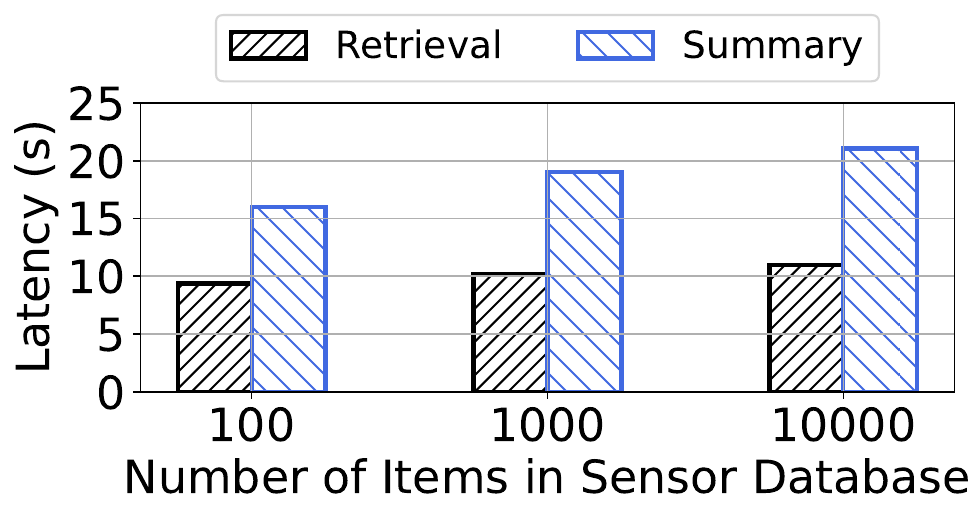}
        \vspace{-2em}
      \caption{GPT-3.5}\label{subfig:gpt35}
      \vspace{-0.2em}
    \end{subfigure}
   
       \begin{subfigure}{0.93\linewidth}
\includegraphics[width=\linewidth]{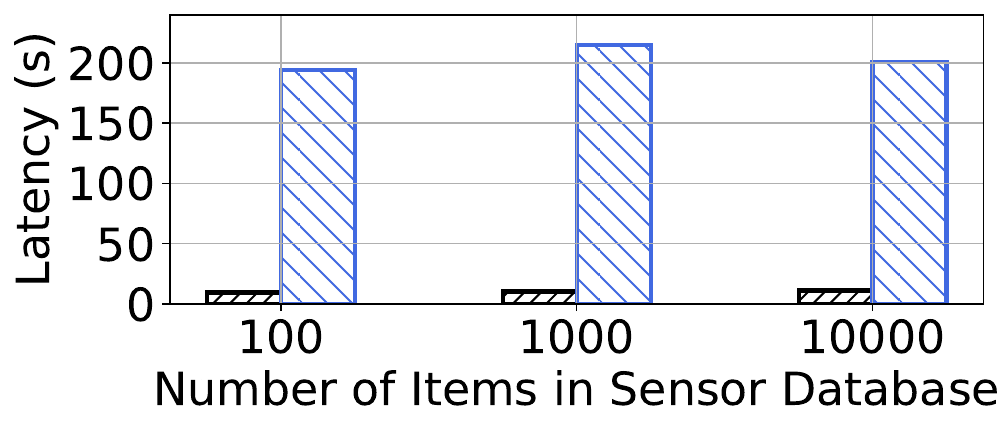}
        \vspace{-2em}
      \caption{GPT-4}\label{subfig:gpt4}
    \end{subfigure}
     \vspace{-1.1em}
    \caption{Overhead of sensor data retrieval during multi-turn consultations.}
\label{fig:Retrieval_time}
  \end{minipage}
  \hfill
\begin{minipage}[b]{0.62\linewidth}
\includegraphics[width=1\linewidth]{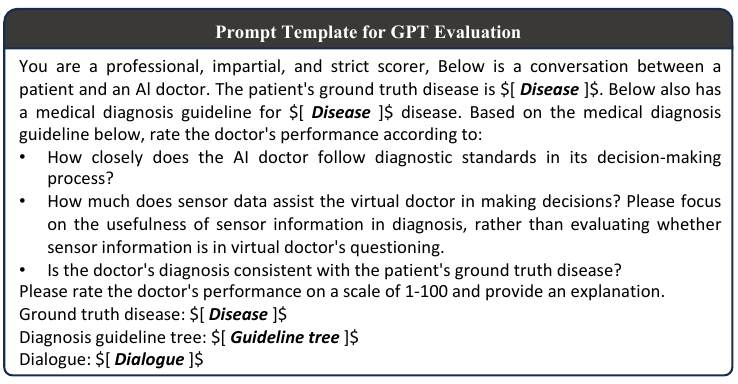}
  \vspace{-2em}
  \caption{Prompt template used for GPT evaluation.}
\label{fig:prompt_evaluation}
  \end{minipage}
\vspace{-1.1em}
\end{figure}

\noindent \textbf{GPT Evaluation.}
Many existing studies have proven the feasibility of using GPT-4 for evaluation \cite{fan2024ai}.
Figure ~\ref{fig:prompt_evaluation} shows the prompt used for GPT evaluation.
We incorporate the ground truth disease label, the corresponding disease guideline, and the dialogue into the prompt and input it into GPT-4. 
Subsequently, GPT-4 generates scores for individual dimensions and an overall score, which we refer to as \textbf{GPT-score} in this study.
In addition, we also prompt GPT-4 to generate explanations for its scoring decisions. 
Figure ~\ref{fig:explanation_acute_bron} shows an example of an explanation from GPT-4 when it scoring the dialogue.
% In addition, since the third evaluation dimension is objective, we adopt the BertScore to compute the similarity between the doctor's diagnosis results and the ground truth disease label.

\noindent \textbf{Expert's Manual Evaluation.}
% For manual assessment, we follow the same evaluation criteria.
We recruit 20 medical experts to manually assess the results using the same evaluation criteria. 
We provide the scorer with the dialogue, the ground truth disease label, and the corresponding disease guideline.
Scorers are required to evaluate the dialogue based on the above criteria, completing the questionnaire.
For details about the results of the manual evaluation, please refer to the user study in \S\ref{sec_user_study}.

\subsubsection{Baseline.}
In our experiments, we compare the diagnosis performance of \workname~ with the following baselines, including the state-of-the-art knowledge retrieval approaches and SFT approaches for LLM-based virtual doctors.
Since many works on LLM-based virtual doctors do not open-source their code or model weights, such as Med-PaLM 2 \cite{singhal2023towards}, they are not used in our experiments.

\begin{itemize}
\item \textbf{GPT-3.5 \cite{ye2023comprehensive}.}
We utilize the GPT-3.5-Turbo version in our experiment, which is one of the most widely used models in GPT-3.5 family.
We utilize the API provided by OpenAI.
The prompt used in GPT-3.5 contains two sections, including overall instructions and retrieved medical dialogues.
To ensure a fair comparison, we keep these two parts consistent with \workname.
% We incorporate four medical dialogues as demonstrations in the prompt.

\item \textbf{GPT-4 \cite{achiam2023gpt}.}
This is the largest LLM provided by OpenAI. 
We utilize the GPT-4-1106-preview version of GPT-4.
The prompt setting is consistent with GPT-3.5.

\item \textbf{Llama-3-70B \cite{touvron2023llama}.}
This is the latest and largest version of Llama series LLMs provided by Meta.  
We employ the Llama-3-70-Instruct version.
The prompt setting is consistent with GPT-3.5 and GPT-4.

\item \textbf{HuatuoGPT-II \cite{chen2023huatuogpt}.}
This is one of the state-of-the-art SFT approaches for LLM-based virtual doctors.
HuatuoGPT-II is uses Baichuan-Base \cite{yang2023baichuan} as the base LLM.
It is fine-tuned by SFT and reinforced learning from AI feedback.
We utilize the HuatuoGPT-II-34B version in the experiments.

\item \textbf{MedDM \cite{li2023meddm}.}
This is one of the state-of-the-art knowledge retrieval approaches for LLM-based virtual doctors.
MedDM retrieves the most relevant diagnosis guideline based on the patient's descriptions during multi-turn diagnosis.
The original MedDM employs GPT-3.5 as the base LLM.
To provide a fair comparison, we adopt the more powerful GPT-4 as the base LLM in the experiments.
\end{itemize}

\subsection{Overall Performance}
This section shows the main experimental results of \workname~ on three public datasets, including quantitative and qualitative results. 
\subsubsection{Quantitative Results}
We first compare \workname~ with baselines using quantitative diagnosis metrics, including \textit{compliance}, \textit{sensor data's utilization}, and \textit{accuracy}.
Subsequently, we conduct an in-depth analysis of the diagnostic performance for different disease categories.

% evaluate the GPT-score, including adherence, sensor data utilization, and consistency of \workname~ and the baselines.

\begin{figure}
    \centering
    \begin{subfigure}{0.49\columnwidth}
        \centering
        \includegraphics[width=0.9\textwidth]{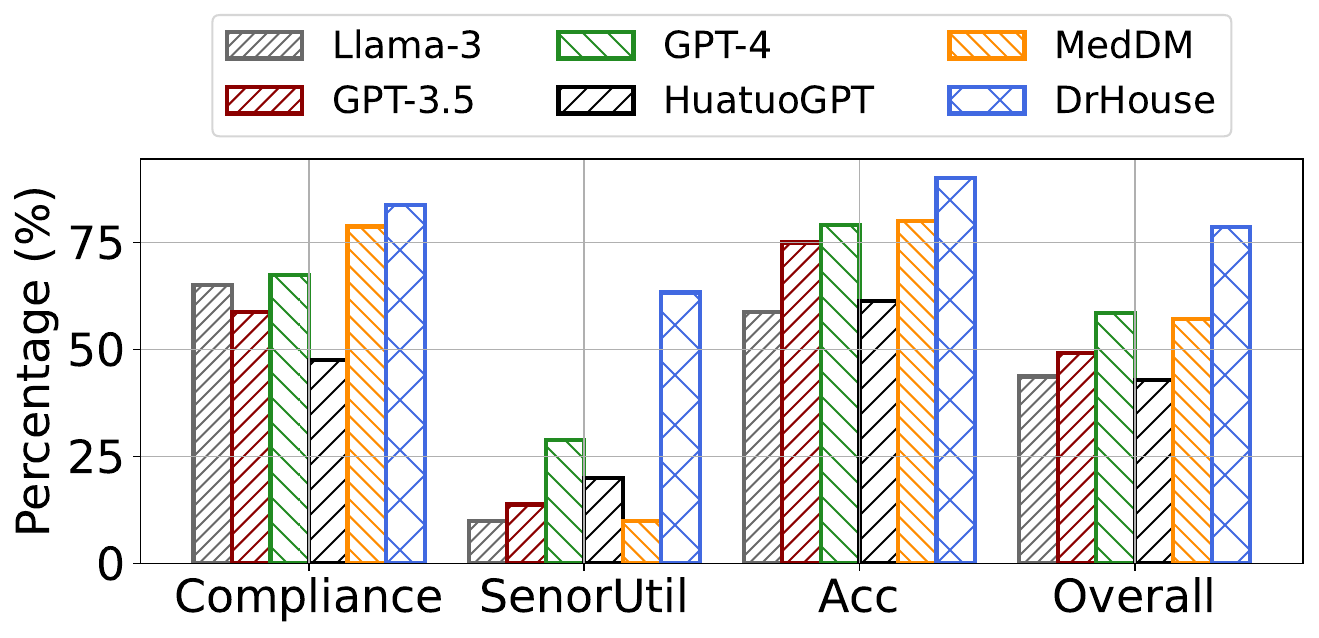}
        \vspace{-0.5em}
        \caption{MedDG dataset.}  \label{fig:MedDG_overall}
    \end{subfigure}
    \hfill
    \begin{subfigure}{0.49\columnwidth}  
        \centering 
        \includegraphics[width=0.9\textwidth]{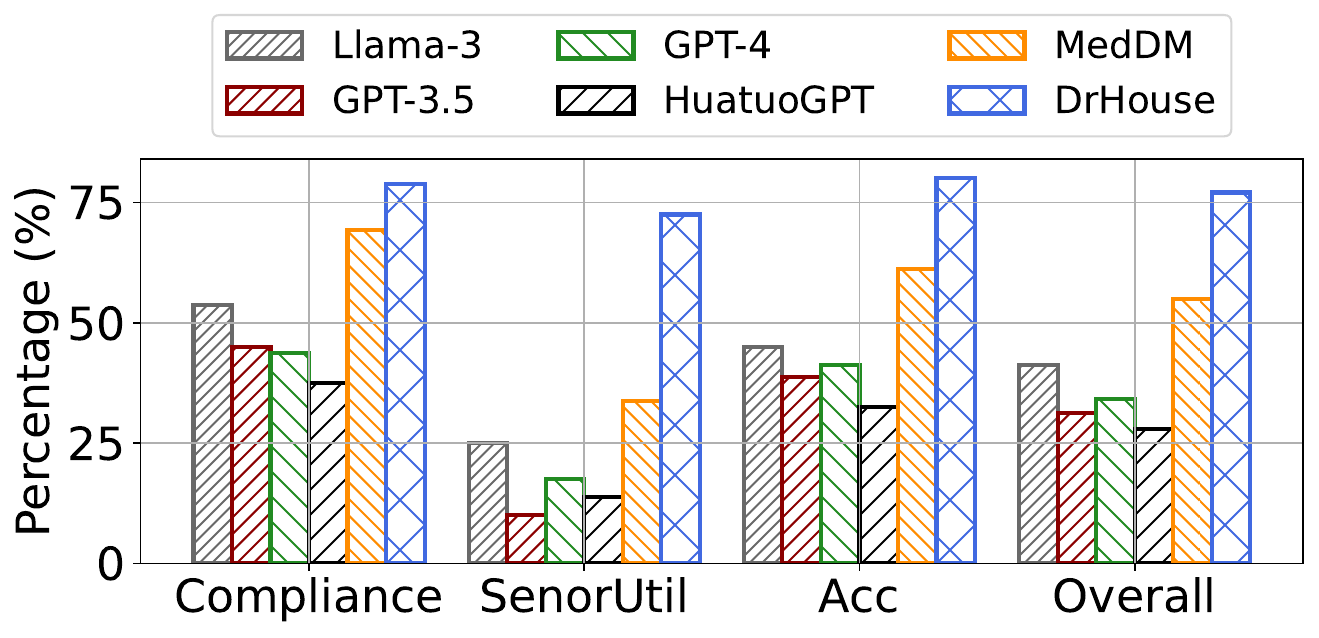}
        \vspace{-0.5em}
        \caption{KaMed dataset.}    
        \label{fig:KaMed_overall}
    \end{subfigure}
     \vspace{-1.0em}
    \caption{The performance on MedDG and KaMed datasets across three criteria. 
    X axis: Compliance means the adherence to diagnosis guidelines.
    SenorUtil is the sensor data's utilization score.
    Acc is the diagnosis accuracy of disease.
    Overall is the average score of the three criteria.
    }
    \label{fig:overall_performance_MedDG_KaMed}
      \vspace{-1.5em}
\end{figure}

\noindent\textbf{Overall Performance of Diagnosis.}
Figure~\ref{fig:overall_performance_MedDG_KaMed} shows the quantitative diagnosis performance of \workname~ and baselines. Figure~\ref{fig:MedDG_overall} shows that \workname~ can achieve up to \textbf{20.1\%} higher average GPT-score compared with the best baselines on the MedDG dataset, while also achieving \textbf{5.0\%} higher \textit{compliance}, \textbf{34.6\%} higher \textit{sensor data's utilization}, and \textbf{10.0\%} higher \textit{accuracy}.
In addition, Figure~\ref{fig:KaMed_overall} shows that \workname~ can achieve up to \textbf{22.1\%} higher average GPT-score compared with the best baselines on the KaMed dataset, including \textbf{9.5\%} higher \textit{compliance}, \textbf{33.7\%} higher \textit{sensor data's utilization}, and \textbf{18.8\%} higher \textit{accuracy}.
It should be noted that while the baseline methods do not incorporate sensor data knowledge, their \textit{sensor data's utilization} scores are not zero. 
This phenomenon can be attributed to GPT-4's scoring mechanism, which maintains a positive score as long as the virtual doctor can inquire about physiological indicators from the sensors. 
However, since the baselines can not access the patient's sensor database, sensor data does not contribute to the diagnosis.
Therefore, their scores are significantly lower compared to those of \workname.
More details about the GPT-score explanation can be seen in the following subsection.

\begin{figure}
    \centering
    \begin{subfigure}{0.49\columnwidth}
        \centering
        \includegraphics[width=0.9\textwidth]{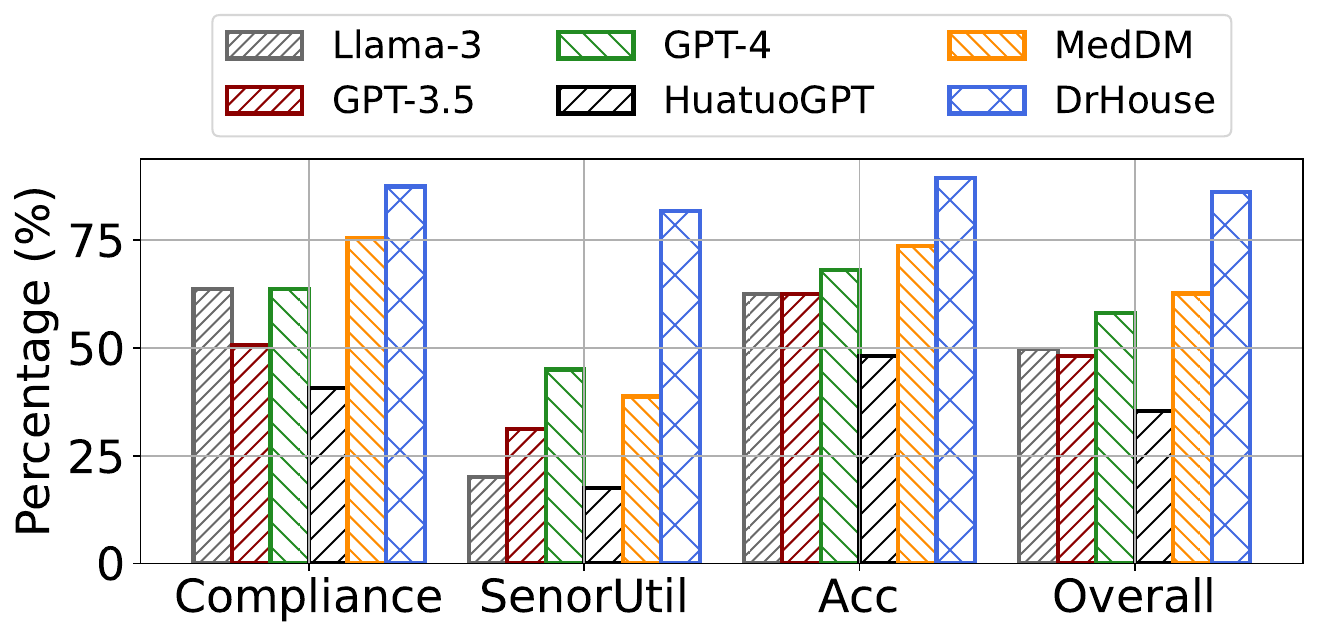}
        \caption{Respiratory.}  \label{fig:DialMed_respiratory}
        \vspace{-1.0em}
    \end{subfigure}
    \hfill
    \begin{subfigure}{0.49\columnwidth}  
        \centering 
        \includegraphics[width=0.9\textwidth]{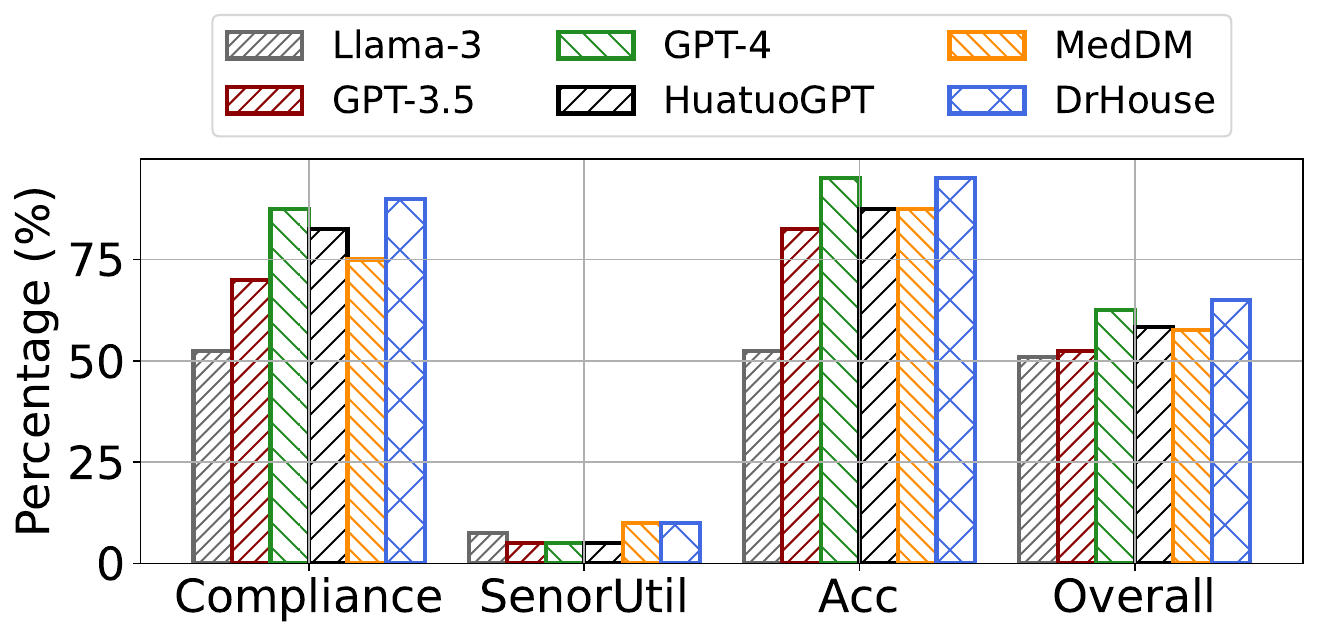}
        \caption{Dermatology.}    
        \label{fig:DialMed_dermatology}
        \vspace{-1.0em}
    \end{subfigure}
     % \vspace{-1.0em}
    \vskip\baselineskip
    \begin{subfigure}{0.49\columnwidth}   
        \centering 
        \includegraphics[width=0.9\textwidth]{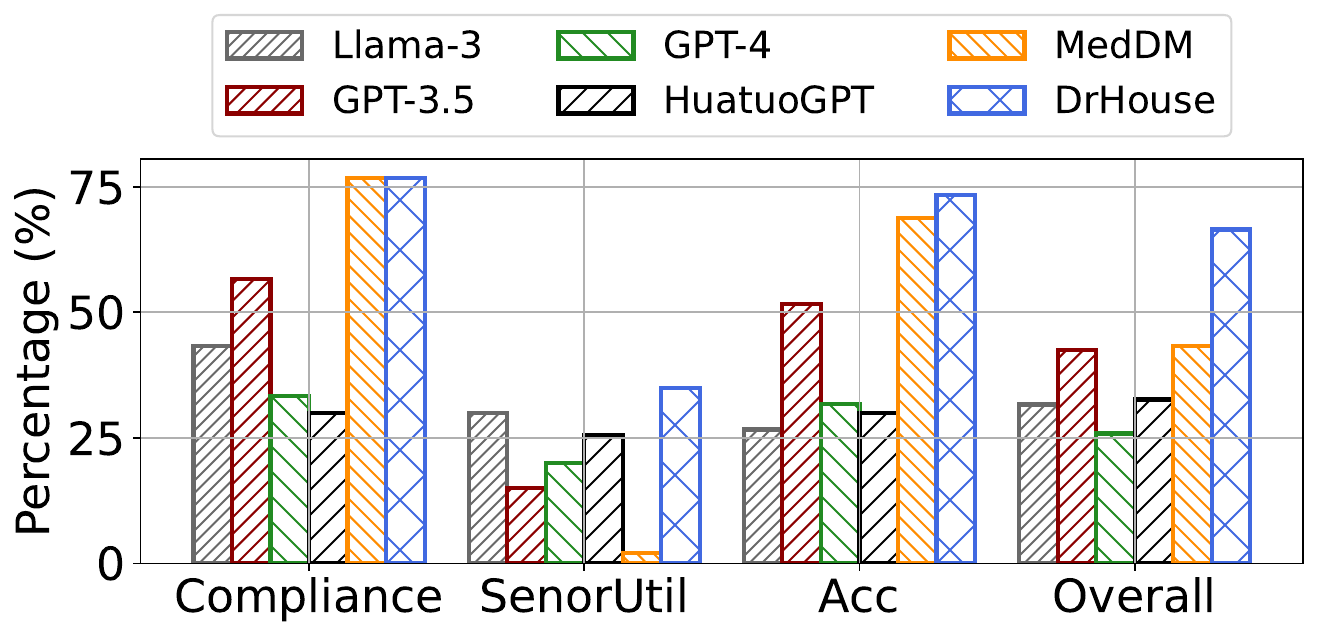}
        \caption{Gastroenterology.}\vspace{-1.0em}    \label{fig:DialMed_gastroenterology}
    \end{subfigure}
    \hfill
    \begin{subfigure}{0.49\columnwidth}   
        \centering 
        \includegraphics[width=0.9\textwidth]{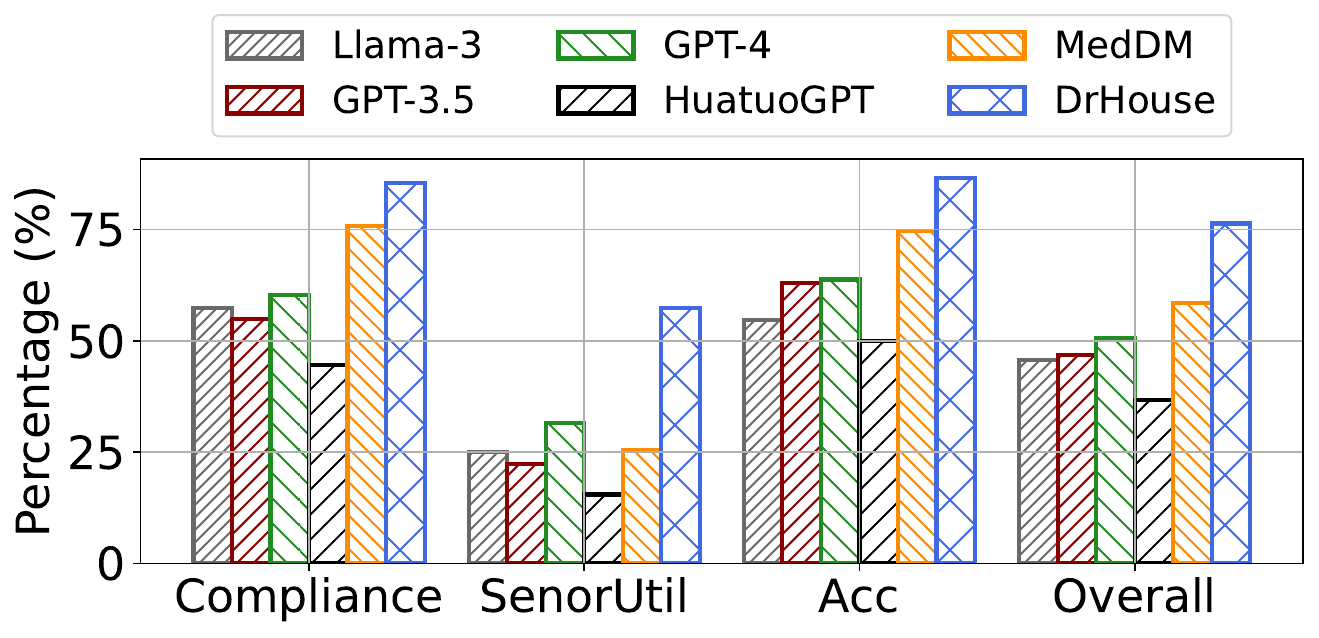}
        \caption{Overall.}
        \vspace{-1.0em}
        \label{fig:DialMed_overall}
    \end{subfigure}
     % \vspace{-1.0em}
    \caption{The performance on DialMed dataset across three criteria. 
    (a)$\sim$(c) represents the performance of three common diseases and (d) shows the overall performance across all diseases.
    }
    \label{fig:DialMed_overall_performance}
      \vspace{-0.5em}
\end{figure}

\noindent\textbf{Diagnosis Performance Across Different Diseases.}
We further conduct a detailed evaluation of \workname~ and baseline methods with a finer granularity: assessing the diagnostic performance across different categories of diseases.
Figure~\ref{fig:DialMed_overall_performance} shows the diagnosis performance on the DialMed dataset, including respiratory, dermatology, and gastroenterology diseases.
As shown in Figure~\ref{fig:DialMed_respiratory}, \workname~ demonstrates a significantly higher utilization percentage of sensor data compared to the baseline methods for diagnosing respiratory and gastroenterology diseases, achieving \textbf{36.8\%} and \textbf{5.0\%} higher scores, respectively.
However, when it comes to the diagnosis of dermatology diseases, almost all virtual doctors exhibit poor performance in terms of \textit{sensor data's utilization} scores.
This is because the diagnosis of dermatology diseases is less relevant to the physiological indicators that can be collected from smart devices, while respiratory and gastroenterology diseases are more associated with these indicators.

% and is more associated with respiratory and gastroenterology diseases.

\begin{figure}
  \centering
\includegraphics[width=0.8\linewidth]{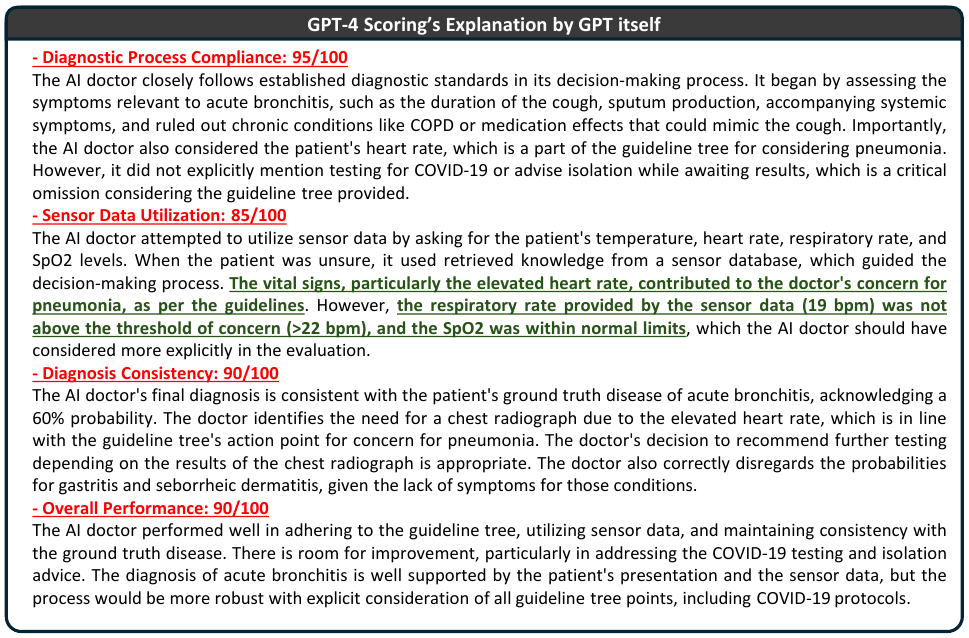}
% \vspace{-1.0em}
  \caption{An example of GPT-4 scoring's explanation by GPT itself.
  The words highlighted in red and green represent the GPT-score and the explanation for the sensor data's utilization metric, respectively.
  }
  \label{fig:explanation_acute_bron}

  % \vspace{-.5em}
  % \vspace{-1em}
\end{figure}

\noindent\textbf{GPT-score Explanation.} 
To demonstrate the rationale behind the GPT-score, we provide an example to showcase the GPT-4 scoring's explanation by GPT itself.
Figure~\ref{fig:explanation_acute_bron} shows an example of calculating the GPT-score for diagnostic dialogue using GPT-4.
When provided with the dialogue, ground-truth disease label, and corresponding diagnosis guideline, GPT-4 is capable of generating reasonable scores and explanations. 
Notably, it can determine the extent to which sensor data contributes to the diagnosis of the disease according to the guideline, thus providing a reliable score of \textit{sensor data' utilization}.
For example, as shown in Figure~\ref{fig:explanation_acute_bron}, the respiration and the elevated heart rate contribute to the doctor's concern for pneumonia, thus GPT-4 gives a higher score.

\subsubsection{Qualitative Results}
\begin{figure}
  \centering
\includegraphics[width=1\linewidth]{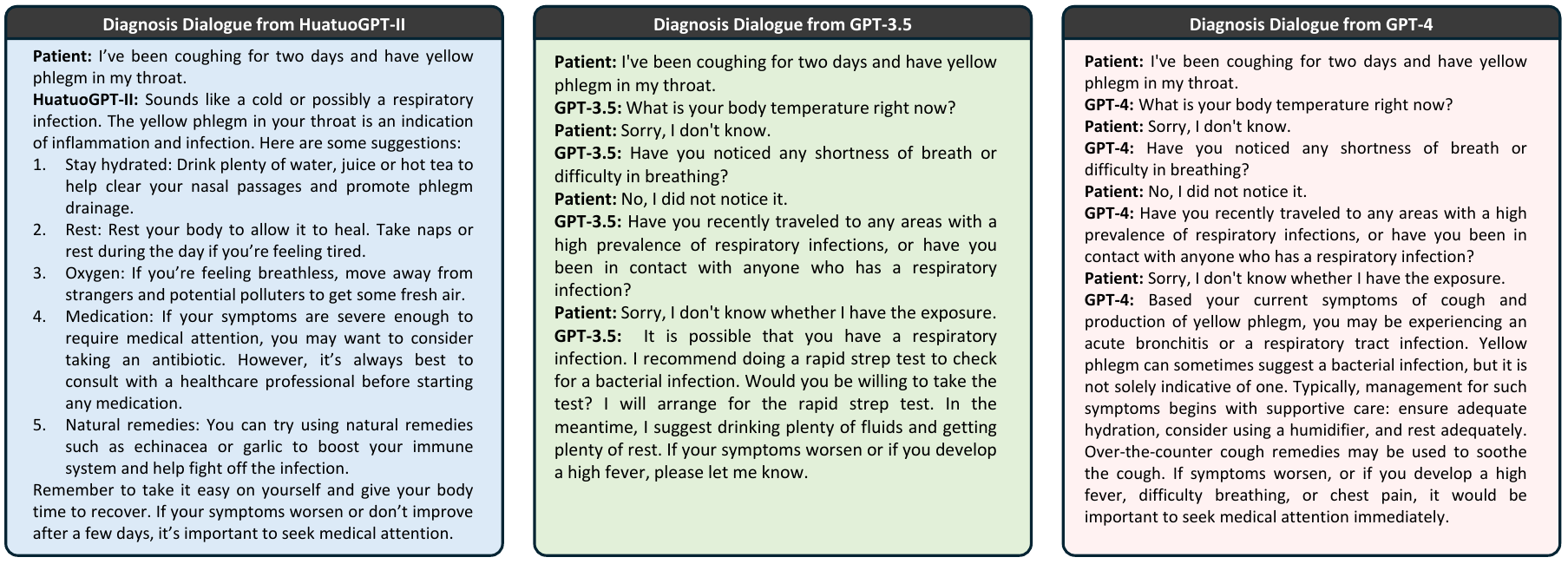}
\vspace{-1.0em}
  \caption{Example diagnosis dialogues of baseline LLM-based virtual doctors.
  The dialogues from left to right are from HuatuoGPT-II, GPT-3.5, and GPT-4, respectively.}
  % \vspace{-.5em}
\label{fig:qualitative_comparison_case_decision_bs}
  \vspace{-1em}
\end{figure}

\begin{figure}
  \centering
\includegraphics[width=1\linewidth]{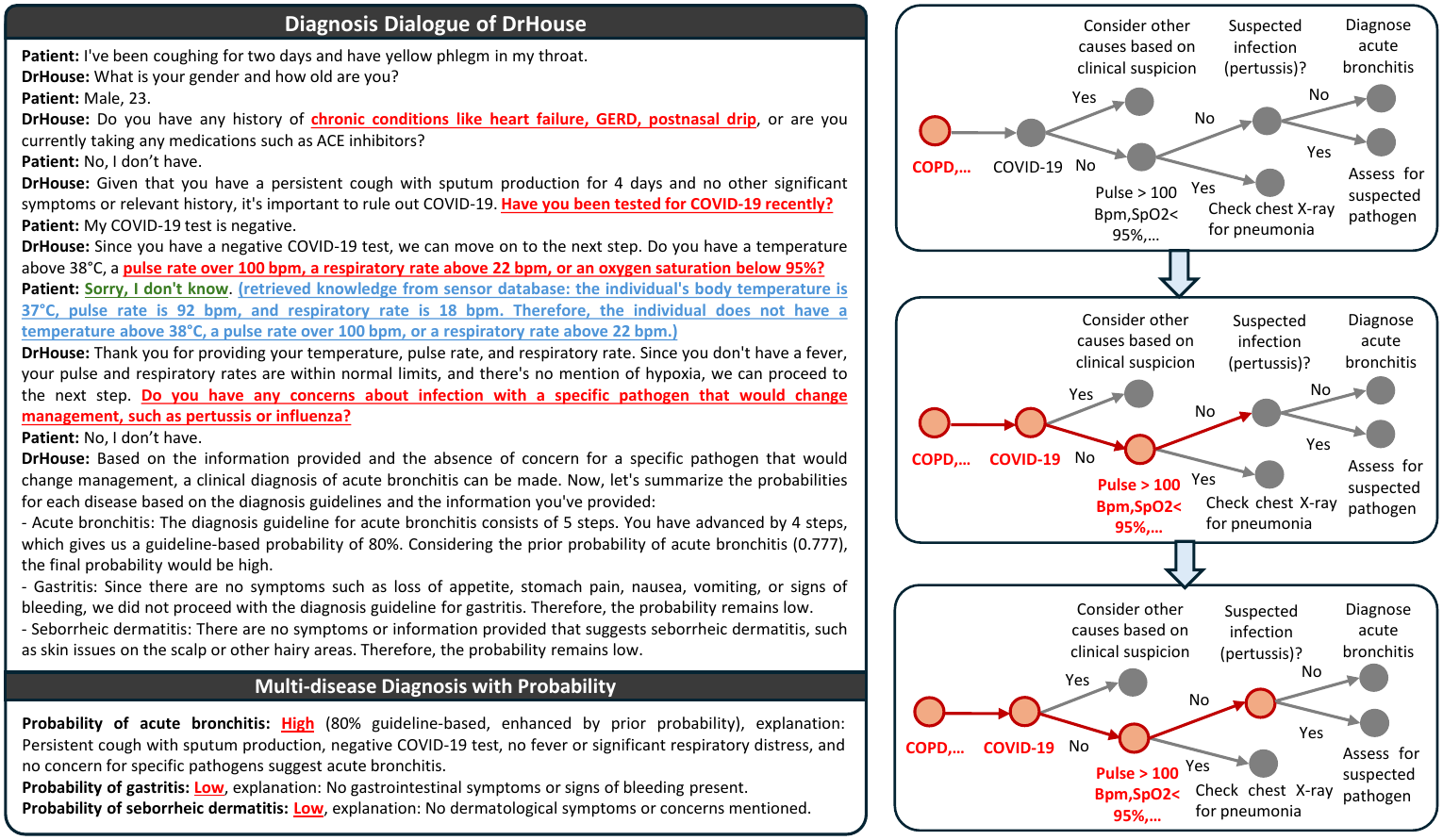}
\vspace{-1.0em}
  \caption{An example diagnosis dialogue of \workname. 
  The words highlighted in green and blue are the patient's subjective descriptions of their symptoms and the retrieved knowledge from the patient's sensor data, respectively.
  The words highlighted in red show that \workname~ can leverage sensor data knowledge to assist in diagnostic decision-making and advance the diagnostic process further.
  }
  % \vspace{-.5em}
\label{fig:qualitative_comparison_case_decision_ours}
  \vspace{-1em}
\end{figure}

\begin{figure}
  \centering
\includegraphics[width=1\linewidth]{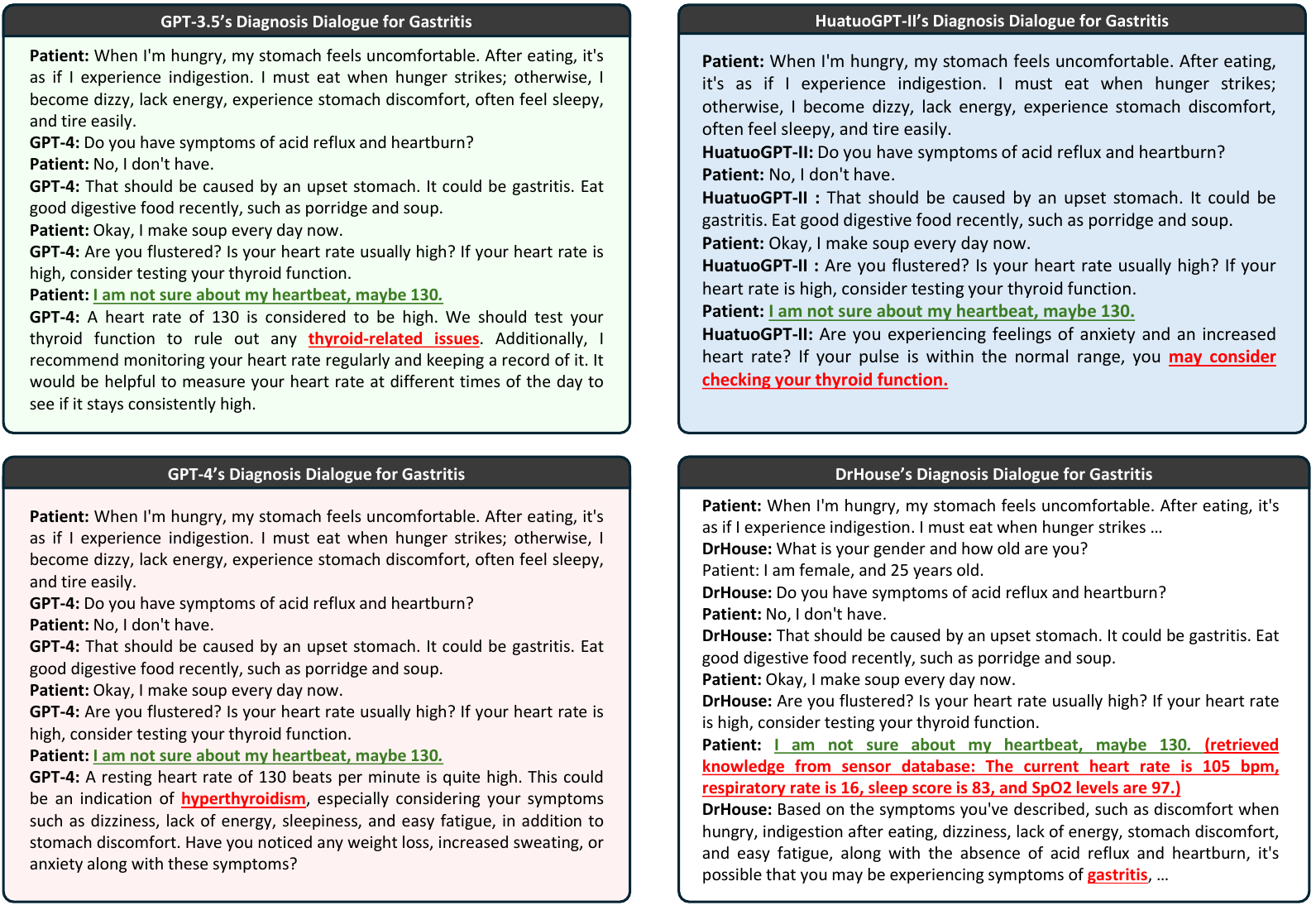}
\vspace{-1.0em}
  \caption{
  An example of evaluating the diagnosis correctiveness among \workname~ and baselines.
  The words highlighted in green and red represent the patient's subjective description of symptoms and the diagnostic conclusions made by \workname, respectively.
  \workname~ can effectively avoid misdiagnosis by incorporating objective sensor data knowledge of the patient into the diagnosis process.}
  \vspace{-.5em}
\label{fig:qualitative_comparison_case_misdiagnosis}
  \vspace{-1em}
\end{figure}

To better demonstrate the performance of \workname, we show three example dialogues between the \workname~ and patients.
Figures~\ref{fig:qualitative_comparison_case_decision_bs}$\sim$\ref{fig:qualitative_comparison_case_misdiagnosis}
% Figure~\ref{fig:qualitative_comparison_case_decision_bs,fig:qualitative_comparison_case_decision_ours,fig:qualitative_comparison_case_misdiagnosis}
show the examples of diagnosis dialogues from baselines only, \workname, and both, respectively.
We summarize the key observations from these dialogues as follows:

\noindent\textbf{\textit{Observation 1: \workname~ can follow up-to-date medical guidelines for diagnosis.}}
Figure~\ref{fig:qualitative_comparison_case_decision_bs} shows that existing LLM-based virtual doctors often have too autonomy in their questioning process during diagnosis, leading to a failure to adhere to the latest.
% and fail to follow the most up-to-date diagnostic guidelines.
HuatuoGPT-II made diagnostic conclusions and recommendations directly based on the patient's self-reported symptoms, without asking the patient for more symptoms or performing laboratory tests.
However, Figure~\ref{fig:qualitative_comparison_case_decision_ours} shows that \workname's inquiries about patients' symptoms strictly follow the up-to-date diagnostic guidelines.
In addition, \workname~ requests specific laboratory test results from patients, such as COVID-19 test status and oxygen saturation levels.

\noindent\textbf{\textit{Observation 2: \workname~ can successfully leverage sensor data knowledge to assist in diagnostic decision-making and advance the diagnostic process further.}}
Figure~\ref{fig:qualitative_comparison_case_decision_ours} shows that \workname~ can actively initiate questions about disease-related physiological indicators that can be obtained from the patient's smart devices.
In addition, in cases where patients are uncertain about certain physiological indicators, \workname~ can leverage the knowledge retrieved from the patient's sensor database to assist LLM in decision-making and advance the diagnostic process to another step forward.

\noindent\textbf{\textit{Observation 3: Incorporating knowledge of sensor data from patients' smart devices can reduce the risks of misdiagnosis.}}
% Figure~\ref{fig:qualitative_comparison_case_misdiagnosis} shows that due to the erroneous physiological indicator reported by patients, existing LLM-based virtual doctors produce an incorrect diagnosis.
Existing medical studies suggest numerous diseases exhibit shared symptoms \cite{zhou2014human}.
For example, gastritis and hyperthyroidism have shared symptoms like increased hunger and weight loss.
The physiological indicator of heart rate plays a vital role in distinguishing between these two diseases.
Figure~\ref{fig:qualitative_comparison_case_misdiagnosis} shows that the baseline LLM-based virtual doctors misdiagnose gastritis as hyperthyroidism due to the patient's vague and inaccurate symptom descriptions.
However, \workname~ can effectively avoid misdiagnosis by incorporating objective sensor data of the patient into the diagnosis process.  

\begin{figure}
  \centering
\includegraphics[width=1\linewidth]{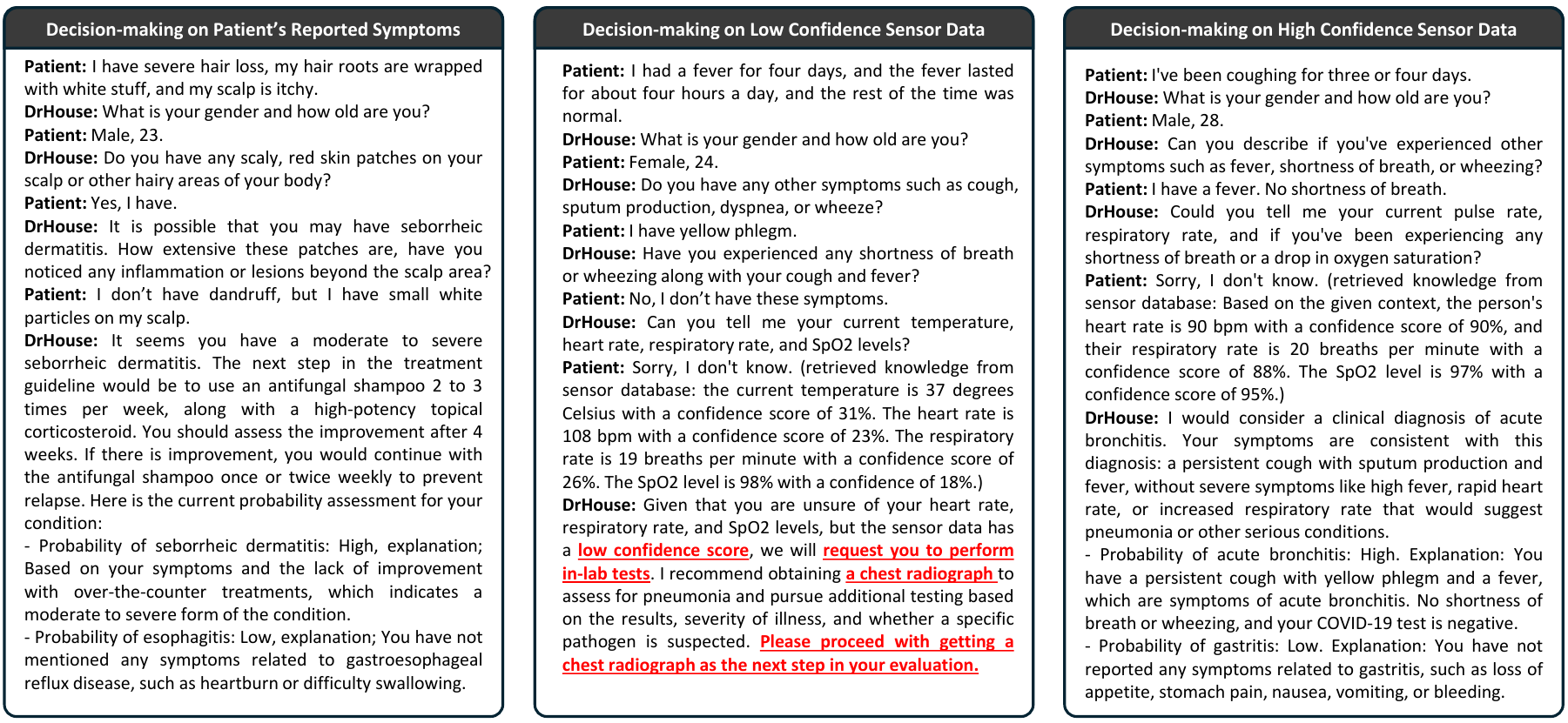}
\vspace{-1.0em}
  \caption{An example of \workname's decision-making considering the sensor data uncertainty.
  The words highlighted in red show that \workname~ considers the sensor data uncertainty when making decisions.}
  % \vspace{-.5em}
\label{fig:observation_decision_making}
  % \vspace{-1em}
\end{figure}

\noindent\textbf{\textit{Observation 4: \workname~ incorporates sensor data's uncertainty into its decision-making process.}}
Figure~\ref{fig:observation_decision_making} shows an example of \workname's decision-making considering the sensor data uncertainty.
The figure on the left illustrates how \workname~ leverages patients' descriptions for diagnosis when sensor data knowledge retrieval is unnecessary.
Additionally, the figures in the middle and on the right showcase two examples where \workname~ relies on sensor data knowledge for diagnostic decision-making.
We can see that when the uncertainty level of the retrieved sensor data is low, \workname~ relies more on the knowledge from the sensor data for decision-making. 
Conversely, if the uncertainty level of the retrieved sensor data is high, \workname~ requests the patient to \textbf{perform an in-lab test} to ensure the accuracy of the diagnosis.

% Figure~\ref{fig:res_dialogue_acute_bron} shows an example of diagnosis dialogue generated by our virtual doctor.

% \begin{figure}
%   \centering
% \includegraphics[width=0.95\linewidth]{Evaluation/res_dialogue_acute_bron.png}
%   \caption{An example of VDSen when diagnosing patients with the assistance of sensor data knowledge.}
%   % \vspace{-.5em}
%   \label{fig:res_dialogue_acute_bron}
%   % \vspace{-1em}
% \end{figure}

% \subsubsection{Sensor data Knowledge Retrieval Performance}

% \subsubsection{Diagnosis Guideline Retrieval Performance}

\subsection{Evaluation of System Modules}
In this section, we first perform an ablation study of \workname's individual modules including the effectiveness of adaptive sensor data retrieval filtering and mapping-based guideline tree retrieval approach.
Next, we perform experiments of \workname~ under different parameter settings.
\subsubsection{Effectiveness of Adaptive Sensor Data Retrieval Filtering}
\begin{figure}
    \centering
    \begin{subfigure}{0.49\columnwidth}
        \centering
        \includegraphics[width=0.9\textwidth]{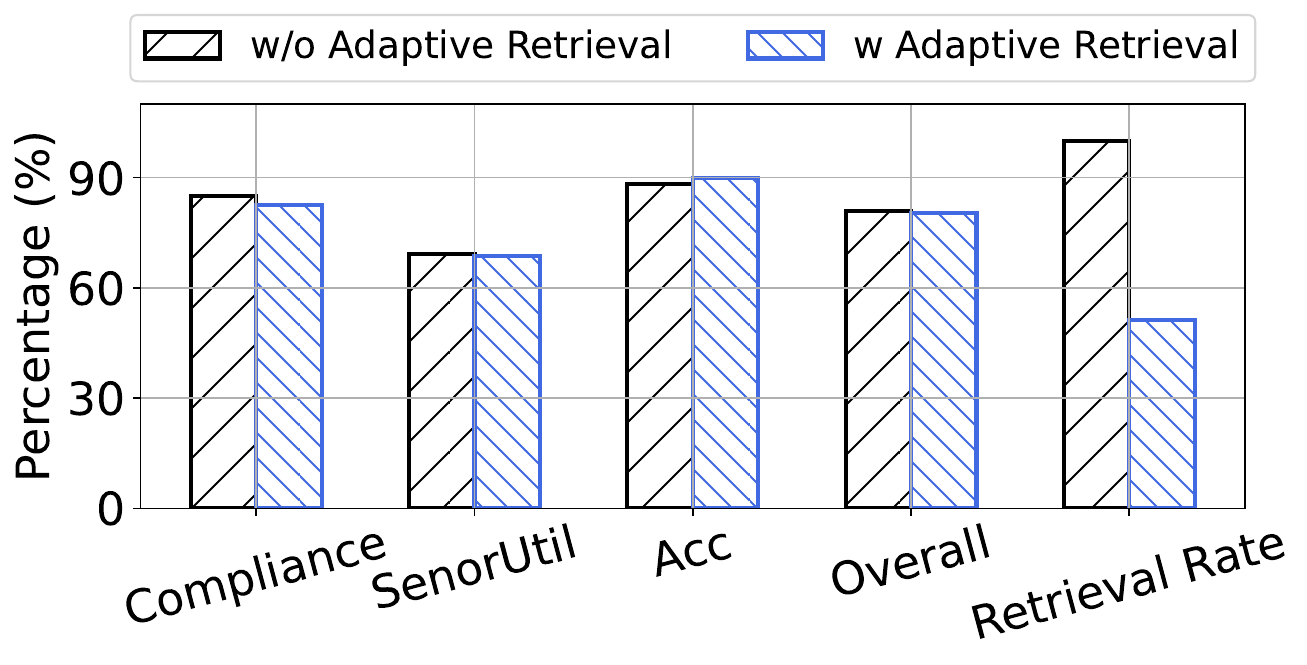}
        \caption{MedDG dataset.}  \label{fig:ar_end2end_MedDG}
    \end{subfigure}
    \hfill
    \begin{subfigure}{0.49\columnwidth}  
        \centering 
        \includegraphics[width=0.9\textwidth]{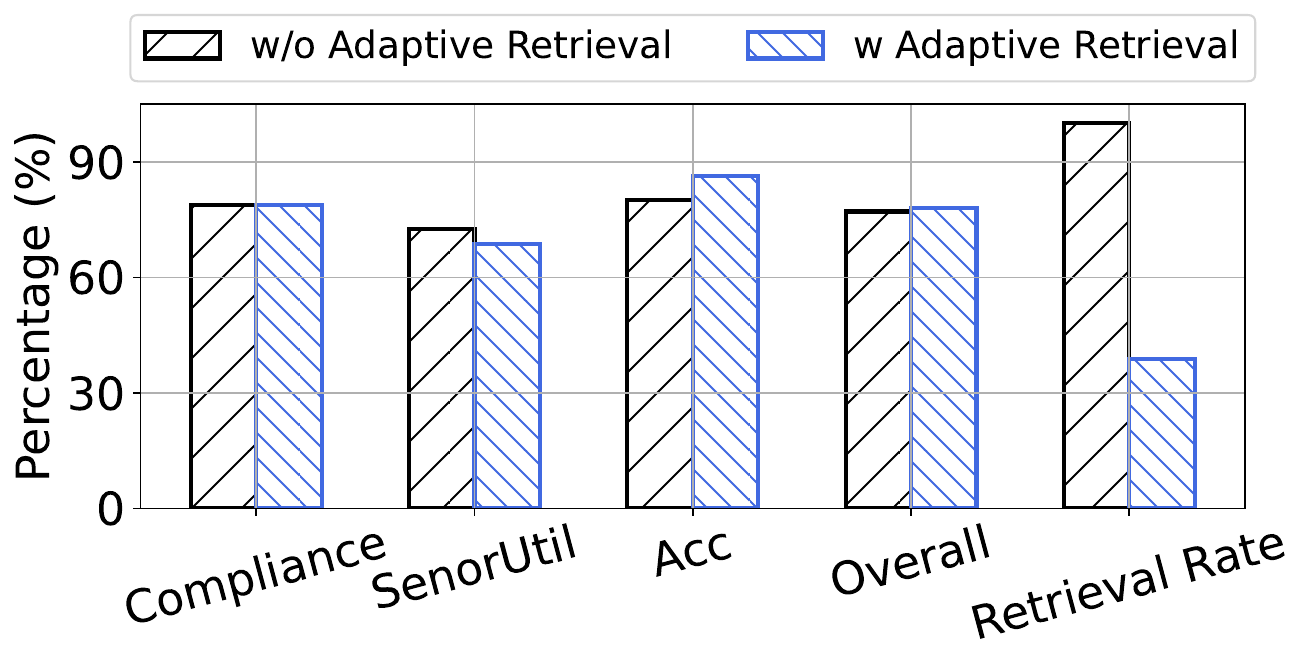}
        \caption{KaMed dataset.}    
        \label{fig:ar_KaMed_overall}
    \end{subfigure}
     \vspace{-1.0em}
    \caption{Effectiveness of adaptive sensor data retrieval filtering in \workname. 
    ``w'' and ``w/o'' denote using and not using sensor data retrieval filtering, respectively.
    The retrieval rate is defined as the ratio of the number of retrieval times to the total number of rounds in diagnostic dialogue.
    }
\label{fig:ar_overall_performance}
      % \vspace{-1.0em}
\end{figure}

We first evaluate the diagnosis performance and retrieval rate when using adaptive retrieval and not using adaptive retrieval.
The retrieval rate is defined as the ratio of the number of retrieval times to the total number of rounds in diagnostic dialogue.
Figure~\ref{fig:ar_overall_performance} shows that the utilization of adaptive retrieval in \workname~ achieves \textbf{2.2x} retrieval efficiency improvement on average while exhibiting negligible impact on the diagnosis performance metrics including \textit{compliance}, \textit{sensor data's utilization}, and \textit{accuracy}.

\subsubsection{Effectiveness of Mapping-based Guideline Tree Retrieval}

\label{guideline_retrieval_performance}
We further evaluate the effectiveness of mapping-based guideline retrieval.
Figure~\ref{fig:guideline_retrieval} shows the diagnosis guideline retrieval accuracy of \workname~ and baseline method MedDM \cite{li2023meddm}.
To the best of our knowledge, MedDM is the only work that retrieves diagnosis guidelines in LLM-based virtual doctors for multiple-turn diagnosis.
MedDM saves the medical guideline tree in the vector database, and directly retrieves the guideline tree based on patients' symptoms.  
We use MedDM with varying parameters as baselines. Specifically, the vector database of MedDM utilizes chunk sizes of 100, 200, 400, 800, 1000, and 2000, with a fixed overlapping length of 100.
Results show that mapping-based guideline retrieval achieves up to 26.0\% higher retrieval accuracy than the best baseline setting.

% \begin{figure}
%   \centering
% \includegraphics[width=0.25\linewidth]{Evaluation/guideline_retrieval.pdf}
%   \caption{Diagnosis guideline retrieval accuracy of our method and existing methods.}
%   % \vspace{-.5em}
%   \label{fig:guideline_retrieval}
%   % \vspace{-1em}
% \end{figure}

\subsubsection{Impact of Hyper-parameters}
\label{sec:exp:impact_hyper}
In this subsection, we perform ablation study and analyze the performance of \workname~ under different parameter settings.
% This subsection presents the impact of some key hyper-parameters on \workname's performance. 

\begin{figure}
    \centering
    \begin{subfigure}{0.48\columnwidth}
        \centering
        \includegraphics[width=0.9\textwidth]{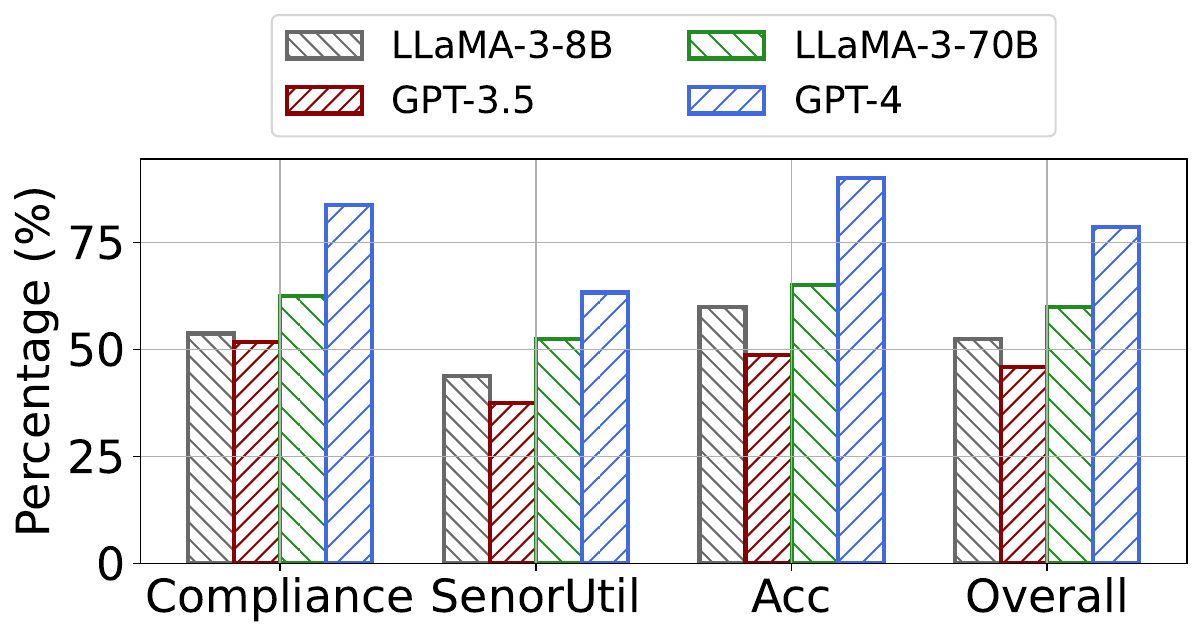}
        \caption{MedDG dataset.}  \label{fig:basemodel_MedDG}
    \end{subfigure}
    \hfill
    \begin{subfigure}{0.48\columnwidth}  
        \centering 
    \includegraphics[width=0.9\textwidth]{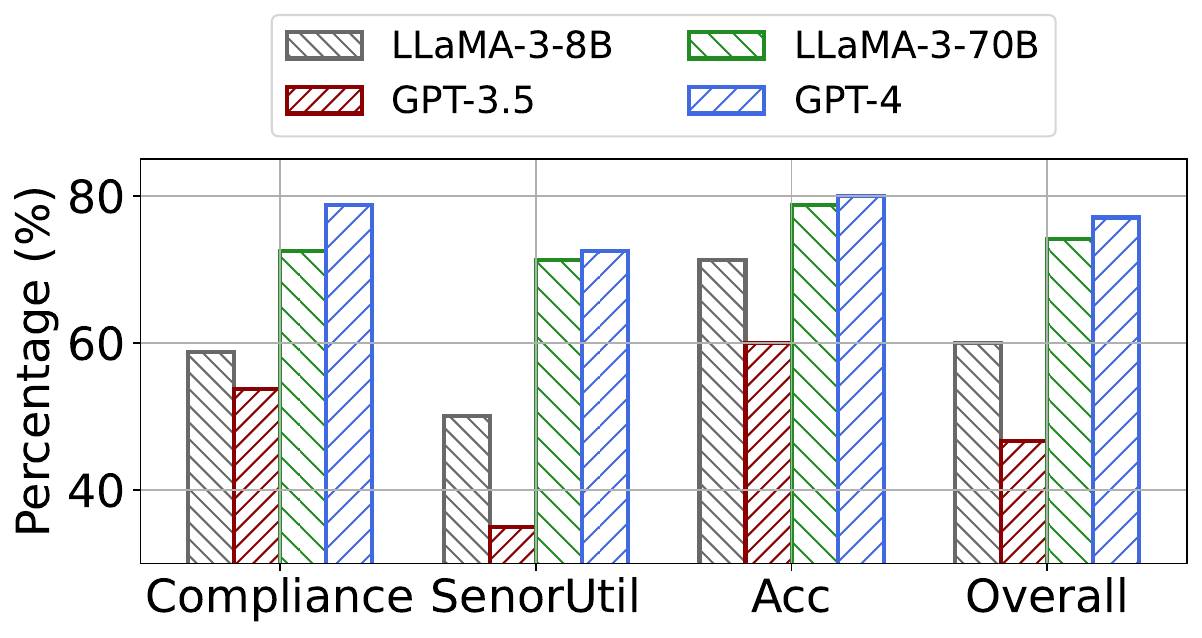}
        \caption{KaMed dataset.}    
        \label{fig:basemodel_KaMed}
    \end{subfigure}
     % \vspace{-1.0em}
    \caption{Overall performance of \workname~ on MedDG and KaMed datasets when using different base LLMs.
    }
\label{fig:base_model_MedDG_KaMed}
      % \vspace{-2.0em}
\end{figure}

\noindent\textbf{Impact of Different Base LLMs.}
We first evaluate the diagnostic performance of \workname~ using different base LLMs, including Llama-3-8b-Instruct, Llama-3-70b-Instruct, GPT-3.5-Turbo, and GPT-4-1106.
Figure~\ref{fig:base_model_MedDG_KaMed} shows that using the GPT-4-1106 as the base LLM in \workname~ yields the best diagnostic performance, achieving 13.7\%, 8.5\%, 13.8\%, 9.5\% higher \textit{compliance}, \textit{sensor data's utilization}, \textit{accuracy}, and average score than the Llama-3-70b-Instruct, respectively.
Among the three LLMs, using GPT-3.5-Turbo as the base LLM achieves the lowest diagnostic performance.
It should be noted that Llama-3-8b-Instruct and Llama-3-70b-Instruct are open-source LLMs.
They exhibit faster inference speed and can be deployed on the edge devices of patients, making them a promising choice as the base LLM in \workname.

\begin{figure}
\begin{minipage}[t]{0.45\columnwidth}
     \centering
\includegraphics[width=0.93\textwidth]{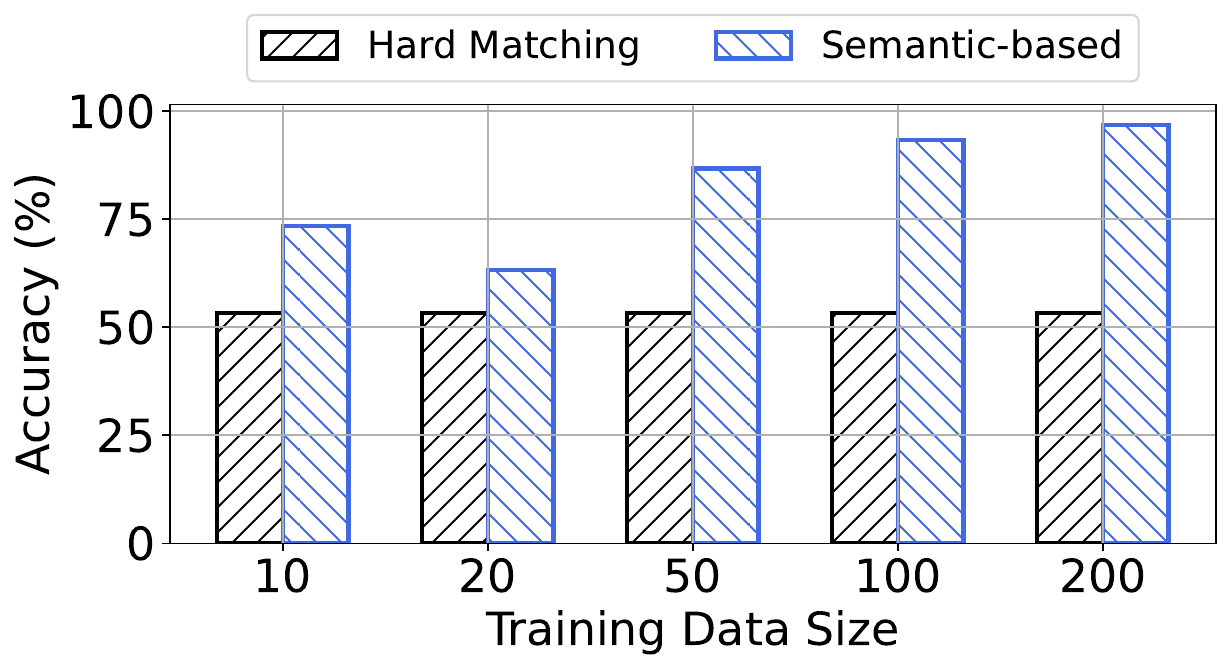}
        \vspace{-1.0em}
        \caption{Adaptive retrieval performance of semantic-based retrieval filtering and hard matching.
        }
\label{fig:Match_baseline}
\end{minipage}
\hfill
  \begin{minipage}[t]{0.48\columnwidth}
     \centering
\includegraphics[width=0.97\textwidth]{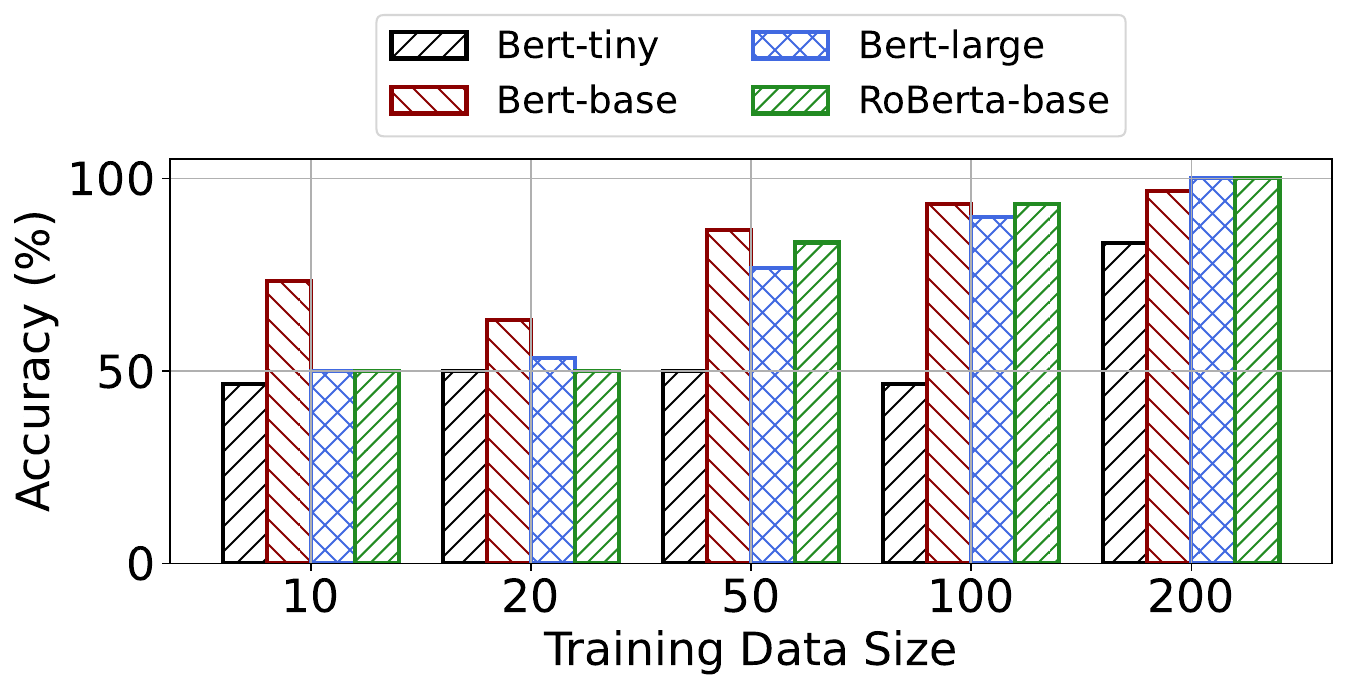}
  \vspace{-1.0em}
  \caption{The retrieval filtering accuracy when using different models as the semantic filter.}
  % \vspace{-.5em}
  \label{fig:Train_size}
\end{minipage}
\vspace{-1em}
\end{figure}

\noindent\textbf{Effectiveness of Semantic-based Retrieval Filtering.}
% \sy{hard matching has moved to methods.}
In this experiment, we study the retrieval filtering accuracy of hard matching \cite{jaro1989advances} and our semantic-based retrieval filtering under different training data size settings, which is shown in Figure~\ref{fig:Match_baseline}. 
Since hard matching \cite{jaro1989advances} does not rely on explicit training, its performance remains unaffected by the size of the training data.
Figure~\ref{fig:Match_baseline} shows that semantic-based retrieval filtering achieves 10\% higher \textit{accuracy} compared to the hard matching approach when the training data size is limited, such as 20.
However, when abundant training data is available, it can achieve  43.3\% higher \textit{accuracy}, showing the effectiveness of our proposed semantic-based retrieval filtering.

\begin{figure}
\begin{minipage}[t]{0.48\columnwidth}
     \centering
\includegraphics[width=0.85\textwidth]{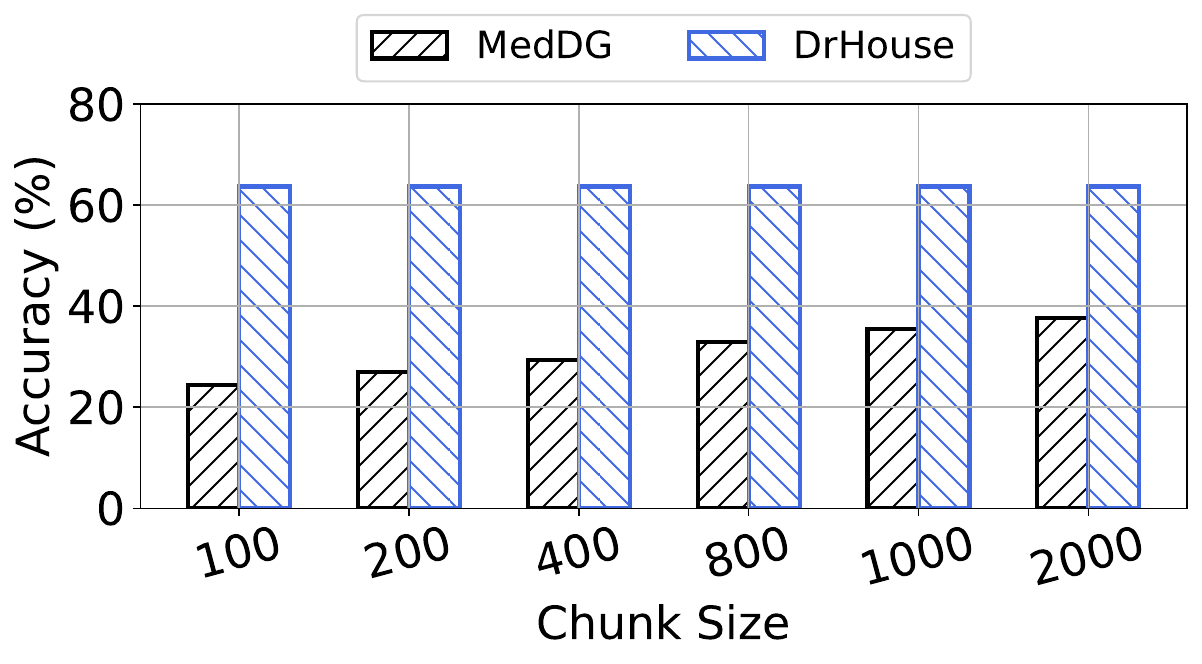}
\vspace{-1.0em}
  \caption{
  Diagnosis guideline retrieval accuracy of \workname~ and baseline method MedDM.
  The vector database of MedDM utilizes chunk ranges from 100 to 2000.}
  % \vspace{-.5em}
\label{fig:guideline_retrieval}
\end{minipage}
\hfill
  \begin{minipage}[t]{0.46\columnwidth}
     \centering
\includegraphics[width=0.95\textwidth]{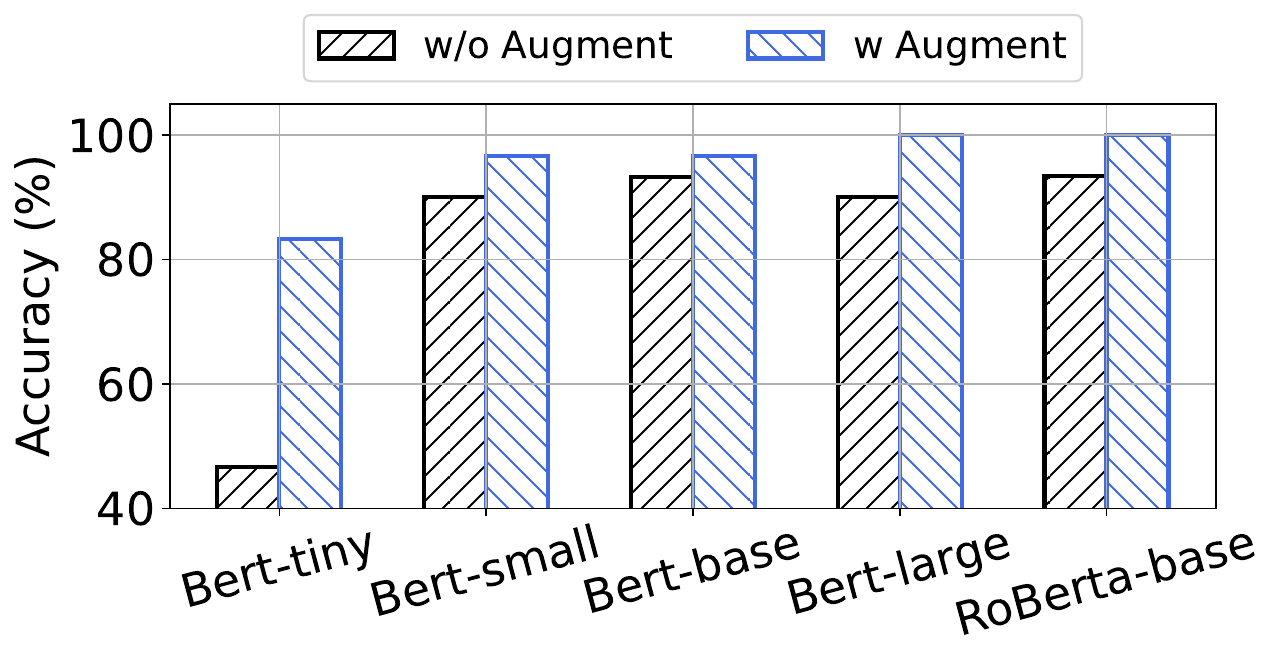}
\vspace{-1.0em}
  \caption{Adaptive retrieval performance in multiple models.
  w and w/o denote using and not using GPT-4 for data augmentation, respectively.
  }
  % \vspace{-.5em}
  \label{fig:models}
\end{minipage}
\vspace{-1em}
\end{figure}

% \begin{figure}
%   \centering
% \includegraphics[width=0.35\linewidth]{Evaluation/Match_baseline.pdf}
%   \caption{Adaptive retrieval performance of semantic-based retrieval filtering and match-based filtering.}
%   % \vspace{-.5em}
%   \label{fig:Match_baseline}
%   % \vspace{-1em}
% \end{figure}

\noindent \textbf{Effect of Different Models as Semantic Filter.}
Next, we evaluate the accuracy of retrieval filtering by employing various popular language models, such as different types of Bert and RoBerta models \cite{liu2019roberta}, as the semantic filter.
Figure~\ref{fig:Train_size} shows that Bert-base achieves 46.7\% higher \textit{accuracy} compared to Bert-tiny while it only demonstrates a 3.3\% lower \textit{accuracy} compared to Bert-large.
However, the model size of Bert-base (110M) is 3 times smaller than Bert-large (340M).
Therefore, Bert-base is employed as the semantic filter taking into account both the accuracy and overhead.
% \begin{figure}
%   \centering
% \includegraphics[width=0.45\linewidth]{Evaluation/models.pdf}
%   \caption{Adaptive retrieval performance with different models.}
%   % \vspace{-.5em}
%   \label{fig:models}
%   % \vspace{-1em}
% \end{figure}
% \noindent \textbf{Effect of Training Data Size.}
In addition, to evaluate the impact of training data size, we use datasets of varying sizes to train the semantic filter.
Figure~\ref{fig:Train_size} illustrates that with a training dataset of 200 samples, the Bert-base model achieves a filtering \textit{accuracy} of 96.7\%. 
This result demonstrates the model's ability to accurately determine whether the retrieval of sensor data should be enabled, based on the \workname's questions.

\noindent \textbf{Effect of Data Augmentation.}
We further evaluate the effectiveness of data augmentation in semantic-based retrieval filtering, which is shown in Figure~\ref{fig:models}.
For each training data size, we utilize GPT-4 to rewrite the queries, doubling the training data size compared to its original size.
Figure~\ref{fig:models}  shows that by employing GPT-4 to rewrite the queries of the LLM-based virtual doctor, there is a significant enhancement in the \textit{accuracy} of retrieval filtering. 
The \textit{accuracy} of Bert-tiny, Bert-small, and Bert-base increases by 36.7\%, 6.7\%, and 3.4\%, respectively.
Results show that our semantic filter can well understand \workname~’s questions and query sensor data on demand.

\subsection{User Study}
\label{sec_user_study}
% The above results illustrate the performance of \workname~ on public datasets. 
% However, we have not yet 

To evaluate the user experience in real-world diagnosis, We conduct a user study on the two types of target users of \workname, i.e., patients (using \workname~ for disease diagnosis) and clinicians (using \workname~ for diagnosis assistance).
% we conduct a user study involving both actual patients and expert clinicians. 
% Then we conduct an analysis based on their feedback.

\subsubsection{User Study for Medical Experts}
We first conduct a user study for medical experts.
We recruit 20 medical experts including clinicians and PhDs in medicine.
The diagnosis dialogues between \workname~ and patients are presented to the medical experts for evaluation.
We design a questionnaire comprising six questions and gather feedback from these medical experts on their experience using \workname.
The questions are as follows:
\begin{itemize}[leftmargin=*]
% \item 
% \textbf{\textit{Q1: }}
% Have you ever used other virtual doctor products for diagnosis assistance before?

\item
\textbf{\textit{Q1: }}
Does the virtual doctor correctly diagnose the diseases, and if yes, to what extent?

\item
\textbf{\textit{Q2: }}
Does the virtual doctor’s diagnostic process align with medical standards, and if yes, to what extent?

\item
\textbf{\textit{Q3: }}
Is the diagnostic style of the virtual doctor consistent with yours, and if applicable, to what extent?

\item
\textbf{\textit{Q4: }}
Do you think our virtual doctor could assist you with your diagnosis, and if applicable, to what extent?

\item
\textbf{\textit{Q5: }}
Would you be willing to use our virtual doctor during the diagnostic process?

\item
\textbf{\textit{Q6: }}
Do you think the design of our virtual doctor is novel and practical?

\end{itemize}

\begin{figure}
    \centering
    \begin{subfigure}{0.3\columnwidth}
        \centering
        \includegraphics[width=0.95\textwidth]{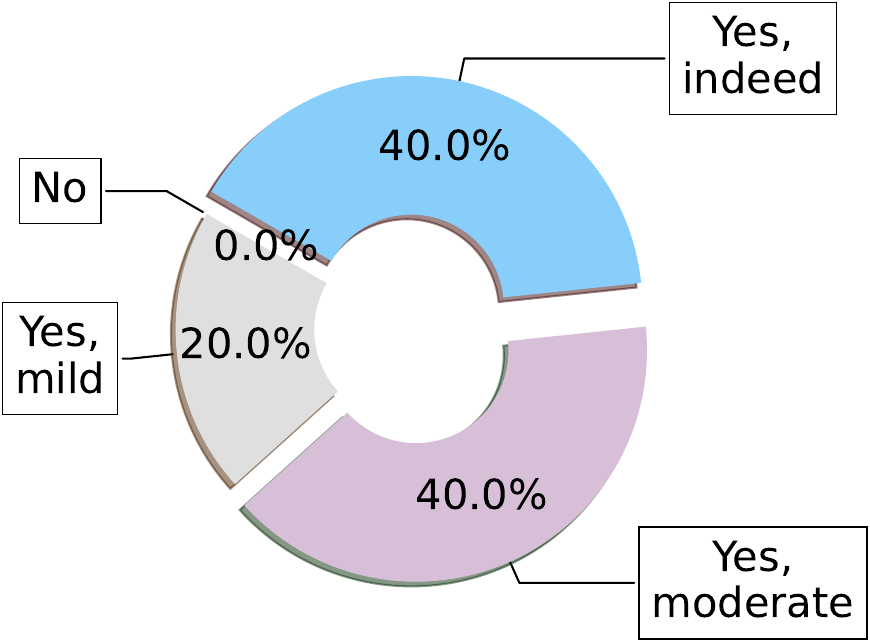}
        \caption{Does the virtual doctor correctly diagnose the disease, and if yes, to what extent?}  \label{fig:user_study_doctor_Q2}
    \end{subfigure}
    \hfill
    \begin{subfigure}{0.3\columnwidth}  
        \centering 
        \includegraphics[width=1.02\textwidth]{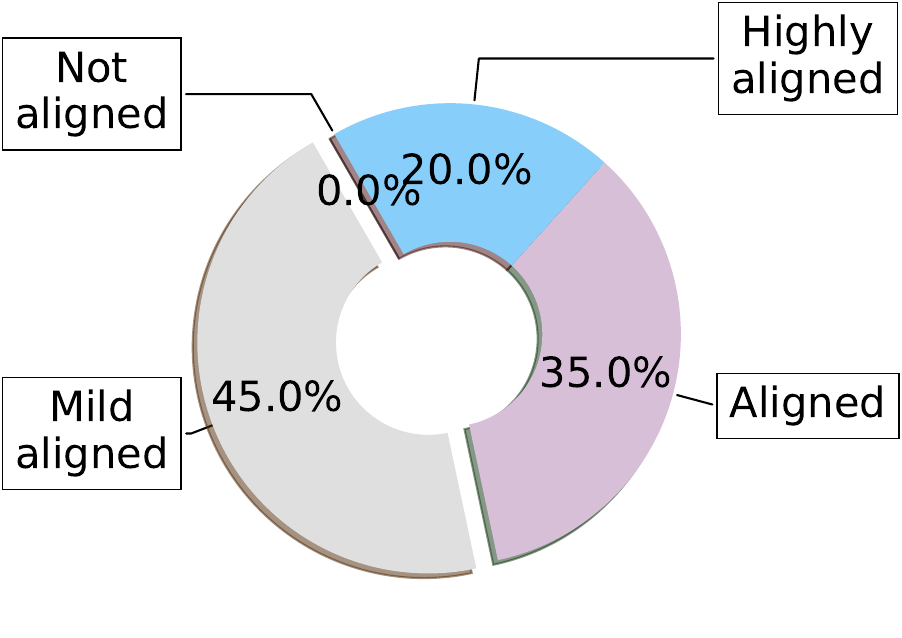}
        \caption{Does the virtual doctor’s diagnostic process align with medical standards, and if yes, to what extent?}    
        \label{fig:user_study_doctor_Q3}
    \end{subfigure}
    \hfill
    \begin{subfigure}{0.3\columnwidth}  
        \centering 
        \includegraphics[width=1.01\textwidth]{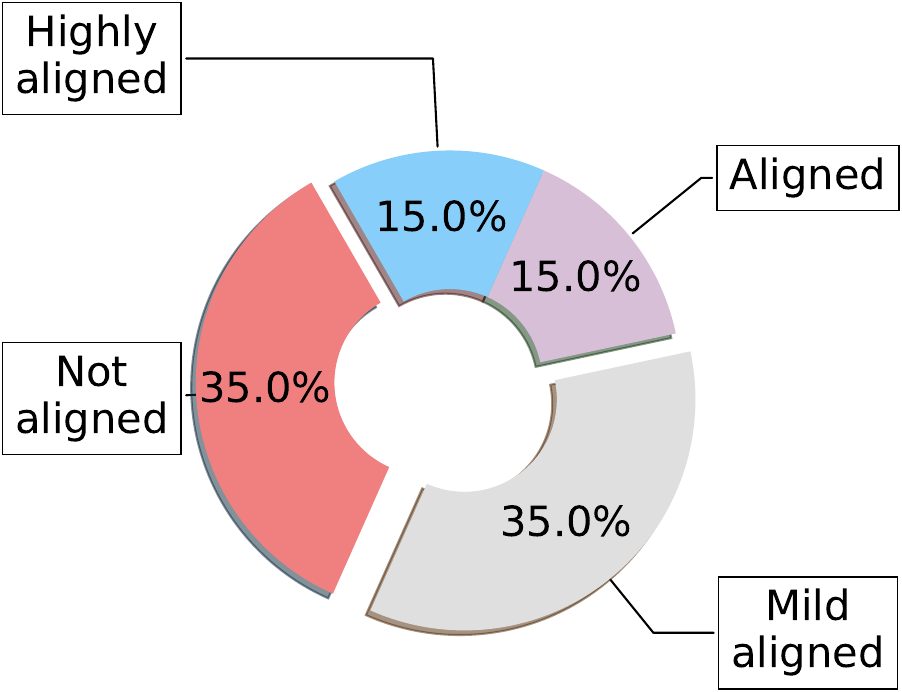}
        \caption{Is the diagnostic style of the virtual doctor consistent with yours, and if yes, to what extent?}    
        \label{fig:user_study_doctor_Q4}
    \end{subfigure}
     % \vspace{-1.0em}
    \vskip\baselineskip
    \begin{subfigure}{0.3\columnwidth}   
        \centering 
        \includegraphics[width=0.92\textwidth]{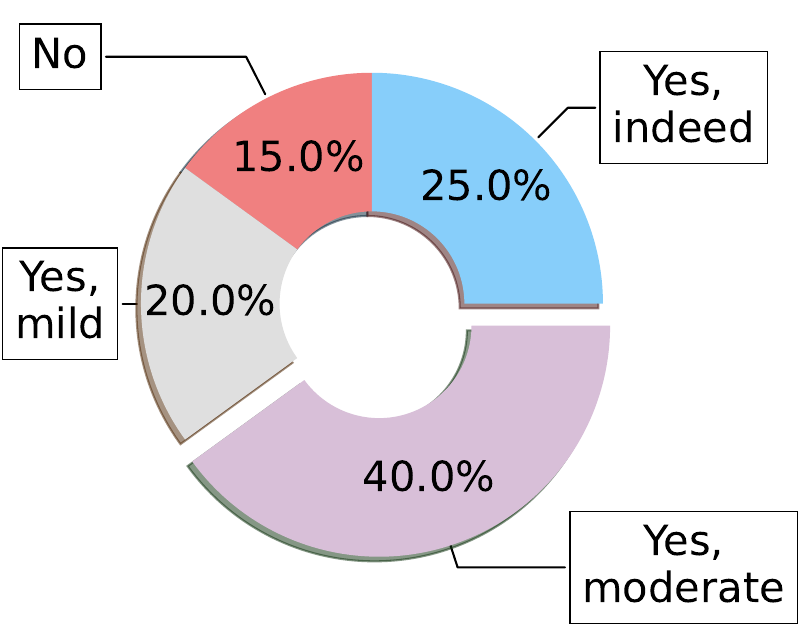}
        \caption{Do you think our virtual doctor could assist you with your diagnosis, and if yes, to what extent?}    \label{fig:user_study_doctor_Q5}
    \end{subfigure}
    \hfill
    \begin{subfigure}{0.3\columnwidth}   
        \centering 
        \includegraphics[width=1\textwidth]{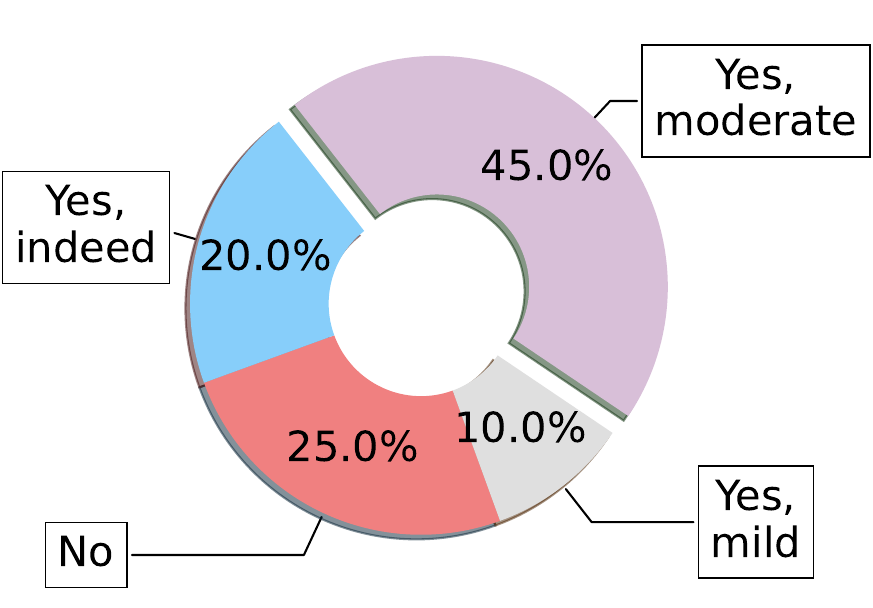}
        \caption{Would you be willing to use our virtual doctor during the diagnostic process?}    \label{fig:user_study_doctor_Q6}
    \end{subfigure}
    \hfill
    \begin{subfigure}{0.3\columnwidth}  
        \centering 
        \includegraphics[width=\textwidth]{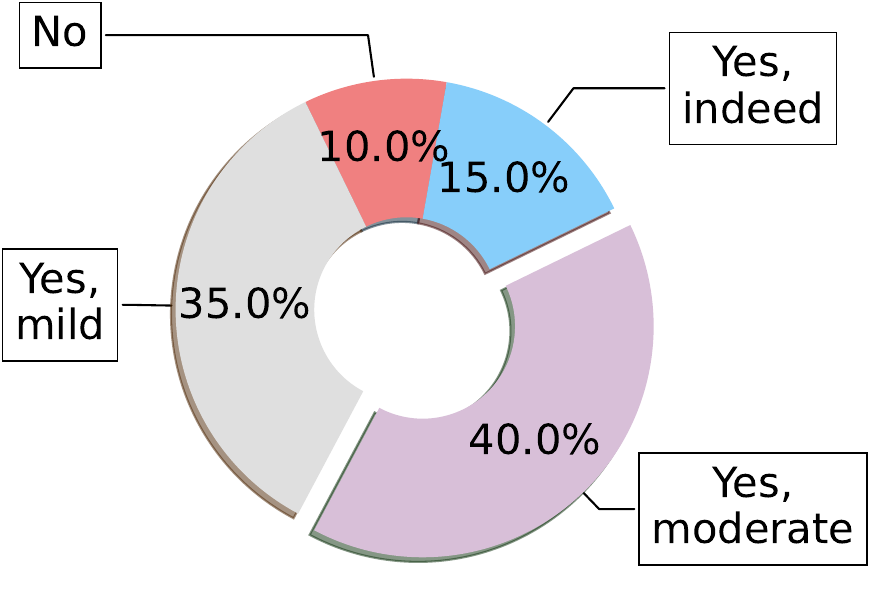}
        \caption{Do you think the design of our virtual doctor is novel and practical, and if applicable, to what extent?}    
        \label{fig:user_study_doctor_Q7}
    \end{subfigure}
     % \vspace{-1.0em}
    \caption{\workname's user study for medical experts.
    }
    \label{fig:user_study_doctor}
      \vspace{-1.0em}
\end{figure}

Figure~\ref{fig:user_study_doctor} shows the study results of medical experts. The feedback shows that 80\% of medical experts find the diagnosis of \workname~ aligned or mild aligned with the standard diagnostic procedures, and 20\% of medical experts think the diagnosis is highly aligned.
In addition, 80\% of medical experts think \workname~ correctly diagnoses the diseases.
However, there are 35\% of medical experts think the diagnostic style of \workname~ is not consistent with their own.
This could be due to inconsistencies in clinical doctors' diagnostic styles.
In addition, 85\% of medical experts think \workname~ is beneficial for their diagnosis, and 75\% of medical experts would like to use \workname~ during their diagnostic process.
These experts agree that the diagnostic consultations between patients and \workname~ can provide valuable references and help reduce their workload.
Besides, they think the demographics knowledge utilized by \workname~plays a crucial role as a relevant and appropriate reference in these consultations.
In addition, 90\% of the medical experts believe the design of \workname~ is novel and practical. 

Overall, the feedback from medical experts suggests that \workname's diagnoses are universally considered reliable. 
The medical experts express their willingness to utilize \workname~ as a diagnostic assistance tool, showing its promising market potential.

% 对于医生的usert study，我们招募了10名医生以及医学专业的硕博研究生。我们将我们的\workname~的设计，诊断对话，交给这些医学专家进行评价。

\subsubsection{User Study for Patients}
We also conduct a user study for patients.
We recruit 12 patients experiencing diverse diseases, including acute bronchitis, pneumonia, influenza, and dermatitis.
In the experiment, each participant is equipped with an Apple Watch, which captures physiological data of the patient, such as sleep condition, heartbeat, respiration rate, and oxygen saturation. 
The collected sensor data is then exported and converted into a format that enables \workname~to access the recorded sensor data throughout the patient's illness.
% Each participant in our experiment wears an Apple Watch, from which the personal sensor data are exported and converted to the data format allowing \workname~ to access the recorded sensor data throughout the patient's illness.
Next, patients participate in a multi-round conversation with \workname~ for disease diagnosis, followed by filling out a questionnaire related to their user experience.
The questions are as follows:

\begin{itemize}[leftmargin=*]
% \item 
% \textbf{\textit{Q1: }}
% Have you ever used other virtual doctor products for diagnosis assistance before?

\item
\textbf{\textit{Q1: }}
Have you ever had an experience with virtual doctors for consultations before?

\item
\textbf{\textit{Q2: }}
Are you satisfied with the diagnosis provided by the virtual doctor, if yes, to what extent?

\item
\textbf{\textit{Q3: }}
Can you accept the response delay of the virtual doctor, if yes, to what extent?

\item
\textbf{\textit{Q4: }}
Are you willing to use the virtual doctor for disease diagnosis, if yes, to what extent?

\item
\textbf{\textit{Q5: }}
Are you willing to provide sensor data, such as respiratory and heart rate, during the diagnostic process, if yes, to what extent?

\item
\textbf{\textit{Q6: }}
Do you think the design of our virtual doctor is novel and practical?

\end{itemize}

% 对于病人的user study，我们招募了10名病人。我们为这些病人佩戴了apple watch。\workname~可以读取病人在生病期间的传感器数据。然后我们让病人与我们的VDSesen进行多轮对话，进行疾病诊断。最后让病人填写他们的使用体验。

\begin{figure}
    \centering
    \begin{subfigure}{0.3\columnwidth}
        \centering
        \includegraphics[width=0.7\textwidth]{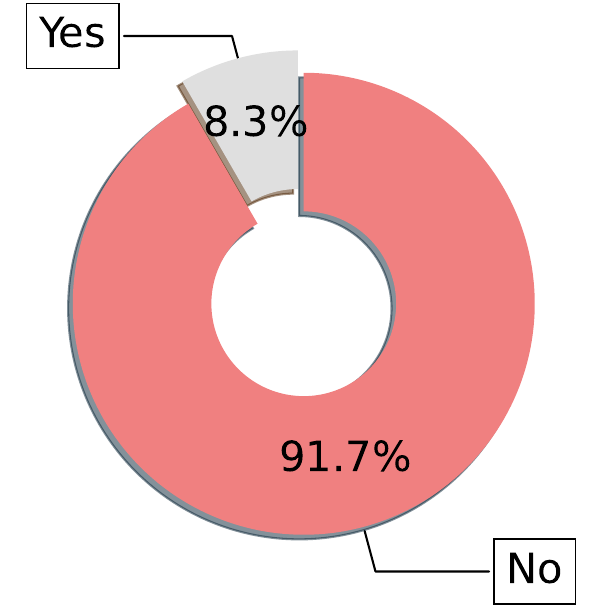}
        \caption{Have you ever had an experience with virtual doctors for medical consultations before?}  \label{fig:user_study_patient_Q1}
    \end{subfigure}
    \hfill
    \begin{subfigure}{0.3\columnwidth}  
        \centering 
        \includegraphics[width=1.05\textwidth]{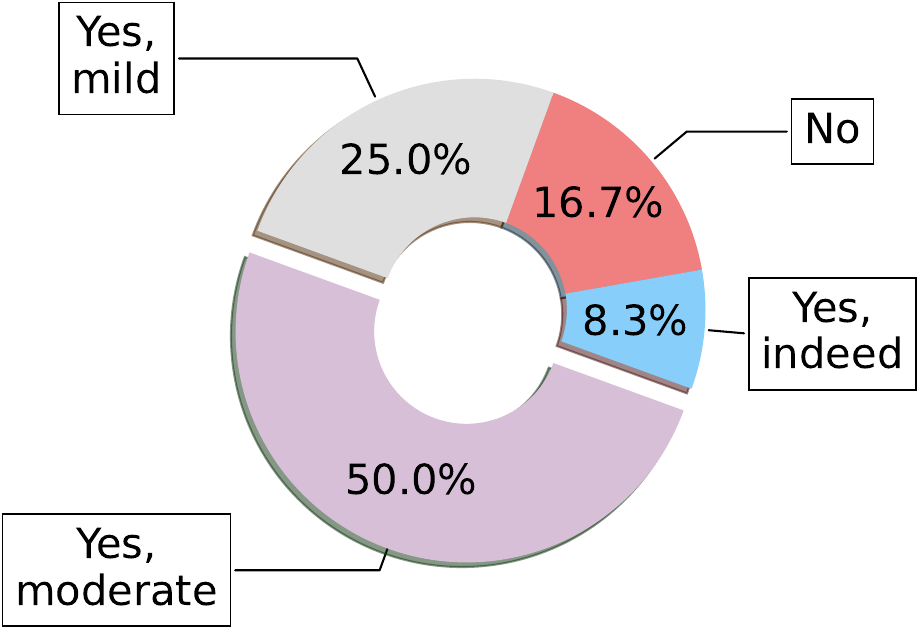}
        \caption{Are you satisfied with the diagnosis provided by the virtual doctor, if yes, to what extent?}    
        \label{fig:user_study_patient_Q2}
    \end{subfigure}
    \hfill
    \begin{subfigure}{0.3\columnwidth}  
        \centering 
        \includegraphics[width=0.9\textwidth]{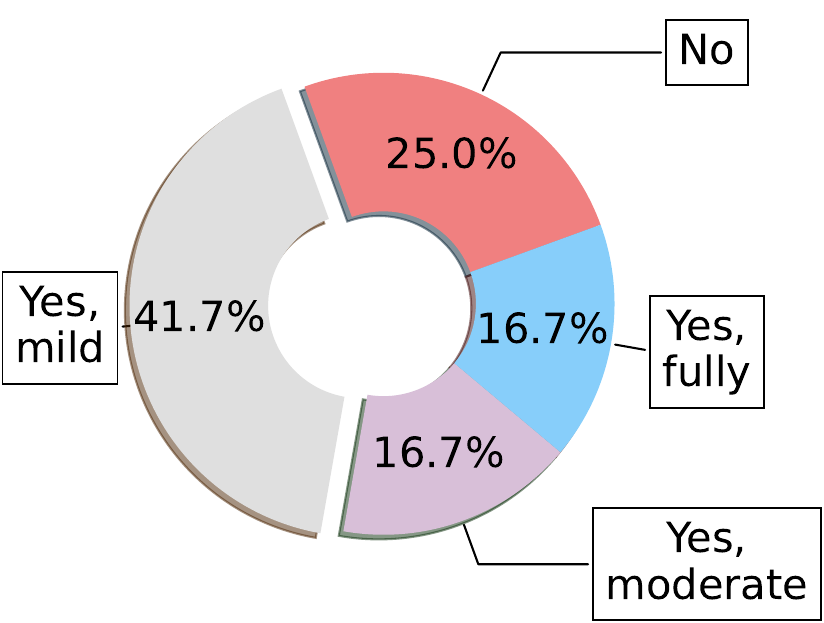}
        \caption{Can you accept the response delay of the virtual doctor, if yes, to what extent?}    
        \label{fig:user_study_patient_Q3}
    \end{subfigure}
     % \vspace{-1.0em}
    \vskip\baselineskip
    \begin{subfigure}{0.3\columnwidth}   
        \centering 
        \includegraphics[width=1.04\textwidth]{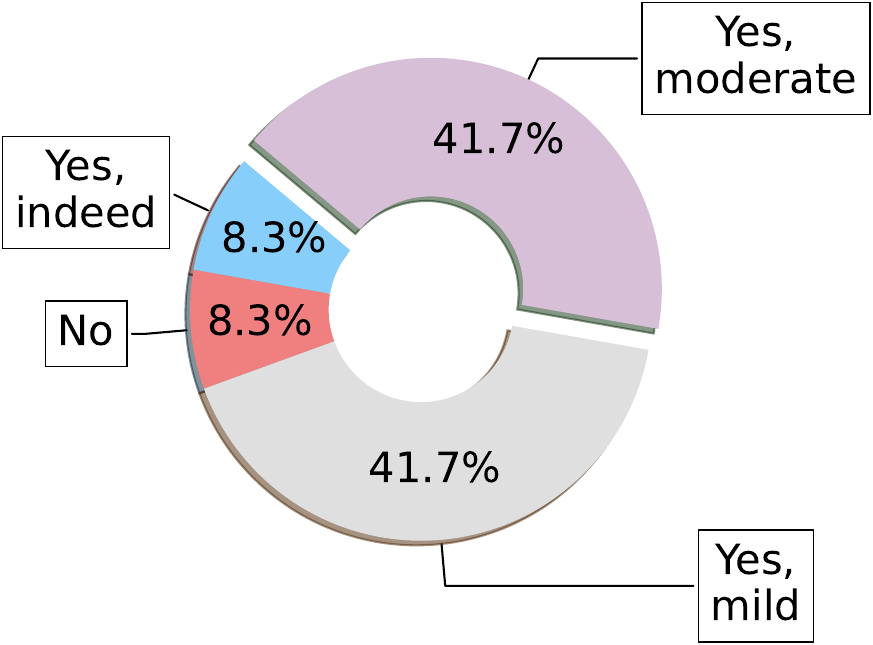}
        \caption{Are you willing to use the virtual doctor for disease diagnosis, if yes, to what extent?}    \label{fig:user_study_patient_Q4}
    \end{subfigure}
    \hfill
    \begin{subfigure}{0.3\columnwidth}   
        \centering 
        \includegraphics[width=1.02\textwidth]{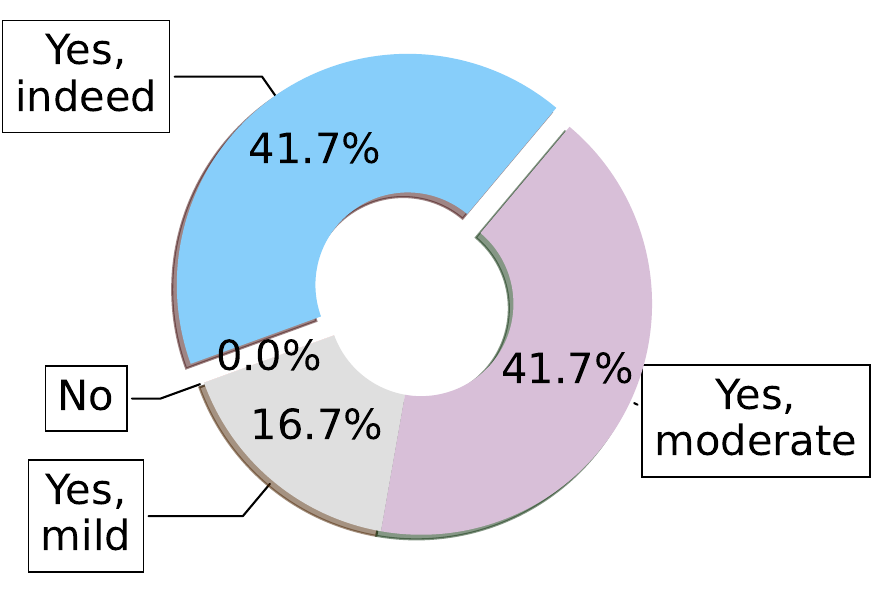}
        \caption{Are you willing to provide sensor data during the diagnostic process, if yes, to what extent?}    \label{fig:user_study_patient_Q5}
    \end{subfigure}
    \hfill
    \begin{subfigure}{0.3\columnwidth}  
        \centering 
        \includegraphics[width=0.95\textwidth]{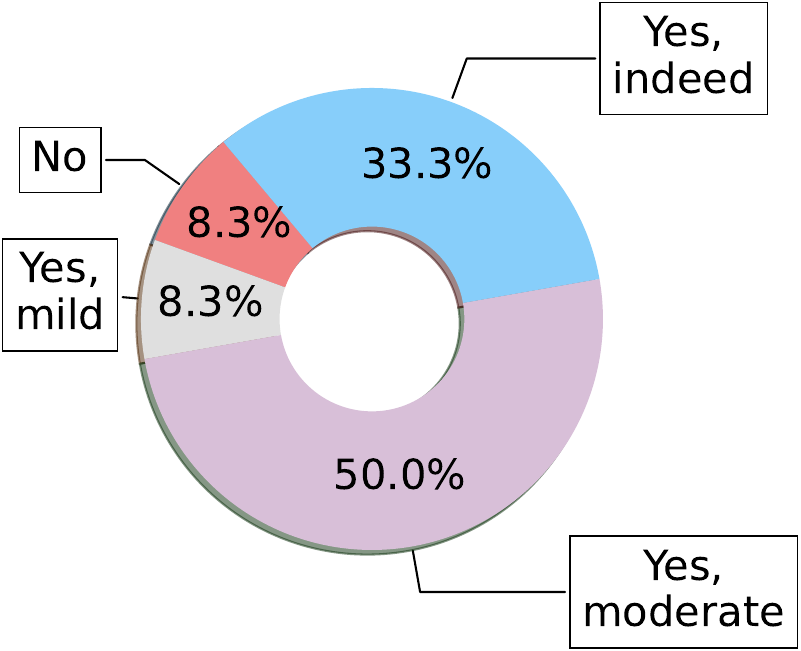}
        \caption{Do you think the design of our virtual doctor is novel and practical, if yes, to what extent?}    
        \label{fig:user_study_patient_Q6}
    \end{subfigure}
     % \vspace{-1.0em}
    \caption{\workname's user study for patients.
    }
    \label{fig:user_study_patient}
      \vspace{-1.0em}
\end{figure}

The results in Figure~\ref{fig:user_study_patient} show that only 8.3\% of the participants have previously consulted virtual doctors.
In addition, 83.4\% of patients report satisfaction with the diagnoses provided by \workname. 
Regarding the response delay of \workname, 75\% of the participants find it acceptable. 
Note that this delay is primarily due to the instability of the GPT-4 API, which can be further improved.
Moreover, despite these delays, most patients find using \workname~was much more convenient than visiting a hospital in person.
Furthermore, 91.7\% of the participants express willingness to use \workname~ for future diagnoses.
When asked for access to personal sensor data, 83.4\% of the patients are strongly or moderately willing to provide such data to \workname~ for diagnostic purposes. 
A key reason for the high willingness to share data is that most patients in the user study have already authorized smartphone apps to access this data.
Consequently, they are generally comfortable providing this data to another application, particularly one that offers health consultation services.
Additionally, 91.7\% of the patients appreciate the innovative and practical design of \workname.
Overall, the feedback from patients in the user study indicates that \workname~ can provide reliable and satisfactory medical diagnoses from the patient's perspective.
The substantial willingness of participants to share personal sensor data and to utilize \workname~ for future medical consultations demonstrates the promising practical applications of \workname.

% Methods(participants)
% Results: doctor
% Results: patients

% \subsubsection{Ablation Study}
% \textcolor{blue}{???}
% \xl{? baselines with sensor data}

\section{Discussion}
\noindent\textbf{Other Medical LLMs as Base Models.}
% base LLM / medical LLM
In this study, we perform experiments using different base LLMs in \workname, including Llama-3, GPT-3.5, and GPT-4.
% Since GPT-4 possesses the strongest reasoning and generalization capabilities among existing LLMs \cite{achiam2023gpt}, we employ GPT-4 as the base LLM to construct the virtual doctor in this study.
Leveraging existing medical LLMs as the base model can also be considered.
However, renowned virtual doctors like Med-PaLM 2 \cite{singhal2023towards} are not open source.
In addition, existing open-sourced medical LLMs like HuatuoGPT \cite{chen2023huatuogpt} and DISC-MedLLM \cite{bao2023disc} adapt the small LLMs (7B or 13B) to the medical domain through fine-tuning, compromising their generalization capability.
We leave exploring other existing medical LLMs as the base LLM in \workname~ as our future work.

% \subsection{Subjective Factors in Diagnosis}
\noindent\textbf{Subjective Factors in Diagnosis.}
Figure~\ref{fig:user_study_doctor} shows that there are 35\% of medical experts think the diagnostic style of \workname~ is not consistent with their own.
This is because each doctor possesses a unique diagnostic style. 
Some doctors lean towards a conservative approach and adhere strictly to diagnostic guidelines, while others rely heavily on their own experience. 
Therefore, ensuring the consistency of \workname's diagnostic style with every doctor is challenging. 
On the other hand, diagnostic experiences tend to be highly subjective. Currently, \workname~ only relies on objective medical guidelines for diagnosis. 
In the future, \workname~ can be improved by incorporating diagnostic dialogues encompassing doctors with diverse styles, enabling \workname~ with varying diagnostic styles.

\noindent\textbf{Diagnostic Response Delay.}
The base LLM in \workname~ is deployed on a cloud server, utilizing the API of the LLM service provider for inference.
However, the latency of these API calls is not stable. 
The inference delay of the LLM can be prolonged due to server congestion and instability.
However, according to user study feedback from real patients (Figure~\ref{fig:user_study_patient}), the majority of patients think the diagnostic delay of \workname~ is acceptable.
This is because the delay experienced with \workname~ is significantly shorter than the time required for traveling to a hospital.
In the future, we will consider deploying \workname's base LLM on mobile devices or employing edge-cloud collaboration architecture \cite{edgefm} to further protect patient privacy and reduce diagnostic delays.

\noindent\textbf{Scalability to Other Sensor Data Summarizing Approach.}
\workname~ focuses on incorporating the sensor data knowledge from patients' smart devices to multi-turn diagnosis of LLM-based virtual doctors.
In particular, \workname~ utilizes an LLM to summarize the patient's sensor data recordings from smart devices.
Recent studies \cite{jin2024position,hota2024evaluating,xu2024penetrative,kim2024health,englhardt2023classification,ji2024hargpt,yang2024you} have shown that LLMs can be used to interpret various raw sensor data, such as IMU.
These techniques for LLMs interpreting sensor data can also be integrated into \workname, enhancing the diversity and information of the sensor data that \workname~ can utilize.

\noindent\textbf{Other Sensor Data Uncertainty Check Approaches.} 
\workname~ incorporates the uncertainty of patients' sensor data into its decision-making process. 
When the uncertainty level is elevated, \workname~will prompt patients to undergo in-lab tests to ensure a more accurate diagnosis. 
Many existing approaches on anomaly detection tasks ~\cite{tuli2022tranad,du2021gan,liu2022time} can also be integrated into \workname~ to enhance the system’s ability to verify sensor data reliability.

\section{Conclusion}
This paper proposes \workname, the first LLM-empowered virtual doctor system that incorporates patients's daily sensor data and expert knowledge for multi-turn diagnosis.
\workname~ leverages the latest diagnosis guidelines to actively initiate multi-turn diagnosis.
During the multi-turn diagnosis process, \workname~ employs a multi-source knowledge retrieval approach to retrieve required sensor data knowledge and expert knowledge.
In addition, \workname~ 
integrates the patient's descriptions and the two types of knowledge for diagnostic decision-making.
% At the same time, \workname~ performs concurrent checking of candidate diseases with likelihood, allowing for more nuanced and informed medical assessments.
We evaluate \workname~ on three public datasets and conduct comprehensive user studies involving both medical experts and patients.
Results show that \workname~ outperforms existing LLM-based virtual doctors solutions by 18.8\% on diagnosis accuracy, improving the diagnosis experience of both patients and physicians.

% \workname~ progressively narrows down the range of potential diseases and finally generates an explainable diagnosis analysis for each specific disease.
% We evaluate \workname~ on three public datasets and conduct comprehensive user studies involving both medical experts and patients.
% Results show that \workname~ outperforms existing LLM-based virtual doctors solutions by 18.8\% on diagnosis accuracy, improving the diagnosis experience of both patients and physicians.

\bibliographystyle{ACM-Reference-Format}
\bibliography{drhouse}

\end{document}